\documentclass[lettersize,journal]{IEEEtran}
\usepackage{amsmath,amsfonts}
\usepackage{algorithmic}
\usepackage{algorithm}
\usepackage{array}
\usepackage[caption=false,font=normalsize,labelfont=sf,textfont=sf]{subfig}
\usepackage{textcomp}
\usepackage{stfloats}
\usepackage{url}
\usepackage{verbatim}
\usepackage{graphicx}
\usepackage{xcolor}
\usepackage{cite}
\usepackage{booktabs} 
\usepackage{multirow}
\usepackage{cleveref}
\usepackage{mathrsfs}  
\usepackage{caption}
\usepackage{marvosym}
\usepackage[normalem]{ulem}

\newcommand\up[1]{\textcolor[rgb]{0.2,0.8,0.2}{$^{\uparrow{#1}}$}}
\newcommand\down[1]{\textcolor{red}{$^{\downarrow{#1}}$}}
\newcommand\smallup[1]{\textcolor[rgb]{0.5,0.6,0.7}{$^{\uparrow{#1}}$}}
\newcommand\smalldown[1]{\textcolor[rgb]{0.5,0.6,0.7}{$^{\downarrow{#1}}$}}
\newcommand\blfootnote[1]{%
\begingroup
\renewcommand\thefootnote{}\footnote{#1}%
\addtocounter{footnote}{-1}%
\endgroup
}

\begin{document}

\title{Hulk: A Universal Knowledge Translator \\ for Human-Centric Tasks}

\author{Yizhou Wang*, Yixuan Wu*, Weizhen He, Xun Guo, Feng Zhu, Lei Bai, Rui Zhao, Jian Wu,~\IEEEmembership{Member,~IEEE,} \\ Tong He, Wanli Ouyang,~\IEEEmembership{Senior Member,~IEEE}, Shixiang Tang\textsuperscript{\Letter}
}

\markboth{Journal of \LaTeX\ Class Files,~Vol.~14, No.~8, August~2021}%
{Shell \MakeLowercase{\textit{et al.}}: A Sample Article Using IEEEtran.cls for IEEE Journals}

\twocolumn[{
\renewcommand\twocolumn[1][]{#1}%
\maketitle
\begin{center}
  \centering
  \vspace{-1em}
  \captionsetup{type=figure}
  \includegraphics[width=\linewidth]{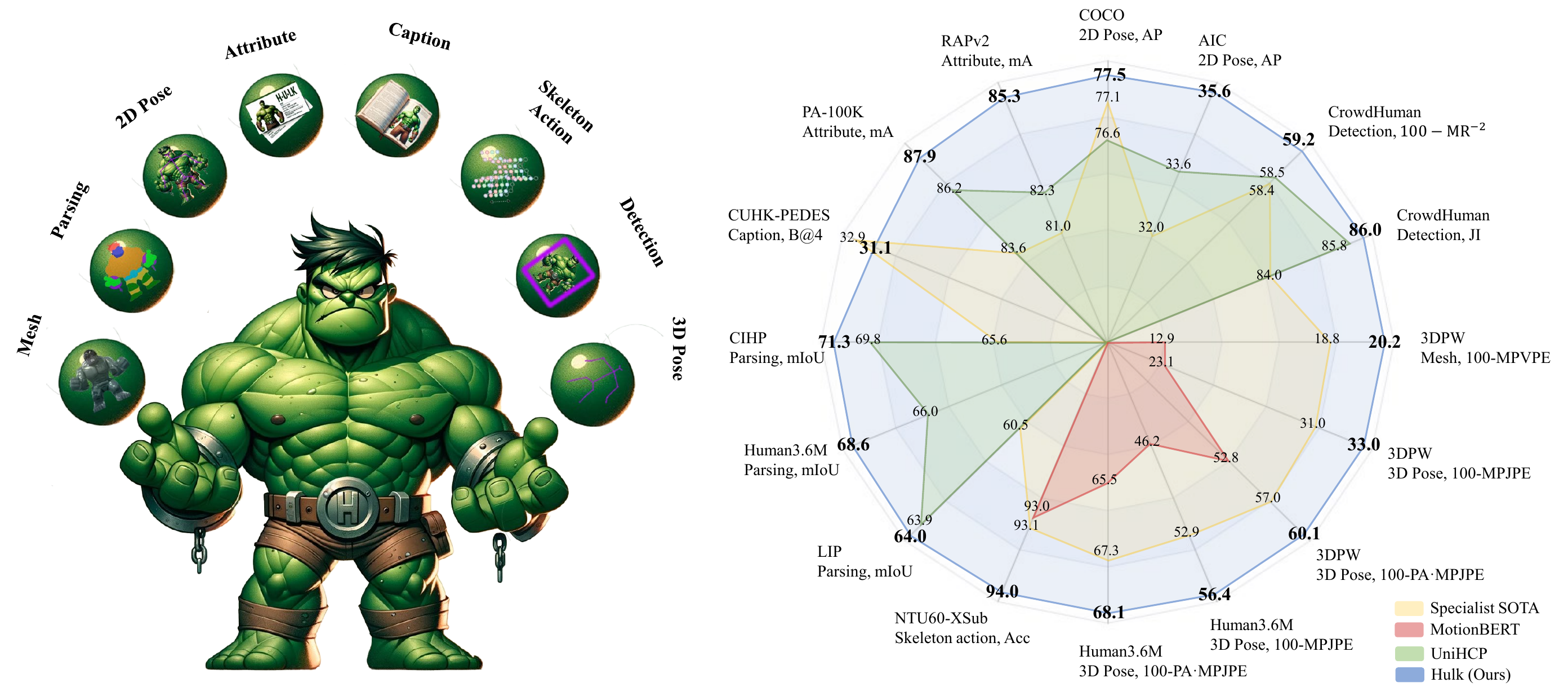}

  \captionof{figure}{
     \textbf{Left}: Our proposed Hulk is a generalist human-centric perceiver that can tackle {several common} 2D vision, 3D vision, skeleton-based, and vision language tasks using a unified framework. 
     \textbf{Right}: Hulk pushes the limits on various human-centric tasks when compared with specialist models, pretraining models, \emph{e.g.}, MotionBERT~\cite{zhu2023motionbert}, and existing generalist models, \emph{e.g.}, UniHCP~\cite{ci2023unihcp}. 
  }
  \label{fig:enter-label}
  \vspace{1em}
\end{center}%
}]

\begin{abstract}
Human-centric perception tasks, \emph{e.g.,} pedestrian detection, skeleton-based action recognition, and pose estimation, have wide industrial applications, such as metaverse and sports analysis. There is a recent surge to develop human-centric foundation models that can benefit a broad range of human-centric perception tasks. While many human-centric foundation models have achieved success, they did not explore 3D and vision-language tasks for human-centric and required task-specific finetuning. These limitations restrict their application to more downstream tasks and situations. To tackle these problems, we present Hulk, the first multimodal human-centric generalist model, capable of addressing 2D vision, 3D vision, skeleton-based, and vision-language tasks without task-specific finetuning. The key to achieving this is condensing various task-specific heads into two general heads, one for discrete representations, \emph{e.g.,} languages, and the other for continuous representations, \emph{e.g.,} location coordinates. 
The outputs of two heads can be further stacked into four distinct input and output modalities. This uniform representation enables Hulk to treat diverse human-centric tasks as modality translation, integrating knowledge across a wide range of tasks.
Comprehensive evaluations of Hulk on 12 benchmarks covering 8 human-centric tasks demonstrate the superiority of our proposed method, achieving state-of-the-art performance in 11 benchmarks. The code will be available on {https://github.com/OpenGVLab/Hulk}.

\end{abstract}


\section{Introduction}
\IEEEPARstart{R}ECENT years have witnessed increasing research attention to human-centric tasks, \emph{e.g.,} person re-identification~\cite{zhu2017fast,hou2021feature,zhang2020semantic,fu2021unsupervised}, pose estimation~\cite{rogez2019lcr,yang2012articulated,yu2018doublefusion,fang2022alphapose}, and human mesh recovery~\cite{zhu2023motionbert,loper2023smpl}, 
due to their widespread applications, including sports analysis~\cite{fernandez2019decomposing,decroos2019actions,honda2022pass},  surveillance~\cite{zahrawi2023improving,litoriya2022integrated,yadav2022real,shou2023crowdq}, augmented reality~\cite{8972114,hu2023understanding,zhen2021augmented}, and video production~\cite{su2022robustfusion,Lee_2023_CVPR,Kim_2023_CVPR}. 
\blfootnote{
$^{*}$Yizhou Wang and Yixuan Wu are contributed equally. $\textsuperscript{\Letter}$Shixiang Tang is the corresponding author. This work was done when Yixuan Wu, Xun Guo were interns, Yizhou Wang was a research assistant at Shanghai Artificial Intelligence Laboratory. Yizhou Wang and Wanli~Ouyang are with Department of Information Engineering, The Chinese University of Hong Kong, Hong Kong, China. Yixuan Wu is with the School of Medicine, Zhejiang University, Hangzhou, 310027, China. Weizhen He is with the College of Electrical Engineering, Zhejiang University, Hangzhou, 310027, China. Xun Guo is with the Department of Automation, University of Science and Technology of China, Hefei, 230052, China. Shixiang Tang is with CUHK Interdisciplinary Artificial Intelligence Research Institute, Hong Kong, China. Feng Zhu and Rui Zhao are with SenseTime Group Limited, China. 
Jian Wu is with School of Public Health, Zhejiang University, Hangzhou, 310058, China. Lei Bai, Tong He, and Wanli~Ouyang are with Shanghai Artificial Intelligence Laboratory, 
Shanghai, 200232, China.
}
While most of the previous methods typically emphasized unique designs tailored for individual tasks, involving heavy work for task-specific design, fine-tuning, and deployment, 
an emerging trend is to explore learning general representative encoders, also known as human-centric foundation models, that can be fine-tuned or prompt-tuned for different human-centric visual tasks. These models leverage large-scale pretraining on diverse datasets, and the high-related human semantics in different human-centric tasks, achieving remarkable performance across a wide range of tasks, particularly in 2D vision applications such as person re-identification~\cite{zhu2017fast,hou2021feature,zhang2020semantic,fu2021unsupervised}, pose estimation~\cite{rogez2019lcr,yang2012articulated,yu2018doublefusion}, and human parsing~\cite{zhang2022aiparsing,zhang2021correlation,liang2016clothes}. Some notable examples of such models include SOLIDER~\cite{solider}, HCMoCo~\cite{hcmoco}, MotionBERT~\cite{zhu2023motionbert}, PATH~\cite{tang2023humanbench}, HAP~\cite{yuan2023hap}, and SMPLer-X~\cite{cai2023smpler}. 
\begin{table*}[ht!]
  \centering
  \caption{Supporting tasks of existing human-centric pretraining models and generalist models. Pretraining models require fine-tuning for downstream tasks, whereas generalist models like Hulk can be directly used. Compared with other methods, Hulk can perform all mentioned tasks using shared parameters except for Person ReID and Person Search due to ethical concerns.}
    \begin{tabular}{llccccccc}
    \toprule
    \multicolumn{1}{l}{\multirow{2}[4]{*}{Task Type}} & \multirow{2}[4]{*}{Tasks} & \multicolumn{5}{c}{Pretraining}       & \multicolumn{2}{c}{Generalist} \\
\cmidrule(r){3-7} \cmidrule(r){8-9}          &       & HCMoCo & SOLIDER & PATH  & HAP   & MotionBERT & UniHCP & Hulk (ours) \\
    \midrule
    \multirow{6}[2]{*}{2D Vision Tasks} &Person ReID  &       & \checkmark   & \checkmark   & \checkmark   &       & \checkmark   &  \\
          & Person Search &       & \checkmark   &       &       &       &       &  \\
          & Human Parsing & \checkmark   & \checkmark   & \checkmark   &       &       & \checkmark   & \checkmark \\
          & 2D Pose Estimation &  \checkmark     & \checkmark   & \checkmark   & \checkmark   &       & \checkmark   & \checkmark \\
          & Pedestrian Detection &       & \checkmark   & \checkmark   &       &       & \checkmark   & \checkmark \\
          & Attribute Recognition&       & \checkmark   & \checkmark   & \checkmark   &       & \checkmark   & \checkmark \\
    \midrule
    Skeleton-based Tasks & Skeleton-based action &       &       &       & \checkmark   & \checkmark   &       & \checkmark \\
    \midrule
    Vision-language Tasks & Image Caption &       &       &       &       &       &       & \checkmark \\
    \midrule
    \multirow{2}[2]{*}{3D Vision Tasks} & Mesh Recovery &       &       &       & \checkmark   & \checkmark   &       & \checkmark \\
          & 3D Pose Estimation & \checkmark   &       &       & \checkmark   & \checkmark   &       & \checkmark \\
    \bottomrule
    \end{tabular}%
    \vspace{-1em}
  \label{tab:supporting tasks}%
\end{table*}%

Despite the success of these human-centric foundation models, they still need to tune the model for every specific downstream task, resulting in one unique model for each downstream task, \emph{e.g.,} 5 models for 5 downstream tasks. 
While UniHCP~\cite{ci2023unihcp} unified five 2D human-centric tasks with a single model, it still struggled when dealing with multimodal inputs and outputs and more complex task modeling beyond 2D human-centric vision tasks.
Therefore, existing works still have not reached the ultimate goal of the \emph{generalist human-centric perceiver} - one versatile model for all 2D vision, 3D vision and mulitmodal human-centric tasks, which mirrors ChatGPT~\cite{openaichatgptblog}, LLama~\cite{touvron2023llama} and InternLM~\cite{team2023internlm} in the field of natural language processing.

The core challenge of developing the \emph{generalist human-centric perceiver} with better versatility is the unified modeling of 2D, 3D, and multimodal tasks, considering incredibly diverse data formats and highly specialized designs in human-centric tasks:
 \begin{itemize}
     \item Human-centric perception tasks typically include 2D vision tasks, 3D vision tasks, skeleton-based tasks, and vision-language tasks, whose inputs and outputs cover diverse formats, \emph{i.e.,} images, languages, key points, segmentation maps, and categories. Unifying these incredibly heterogeneous inputs and outputs is nontrivial for a generalist human-centric perceiver.
     \item Current state-of-the-art methods for human-centric tasks differ in backbones (\emph{e.g.,} GNN~\cite{zhou2020graph} for skeleton-based action recognition and HRFormer~\cite{yuan2021hrformer} for pose estimation) and task heads (\emph{e.g.}, class guidance blocks~\cite{liu2022cdgnet} for human parsing and progressive dimensionality reduction heads~\cite{metro, fastmetro} for mesh recovery). 
    Unifying these specialized but effective architectural designs for unique tasks is demanding for developing the generalist human-centric perceiver. 
 \end{itemize}


To address the challenges, we present Hulk, a \underline{H}uman-centric \underline{u}niversa\underline{l} \underline{k}nowledge translator that can, for the first time, handle {several common} large-scale 2D vision, 3D vision, skeleton-based and vision-language tasks listed in Table~\ref{tab:supporting tasks} without task-specific finetuning. Unlike previous methods, Hulk unifies data formats and task modeling by condensing diverse task inputs and outputs into four modalities and treating heterogeneous human-centric tasks as modality translation tasks. First, Hulk condenses diverse task inputs and outputs into only two basic formats, \emph{e.g.,} continuous digits and discrete words, which can be further stacked into four modalities, \emph{i.e.,} \textit{Text} - discrete natural language tokens; \textit{Image} - normalized RGB images with continuous values from 0 to 1; \textit{Sparse Label} - 2D/3D continuous location coordinates plus discrete words as its semantic meanings; \textit{Dense Label} - per-pixel discrete natural language tokens such as segmentation maps. Compared with existing human-centric models, we avoid designing heterogeneous task heads but explore using four modality tokenizers/de-tokenizers to cover diverse human-centric tasks. These more concise formats facilitate flexibility and scalability for various tasks and datasets. Second, Hulk reformulates diverse human-centric multimodal tasks into the modality translation task and accordingly unifies different task-specific model designs into the encoder-decoder architecture which is similar to models in the machine translation task. The intuition behind this is that if the \emph{generalist human-centric perceiver} clearly perceives 2D and 3D information of human beings in an image, skeletons, or pedestrian appearance descriptions, we only need to teach it to translate between any input and output modalities. Following recent advances in machine translation~\cite{zhang2018neural, vaswani2017attention}, Hulk unifies different task-specific designs into five simple and effective components: a modality-specific tokenizer, a modality-shared encoder, a modality-specific indicator to tell the decoder which modality the inputs should be translated to, a modality-shared decoder, and a modality-specific de-tokenizer after the decoder. The tokenizers embed multimodal data into token sequences in a common manifold space. The modality-shared encoder then extracts general human-centric representations from the token sequence. The modality-specific indicator guides the modality-shared decoder to translate the human-centric representations into the tokens of the output modality. Finally, the modality-specific de-tokenizer reconstructs the tokens into the desired output modalities to complete any human-centric task.

To summarize, our contributions are three folds:
\begin{itemize}
    \item To the best of our knowledge, Hulk is the first \emph{generalist human-centric perceiver} achieving competitive results on {several common} large-scale 2D vision, 3D vision, skeleton-based and vison-language multimodal human-centric tasks\footnote{Tasks such as person re-identification and person search are not explored due to human ethical considerations.}, including pedestrian attribute recognition, pedestrian detection, 2D pose estimation, human parsing, pedestrian image caption, skeleton-based action recognition, 3D pose estimation and 3D human mesh recovery.
    \item To facilitate unified modelling, we propose to design only two types of basic blocks, \emph{i.e.,} one for predicting semantics and the other for predicting locations, to establish tokenizers and de-tokenizers of the output modalities.
    \item Trained on a massive collection of about 30 million labeled human-centric datasets of eight tasks, Hulk can directly handle a broad range of tasks without further task-specific adaptation, even pushing the performance limits of the existing state-of-the-art methods. Specifically, as shown in Fig.~\ref{fig:enter-label}, 
Hulk outperforms current leading human-centric specialist models and pretraining models by \textbf{+1.5\%} mIoU on CIHP for human paring, \textbf{+3.1\%} mA on RAPv2 for pedestrian attribute recognition, \textbf{+2.0\%} AP on AIC for 2D pose estimation, \textbf{-0.7\%} MR$^{-2}$ (Missing rate) on CrowdHuman for pedestrian detection, \textbf{-1.4} MPVPE (Mean-Per-Vertex-Position-Error) and \textbf{-2.0} MPJPE (Mean-Per-Joint-Position-Error) on 3DPW for mesh recovery and 3D pose estimation, respectively.
\end{itemize}

\section{Related Work}

\subsection{{Specialist Models on Human-centric Perception Tasks}}
{
Researchers have been developing specialist models tailored for individual tasks. To address challenges in human-centric perception tasks, four key strategies have been employed to design these specialist models: (1) \textbf{Incorporating multi-granularity feature extraction}. Tailored backbones, \emph{e.g.}, HRNet~\cite{sun2019deep} and Swin-Transformer~\cite{liu2021swin} are designed to extract representative human-centric features in 2D pose estimation~\cite{yang2023effective,sun2019deep,li2021tokenpose} and pedestrian detection tasks~\cite{zheng2022progressive}, pushing the limits on human-centric benchmarks. (2) \textbf{Leveraging the inherent relationships among body parts and attributes}. For example, GCN-based methods~\cite{stgcn,agcn,mcc,scc,unik,ctrgcn,shiftgcn} and transformer-based methods~\cite{plizzari2021skeleton,shi2020decoupled,qiu2022spatio} have achieved satisfactory results in skeleton-based action recognition tasks, as they can effectively model the positional relationships among human joints. In addition, modeling the co-occurrence of human attributes~\cite{li2022label2label,zhou2023solution} has also facilitated attribute recognition and human parsing. (3) \textbf{Designing for complex scenarios.} As the information of a pedestrian in the crowd might be hard for models to learn directly, progressive refinement~\cite{zheng2022progressive} and model distillation~\cite{yang2023effective} are proposed for better performance. (4) \textbf{Utilizing knowledge from related tasks.} For example, ED-Pose~\cite{yang2023explicit} and BUCTD~\cite{zhou2023rethinking} utilize detection information to aid 2D pose estimation, while~\cite{lip} and~\cite{zhao2022pose} use pose supervision to enhance human part segmentation. Despite the achievements, specialist models involve \textit{different} task-specific designs, leading to multiple times deployments for multiple tasks. In this paper, we propose to unify diverse human-centric tasks into modality transition tasks to learn them all at once.}

\subsection{Multimodal Foundation Models}

Multimodal foundation models have emerged as a significant research direction towards artificial general intelligence (AGI), which aims to synthesize information from various sensory inputs to create more comprehensive AI systems. 

Alignment with paired multimodal data, as the common way to develop multimodal foundation models, can be generally divided into {four} different categories: \textbf{(1) Contrastive methods}, which utilize contrastive loss to pull features with the same content close and push features with different contents apart. A famous example is CLIP~\cite{radford2021learning}, which aligns image-text pairs on websites, achieving remarkable zero-shot classification performance. VideoCLIP~\cite{xu2021videoclip} and AudioCLIP~\cite{guzhov2022audioclip} further extend this idea to alignments text with video and audio, respectively. ImageBind~\cite{girdhar2023imagebind} binds five different modalities with images, achieving strong zero-/few-shot performance on cross-modality recognition tasks.
\textbf{(2) Multimodel reconstruction methods}, which learn modality-agnostic representation through multi-modal masked autoencoding~\cite{yang2023unipad,yang2023gd}. For instance, MultiMAE~\cite{bachmann2022multimae} utilizes a modality-share encoder with modality-specific decoders to reconstruct masked RGB, depth and semantic patches, achieving desirable performance on downstream tasks in these modalities. MaskedVLM~\cite{kwon2022masked} and SMAUG~\cite{lin2023smaug} adopt masked image/text modeling to learn general representations for vision-language tasks.
{(3) \textbf{Multimodel image generation methods} proposes to deal with diverse perception tasks as generating the target images with prediction results, \emph{e.g.}, red bounding boxes for object detection~\cite{bai2024sequential}, colorful region masks for instance segmentation~\cite{wang2023images,geng2024instructdiffusion,wang2023seggpt}. Although these methods unify 2D vision tasks, it is hard to extend them in the field of 3D vision.}
\textbf{(4) Multimodal large language models (MLLM)}~\cite{yin2023lamm,chen2023shikra,li2023blip,LLaVA,Otter,zeng2023matters,PandaGPT,xu2023pointllm,xue2023ulip,gardner2023llark,zhang2023motiongpt} proposes a new approach to handling diverse perception tasks: aligning all modalities into language and formatting all perception tasks into Visual Question Answering (VQA) tasks.
Leveraging effective adaptation methods, \emph{e.g.}, LoRA~\cite{hu2021lora} and LLAMA-Adapter~\cite{zhang2023llama}, MLLM generates language answers using the given samples in different modalities, including image~\cite{li2023blip,LLaVA,Otter}, video~\cite{zeng2023matters,PandaGPT}, point cloud~\cite{xu2023pointllm,PandaGPT,zhu2023ponderv2,huang2023ponder,xue2023ulip} and audio~\cite{PandaGPT,gardner2023llark}. Despite their versatility, MLLMs have not attained state-of-the-art performance in
some perception tasks~\cite{yin2023lamm, chen2023shikra}, such as detection and mesh recovery.

Although these methods have been extensively studied in general images, they have not been explored for human-centric foundation models.
Existing works treat different tasks by different heads~\cite{wang2022ofa,lu2022unified}, leading to sub-optimal structures for human-centric tasks, where diverse inputs and outputs need a more general approach.
In comparison, we propose to unify the input and output formats of diverse human-centric tasks into four general modalities. This unification allows us to formulate diverse human-centric tasks into modality transitions, greatly simplifying the structure and ensuring knowledge sharing among different tasks. Our proposed Hulk can easily scale across a variety of human-centric tasks in 2D vision, 3D vision, vision-language, and skeleton-based tasks, thereby setting new records for these tasks.

\subsection{Human-Centric Foundation Models}

Human-centric foundation models are pivotal in real-world applications, such as social surveillance, sports analysis, and \emph{etc}. Depending on the model structures, they can be categorized into two types: \textbf{(1) Human-centric Pretraining Models}~\cite{hcmoco, yuan2023hap, solider, tang2023humanbench}, which introduce the human prior to pretrain a versatile backbone for extracting general human-centric representation. 
Specifically, HCMoCo~\cite{hcmoco} and HAP~\cite{yuan2023hap} leverage the offline extracted human keypoints as the prior knowledge to learn structure-invariant representation across diverse human poses, while SOLIDER~\cite{solider} proposes to learn representation with more semantic information using token-level semantic classification pretext task. Despite the success, these models still require additional task-specific finetuning for downstream task applications due to their lack of unified task modeling. \textbf{(2) Human-centric General Perceivers}~\cite{ci2023unihcp}, aim for a unified framework to handle diverse human-centric tasks with task-specific finetuning. A notable example, UniHCP~\cite{ci2023unihcp}, introduces a shared decoder head with a task-guided interpreter to handle five 2D vision human-centric tasks without additional task-specific finetuning. Our proposed Hulk aligns with this line and further advances the field by simplifying heterogeneous representation across tasks into only two basic formats - continuous digits and discrete words. This simplification greatly enhances the flexibility and scalability of the model and enables our proposed Hulk to simultaneously address 2D vision, 3D vision, vision-language, and skeleton-based human-centric tasks without any task-specific adaptation.

\begin{table*}[htbp]
  \centering
  \caption{Input and output modalities of eight covered human-centric tasks. Hulk treats all these tasks as modality translation. \dag In the mesh recovery task, many vertices do not have a name therefore we only predict their locations.}
  \resizebox{\linewidth}{!}{
    \begin{tabular}{lllll}
    \toprule
    Task  & Input modality & Input sample & Output modality & Output sample   \\
    \midrule
    Human Parsing & Image & $\mathbf{x}_I$  & Dense label & semantic map representing the occurrence of ``hair"   \\
    2D Pose Estimation & Image & $\mathbf{x}_I$   & Dense label & semantic map representing the occurrence of ``nose"    \\
    Pedestrian Attribute Recognition & Image & $\mathbf{x}_I$   & Text  & ``female", ``age 22-30", ...   \\
    Image Caption & Image & $\mathbf{x}_I$   & Text  & ``A man with black suit standing near the bus station"   \\
    Pedestrian Detection & Image & $\mathbf{x}_I$   & Sparse label & ``pedestrian", [0.12, 0.45, 0.21, 0.6]   \\
    3D Pose Estimation & Image & $\mathbf{x}_I$   & Sparse label & ``nose", [0.45,0.26, -0.18], ``mouth'', [0.46, 0.15, -0.19], ...  \\
    Mesh Recovery & Image & $\mathbf{x}_I$   & Sparse label & [0.61,0.42,0.17], [0.41, 0.21, 0.12], ...\dag   \\
    Skeleton-based Action & Sparse label & [[0.47,-0.21], [0.48,-0.20]...], ``left knee" & Text  & ``kicking", ``clipping'', ...   \\
    \bottomrule
    \end{tabular}%
    }
  \label{tab:task_input_output_modalities}%
  \vspace{-1em}
\end{table*}%

\section{Human-centric 2D Vision, 3D Vision, Skeleton-based and Vision-Language Tasks}  \label{sec:task_models}
Hulk, a \emph{generalist human-centric perceiver}, is designed to handle a wide range of 2D vision, 3D vision, skeleton-based and vision-language tasks in a unified way. We collect vision, skeleton-based, and multimodal datasets from 42 publicly available data sources as training datasets for our model. These datasets detailed in Sec.~\ref{sec:traning_datasets} cover 8 human-centric tasks.

The inputs and outputs of human-centric tasks can be categorized into 4 different modalities: \emph{Images} -  RGB images, which are represented as matrices with elements normalized to range between 0 and 1;
\emph{Texts} - discrete words; \emph{Sparse Labels} - the coordinates with its semantic meaning and locations, \emph{e.g.,} ``nose, [0.45, 0.26]''; \emph{Dense labels} -
per-pixel semantics represented by discrete language tokens.
In the following, we show that eight {common} human-centric tasks can be viewed as the translations among the four modalities. Examples are illustrated in Table~\ref{tab:task_input_output_modalities}.


\noindent \textbf{Human Parsing \& 2D Pose Estimation.} Given an image, human parsing and 2D pose estimation are required to generate the dense labels which are semantic segmentation maps for human parsing and keypoint heatmap for pose estimation\footnote{In this paper, we leverage the heatmap-based methods for 2D pose estimation because we experimentally verify that heatmap-based methods generally achieve better performance than keypoint-regression-based methods.}.

\noindent \textbf{Pedestrian Attribute Recognition \& Image Caption.} The RGB images are projected into texts that describe their semantic meanings. For pedestrian attribution recognition, the output is the name of pedestrian attributes, while for the image caption, the outputs are sentences of discrete words as natural language descriptions of the image.

\noindent \textbf{Pedestrian Detection \& 3D Pose Estimation \& Human Mesh Recovery.} The RGB input images are projected to sparse labels, which include continuous digits to describe their locations and discrete words to describe their semantic meaning. For pedestrian detection, the sparse labels contain the locations of the upper-left point and bottom-right points for bounding boxes and discrete words to describe the object class. For 3D pose estimation, the output is the 3D location of every keypoint and its class names, \emph{e.g.,} noses and heads. For human mesh recovery, the output is only the 3D locations of mesh vertices defined in the SMPL model. Since these vertices do not have explicit semantic meaning, we do not predict the class name of every mesh vertex.

\noindent \textbf{Skeleton-based Action Recognition.} Given sparse labels that are a temporal sequence of keypoints with their semantic meaning and locations, the skeleton-based action recognition transforms the sparse labels to the texts of the action names.

\begin{figure*} 
\centering  
\includegraphics[width=0.99\textwidth]{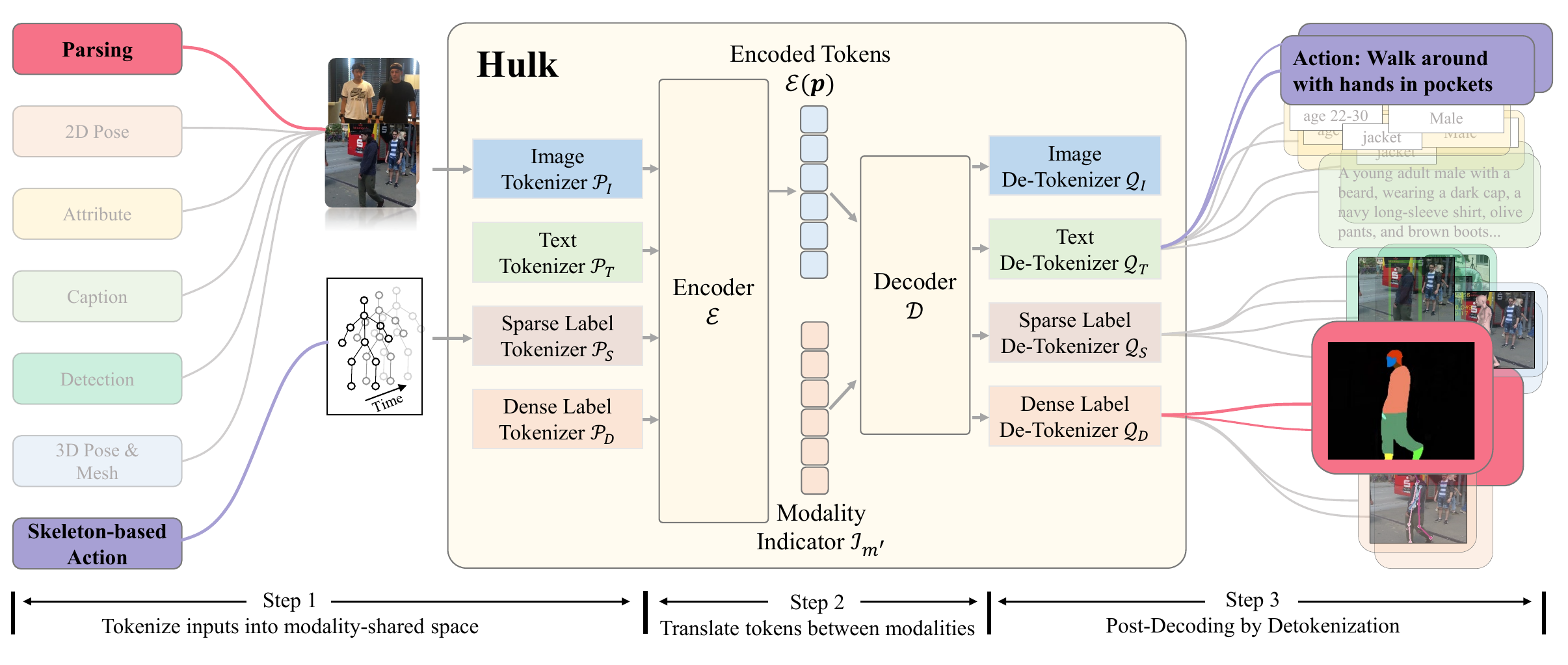}
\caption{The framework of Hulk which can handle eight human-centric tasks in a unified way. We highlight the pipeline for human parsing and skeleton-based action recognition tasks in 
{pink} and 
{purple}, respectively. Take the human parsing task for example, Hulk first tokenizes the input image using the image tokenizer $\mathcal{P}_I$ and sends tokens into the modality-shared encoder $\mathcal{E}$ to learn general human-centric representations. 
Guided by the output modality indicator $\mathcal{I}_m'$, the decoder $\mathcal{D}$ translates the encoded tokens to output tokens. Finally, the dense label de-tokenizer $\mathcal{P}$ transforms output tokens into segmentation maps. }  
\label{fig:pipeline}
\vspace{-1em}
\end{figure*}

\section{Hulk: A Generalist Human-centric \\ Knowledge Translator}
\label{sec:method}
We now introduce Hulk, a general human-centric knowledge translator, which can tackle human-centric 2D vision, 3D vision, skeleton-based, and vision-language tasks in one model without any task-specific adaption. The key challenge is to unify the heterogeneous representations of the inputs and outputs. Our Hulk has two innovations to address the problem. First, our Hulk proposes that heterogeneous inputs and outputs can be unified into four different modalities, \emph{i.e.,} texts, images, sparse labels (locations or location sequences) and dense labels (pixel-wise semantic words), which can be further simplified into two basic input and output formats of discrete semantic words and continuous values such as location coordinates. For instance, the pedestrian caption produces words that describe the appearance, age, dress-ups, \emph{etc.}; the human parsing task predicts pixel-wise semantic words of the given image; the 3D pose estimation task predicts both the location of joints and their corresponding semantic words such as the nose and the right winkle.  Second, our Hulk proposes a modality translation task where {several common} 2D vision, 3D vision, skeleton-based, and vision-language human-centric tasks can be viewed as its special case. The intuition of the modality translation task is that if a model well understands all human-centric knowledge, it only needs to translate from the input modality to the output modality to handle human-centric tasks, \emph{e.g.,} human mesh recovery task translates the image into sparse labels that are 3D location of mesh vertices, and skeleton-based action recognition task translates the human keypoint temporal sequences into texts that describe the action semantic names.

\subsection{Model Architecture}
To realize the general modality translation for human-centric perceptions, our proposed Hulk, \emph{i.e., \underline{H}uman \underline{u}niversa\underline{l} \underline{k}nowledge translator}, consists of three simple yet effective components: a modality-specific tokenizer and de-tokenizer (Sec.~\ref{sec:tokenizer_and_detokenizer}) composed of two different heads to predict semantics and locations, a modality-shared encoder and decoder (Sec.~\ref{sec:token_translation}) and a modality-specific indicator (Sec.~\ref{sec:token_translation}) to guide the modality-shared decoder to translate the features extracted by the encoder to the output modality. Finally, we leverage the semantic contrastive loss to supervise the classification of semantic words and the digit regression loss to supervise the regression of locations. As shown in Fig.~\ref{fig:pipeline}, the pipeline of our proposed Hulk of translating from the input modality to the output modality can be described as follows:

\subsubsection{Step1 -- Tokenize modality-specific inputs into a modality-shared manifold space} Given input data $\mathbf{x}_{m}$ from one of the four modalities, \emph{i.e.,} images~$I$, texts~$T$, sparse labels~$S$ and dense labels~$D$, we leverage the modality-specific tokenizer to project them into tokens $\mathbf{p}$ in the modality-shared manifold, \emph{i.e.,} $\mathbf{p}\!=\!\mathcal{P}_m(\mathbf{x}_m)$, where $\mathcal{P}_m$ is the tokenizer of modality $m$ before the transformer encoder $\mathcal{E}$ and $m\!\in\!\{I, T, S, D\}$ denotes the modality of input $\mathbf{x}_m$.

\subsubsection{Step2 -- Translate the input tokens to output modalities
} Given the input token $\mathbf{p}$ and the output modality indicator $\mathcal{I}_m'$, our process involves two primary steps. First, the encoder $\mathcal{E}$ is employed to extract human-centric representations from $\mathbf{p}$. These representations are then translated into the designated output modality. This translation is facilitated by the modality-shared decoder, which operates in conjunction with the output modality indicator. Formally, the output tokens $\mathbf{q}$ are generated as $\mathbf{q} = \mathcal{D}(\mathcal{E}(\mathbf{p}), \mathcal{I}_{m'})$. In this formulation, $\mathcal{D}$ denotes the transformer decoder, and $m'\in\{I, T, S, D\}$ specifies the output modality.

\subsubsection{Step3 -- Post-Decoding Optimization by Detokenization, Semantic Classification, and Location Regression
} 
Given the translated tokens after the decoder, we generate the output modalities by the modality-specific de-tokenizer, \emph{i.e.,} $\hat{\mathbf{y}}_s = 
\mathcal{Q}_s(\mathbf{q})$. Finally, the parameters in the tokenizers $\mathcal{P}\in\{\mathcal{P}_I, \mathcal{P}_T, \mathcal{P}_S, \mathcal{P}_D\}$, the encoder $\mathcal{E}$, the decoder $\mathcal{D}$, the de-tokenizers $\mathcal{Q}\in\{\mathcal{Q}_I, \mathcal{Q}_T, \mathcal{Q}_S, \mathcal{Q}_D\}$ and the modality indicators $\mathcal{I}_{m'}\in\{\mathcal{I}_I, \mathcal{I}_T, \mathcal{I}_S, \mathcal{I}_D\}$ are optimized by the semantic contrastive loss for classifying semantic words and the digital regression loss for regressing locations.

\subsection{Tokenizer and De-Tokenizer} \label{sec:tokenizer_and_detokenizer}

\begin{figure}[t]
    \centering
    \includegraphics[width=\linewidth]{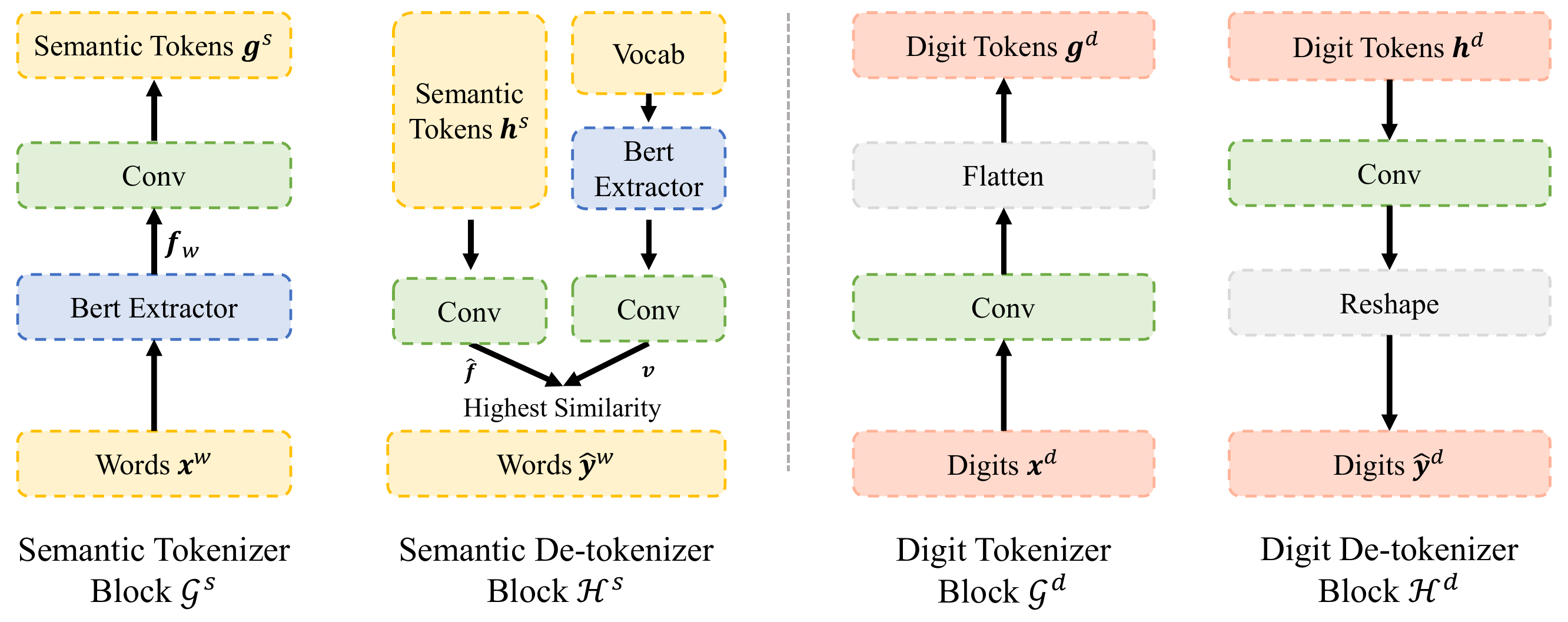}
    \caption{Two tokenizer and de-tokenizer blocks. Semantic (de-)tokenizer classifies the semantics text and digit (de-)tokenizer regresses the locations.}
    \label{fig:tokenizer_detokenizer}
    \vspace{-0.85em}
\end{figure}

One of the most critical challenges to developing the \emph{generalist human-centric perceiver} is to tackle the heterogeneity of the inputs and outputs. 
Unlike previous methods that utilized modality-specific projectors and task-specific heads for generating diverse inputs and outputs, our approach, Hulk, simplifies this process. Hulk employs a streamlined interface involving the tokenization of task inputs and the de-tokenization of representations into task outputs. This is achieved using just two types of blocks: a tokenizer and a de-tokenizer, as depicted in Fig.~\ref{fig:tokenizer_detokenizer}. Specifically, one block is designed to classify semantic texts, and the other to regress locations. This design is based on the fundamental perception task objectives: to recognize semantic meanings and find their locations.
We further showcase these two types of blocks can be stacked to construct the tokenizers and de-tokenizers of four modalities in {common} human-centric tasks. 

\subsubsection{Tokenizer and De-tokenizer Blocks}
We design two pairs of tokenizers and de-tokenizers for classifying semantic words and regressing the digital locations, respectively. 

\noindent \textbf{Semantic Tokenizer and De-tokenizer Block.} The semantic tokenizer block $\mathcal{G}^s$ aims at tokenizing words in a modality-shared manifold, while the semantic de-tokenizer block $\mathcal{H}^s$ aims at decoding the tokens to the words. The semantic tokenizer and de-tokenizer have dual designs. Specifically, given several words $\mathbf{x}_w=\{\mathbf{x}_1^w, \mathbf{x}_2^w, ..., \mathbf{x}_N^w\}$ representing different semantics, we extract the features\footnote{
According to the vocabulary of BERT, different words will be split into different numbers of tokens, leading to various lengths of extracted BERT features in different tasks. However, without loss of generality, we use the pooled BERT feature for all tasks.
} 
$\mathbf{f}_w=\{\mathbf{f}^w_1, \mathbf{f}^w_2, ..., \mathbf{f}^w_N\}$ by the off-shore BERT~\cite{devlin2018bert} model, where $N$ is the number of words. 
In the skeleton-based action recognition task, semantics of joints are also introduced to enhance Hulk's understanding of the movement of joints.
To emphasize the semantic correlation among nearing words, we use a convolutional network to generate the text tokens $\mathbf{g^s}$, \emph{i.e.,}
\begin{equation} \label{eq:semantic_tokenizer_block}
    \mathbf{g}^s = \mathcal{G}^s (\mathbf{x}_1^w, \mathbf{x}_2^w, ..., \mathbf{x}_N^w) = \text{Conv}(\mathbf{f}^w_1, \mathbf{f}^w_2, ..., \mathbf{f}^w_N),
\end{equation}
where $\mathbf{f}^i = \text{BERT}(\mathbf{x}_i^w)$ is the BERT feature of the word $\mathbf{x}_i^w$.

Accordingly, the semantic de-tokenizer block $\mathcal{H}^s$ decodes the tokens into words. Given the words in the vocabulary, we extract the features of all words $\mathbf{v}=\{\mathbf{v_1}, \mathbf{v_2}, ..., \mathbf{v}_{C}\}$ by BERT and leverage the convolutional network to project the output semantic tokens $\mathbf{h}^s$ to the BERT feature space $\hat{\mathbf{f}}$,
\begin{equation} \label{eq:semantic_detokenizer_output_feature}
    \hat{\mathbf{f}} = \text{Conv}(\mathbf{h}^s).
\end{equation}
Then we select the word $\hat{\mathbf{y}}^w$ whose BERT feature is the most similar to the token feature,  \emph{i.e.,} 
\begin{equation}\label{eq:semantic_detokenizer_block}
    \hat{\mathbf{y}}^w = \mathcal{H}^s(\mathbf{h}^s,\mathbf{v}) = \{y_j = \text{argmax}_{k\in [1, C]} \mathbf{v}^\top \hat{\mathbf{f_{j}}} \}, j\in [1,N'],
\end{equation}
where $C$ and $N'$ denote the number of semantic tokens in $\mathbf{v}$ and output tokens, respectively.

\noindent \textbf{Digit Tokenizer and De-tokenizer Block.} The digit tokenizer $\mathcal{G}^d$ and de-tokenizer block $\mathcal{H}^d$ are designed for mutual conversion between continuous digits representing locations and tokens. Given the digit sequence $\mathbf{x}^{d}=[\mathbf{x}_1^{d}, \mathbf{x}_2^{d}, ..., \mathbf{x}_N^{d}]$, where $N$ is the length of digit sequence, we design the convolutional network to extract the information in the digits by transforming the digits to modality-shared tokens, and then flatten it into the token sequence, \emph{i.e.,}
\begin{equation} \label{eq:digit_tokenizer_block}
    \mathbf{g}^d = \mathcal{G}^d([\mathbf{x}_1^{d}, \mathbf{x}_2^{d}, ..., \mathbf{x}_N^{d}]) = \text{Flatten}(\text{Conv}([\mathbf{x}_1^{d}, \mathbf{x}_2^{d}, ..., \mathbf{x}_N^{d}])).
\end{equation}

Accordingly, to decode the digit tokens $\mathbf{h}^d$ into digits indicating locations, we design the convolutional network to generate digits and then reshape the generated digits to the same shape as task outputs $\hat{\mathbf{y}}^{d}$, \emph{i.e.,}
\begin{equation} \label{eq:digit_detokenizer_block}
    \hat{\mathbf{y}}^{d}=[\hat{\mathbf{y}}_1^{d}, \hat{\mathbf{y}}_2^{d}, ..., \hat{\mathbf{y}}_{N^{'}}^{d}] = \mathcal{H}^d(\mathbf{h}^d) = \text{Reshape}(\text{Conv}(\mathbf{h}^d)),
\end{equation}
where $N'$ is the length of the output sequence.

\subsubsection{Stacking Blocks to modality-specific Tokenizers and De-tokenizers} ~\label{sec:stack_tokenizer_detokenizer}
As detailed in Sec.~\ref{sec:task_models}, we categorize the input and output formats of human-centric tasks into 4 different modalities: Text--sentences containing words; Image--RGB images; Sparse labels--a number of location coordinates in the image; Dense labels--per-pixel sentences to describe semantics.

\noindent \textbf{Image.} Most human-centric perception tasks rely on the image modality as input. An image $\mathbf{x}_I \in \mathbb{R}^{C \times H \times W}$, where $C$ is the number of channels, $H$ is the height of the image and $W$ is the width of the image, is a matrix with continuous digits between 0 and 1 after normalization. Therefore, we directly leverage the digit tokenizer block $\mathcal{G}^d$ as the image tokenizer $\mathcal{P}_{I}$, \emph{i.e.,}
\begin{equation}
    \mathbf{p} = \mathcal{P}_I(\mathbf{x}_I) = \mathcal{G}^d(\mathbf{x}_I),
\end{equation}
where $\mathbf{p}$ is the input tokens in the modality-shared manifold, and $\mathcal{G}^d$ is the digit tokenizer block defined in Eq.~\ref{eq:digit_tokenizer_block}. Since Hulk only focuses on human-centric perception tasks, we do not design an image de-tokenizer in this paper and leave it for future works.

\noindent \textbf{Text.} The text modality is an important output modality for human-centric perception tasks, including pedestrian caption, skeleton-based action recognition, \emph{etc.} In practice, we directly leverage the semantic tokenizer block as the text tokenizer and the semantic de-tokenizer block as the text de-tokenizer. Given a text inputs of $N$ words, $\mathbf{x}_T =(\mathbf{x}^w_1, \mathbf{x}^w_2, ..., \mathbf{x}^w_N)$, with the text tokenizer $\mathcal{P}_{T}$, the input tokens $\mathbf{p}$ can be computed by
\begin{equation}
    \mathbf{p} = \mathcal{P}_T(\mathbf{x}_T) = \mathcal{G}^s (\mathbf{x}_1^w, \mathbf{x}_2^w, ..., \mathbf{x}_N^w),
\end{equation}
where $\mathcal{G}^s$ is the semantic tokenizer block defined in Eq.~\ref{eq:semantic_tokenizer_block}.

For attribute or action classification tasks, to obtain the prediction words using text de-tokenizer $\mathcal{Q}_{T}$ and the output token $\mathbf{q}$, 
\begin{equation}
    \hat{\mathbf{y}}_T = \mathcal{Q}_T(\mathbf{q})=\mathcal{H}^{s}(\mathbf{q},\mathbf{v}),
\end{equation}
where $\mathcal{H}^s$ is the semantic de-tokenizer defined in Eq.~\ref{eq:semantic_detokenizer_block}, $\mathbf{v}$ denotes the features of attribute or action classes. 

For the pedestrian caption task, to auto-regressively decode $N$ words in task outputs $\hat{\mathbf{y}}_T=(\hat{\mathbf{y}}^w_1, \hat{\mathbf{y}}^w_2, ..., \hat{\mathbf{y}}^w_N)$, we define the text de-tokenizer $\mathcal{Q}_{T}$ as 
\begin{equation}
\begin{aligned}
    \hat{\mathbf{y}}_T &= \mathcal{Q}_T(\mathbf{q}) = [\mathcal{H}^s(\mathbf{q}_1, \mathbf{v}), \mathcal{H}^s(\mathbf{q}_2, \mathbf{v}), ..., \mathcal{H}^s(\mathbf{q}_N, \mathbf{v})],
    \end{aligned}
\end{equation}
where $\mathcal{H}^{s}$ is the semantic de-tokenizer block (Eq.~\ref{eq:semantic_detokenizer_block}), $\mathbf{q}_i$ denotes the $i$-th output tokens in auto-regressive manner, $\mathbf{v}$ denotes the features of BERT vocabulary.

\noindent \textbf{Sparse Label.} The sparse labels usually refer to points with semantic meanings. For example, the task outputs of the pose estimation and task inputs of skeleton-based action recognition are the point coordinates and their corresponding semantic meanings, \emph{e.g.}, ``nose". Therefore, the sparse label tokenizer and de-tokenizer combine semantic and digit blocks. We leverage the semantic block to encode the semantics of points and the digit blocks to encode the point locations. Speicifically, given the sparse labels $\mathbf{x}_S=(\mathbf{x}_S^s; \mathbf{x}_S^d)$, where $\mathbf{x}_S^s$ denotes the semantics of the sparse labels and $\mathbf{x}_S^d$ denotes the digital locations of the sparse labels. Mathematically, the input tokens $\mathbf{p}$ is computed using the sparse label tokenizer $\mathcal{P}_S$ by 
\begin{equation}
    \mathbf{p}=\mathcal{P}_S (\mathbf{x}_S) = \mathcal{G}^s(\mathbf{x}^s_S) + \mathcal{G}^d(\mathbf{x}^d_S),
\end{equation}
where $\mathcal{G}^s$ and $\mathcal{G}^d$ are the semantic tokenzier block (Eq.~\ref{eq:semantic_tokenizer_block}) and the digit tokenizer block (Eq.~\ref{eq:digit_tokenizer_block}), respectively. Given the output tokens $\mathbf{q}$ after the decoder, the sparse label de-tokenizer $\mathcal{Q}_s$ decomposes it into semantics and digits of sparse label outputs $\hat{\mathbf{y}}_S = (\hat{\mathbf{y}}_S^s, \hat{\mathbf{y}}_S^d)$, where $\hat{\mathbf{y}}_S^s$ and $\hat{\mathbf{y}}_S^s$ are the semantics and digital locations of the output sparse label $\hat{\mathbf{y}}_S$. Specifically, 
\begin{equation}
    (\hat{\mathbf{y}}_S^s, \hat{\mathbf{y}}_S^d) = \mathcal{Q}_S(\mathbf{q}), 
    \hat{\mathbf{y}}_S^s = \mathcal{H}^s(\mathbf{q}), \hat{\mathbf{y}}_S^d = \mathcal{H}^d(\mathbf{q}),
\end{equation}
where $\mathcal{H}^s$ and $\mathcal{H}^d$ are the semantic de-tokenzier block (Eq.~\ref{eq:semantic_detokenizer_block}) and the digit de-tokenizer block (Eq.~\ref{eq:digit_detokenizer_block}).

\noindent \textbf{Dense Label.} The dense labels are the semantic words for every pixel of the image. For the dense label tokenizer $\mathcal{P}_{D}$, given a dense label $\mathbf{x}_D \in \mathbb{R}^{C \times H \times W}$, where $C$ is the number of classes consisting of different words, we leverage the semantic tokenizer block $\mathcal{G}^{s}$ to transform it into input tokens $\mathbf{p}$, \emph{i.e.,}
\begin{equation}
    \mathbf{p}=\mathcal{P}_D(\mathbf{x}_D) = \mathcal{G}^s(\mathbf{x}_D).
\end{equation}

Different from that in the text de-tokenizer, the decoder output $\mathbf{q}$ often embeds the information of an image patch instead of a pixel. Therefore, 
the similarity is computed between the features of words in BERT vocabulary $\mathbf{v}$ and the feature of upsampled output tokens $\hat{\mathbf{f}}$ (Eq.~\ref{eq:semantic_detokenizer_output_feature}).
The dense label predictions $\hat{\mathbf{y}}_D$ are then computed by 
\begin{equation}
\begin{aligned}
     \hat{\mathbf{y}}_D &= \mathcal{Q}_D(\mathbf{q})  \\
     &= \{\hat{\mathbf{y}}_j  = \text{argmax}_{k\in [1, C]} \mathbf{v}^\top\text{Upsample}( \hat{\mathbf{f_{j}}}) \}, j\in [1,N'],
 \end{aligned}
\end{equation}
where $C$ denotes the different semantic classes in $\mathbf{v}$, $N'$ is the number of output tokens, the dense label de-tokenizer $\mathcal{Q}_D$ has the additional ``Upsample'' operator than $\mathcal{H}^s$ defined Eq.~\ref{eq:digit_detokenizer_block}.

\begin{figure*}[t!]
    \centering
    \includegraphics[width=.95\linewidth]{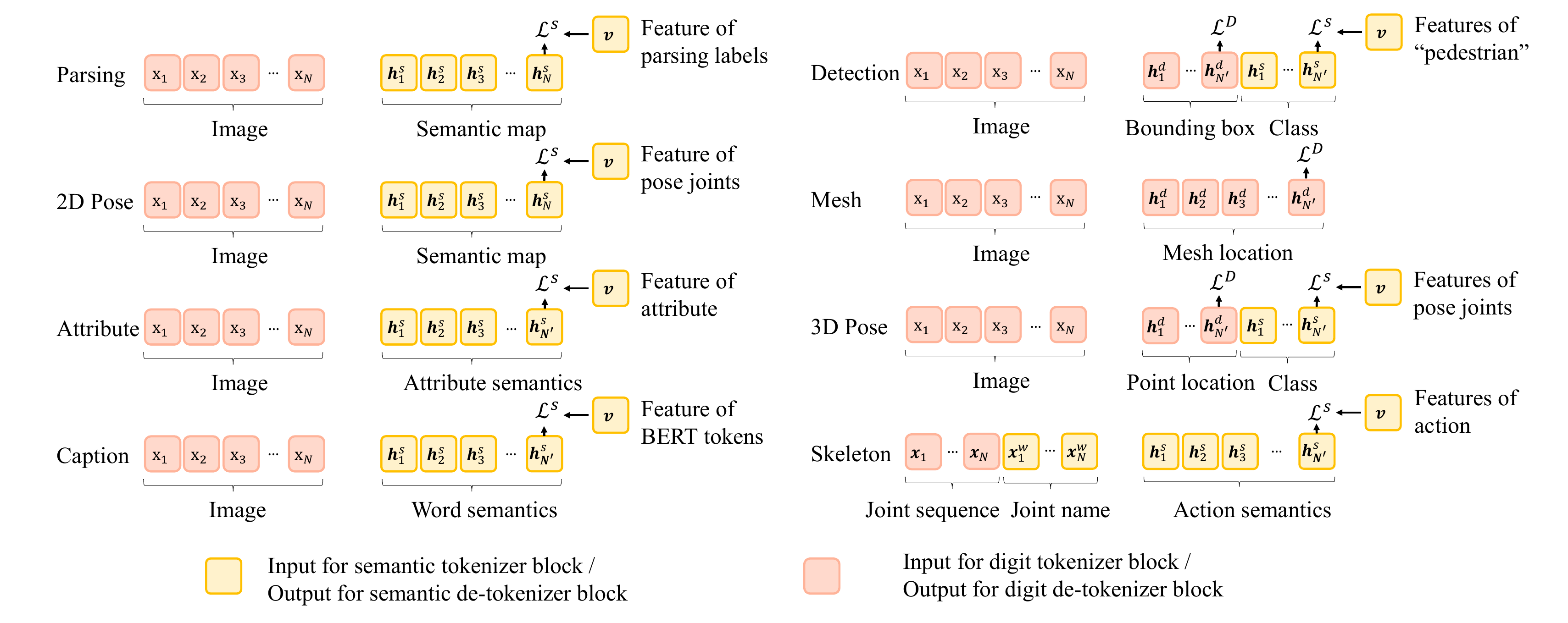}
    \caption{Input, output tokens and objective functions for eight human-centric tasks. The objective function supervises all the output tokens while only one token's loss is depicted here due to the limited space.}
    \label{fig:task_loss}
    \vspace{-1em}
\end{figure*}


\subsection{Token Translation} ~\label{sec:token_translation}
Given the modality-specific tokenizer and the modality-specific de-tokenizer, the encoder and the decoder translate the input tokens to output tokens with the guidance of the modality-specific indicator, \emph{i.e.}, four learnable tokens for four modalities. The design for token translation is similar to the architecture of ``Transformer''. Specifically, give the tokens $\mathbf{p}$ of the input modality, the encoder extracts the general human-centric representations, and then the decoder translates the general representation to the output tokens $\mathbf{q}$, which is guided by the modality-specific indicator $\mathcal{I}_{m'}$. Mathematically, 
this process can be defined as 
\begin{equation}
    \mathbf{q} = \mathcal{D}(\mathcal{E}(\mathbf{p}), \mathcal{I}_{m'}),
\end{equation}
where $\mathcal{E}$ is the transformer encoder, $\mathcal{D}$ is the transformer decoder, and $\mathcal{I}_{m'} \in \mathbb{R}^{N'\times d_{out}}$ is a learnable token repeated $N'$ times for modality $m'$, where $d_{out}$ is the dimension of decoder features, $N'$ is the number of output tokens.

\subsection{Objective Functions} \label{sec:objective}
As illustrated in Sec.~\ref{sec:stack_tokenizer_detokenizer}, the modality-specific de-tokenizer will transform output tokens to the semantic part and the digital location part regarding corresponding human-centric tasks. Therefore, the diverse losses proposed in different human-centric tasks can be unified into two, \emph{i.e.,} the contrastive loss to align the predicted semantics and the ground-truth labels, the digit regression loss to supervise the regressed point locations using the ground truth location.

\subsubsection{Semantic Contrastive Loss}
The semantic contrastive loss aims at aligning the predicted semantic words and the semantic labels of the text, the sparse label, and the dense label modality. Mathematically, given the conv-transformed features $\hat{\mathbf{f}}$ of predicted semantic tokens $\mathbf{h}^s$ , the BERT feature $\mathbf{v}$ of the ground-truth semantic label containing the BERT features of all label candidates $(\mathbf{v}_1, \mathbf{v}_2, ..., \mathbf{v}_C)$, the semantic contrastive loss is defined as
\begin{equation}
    \mathcal{L}^s = \frac{\mathbf{v}_k^\top \hat{\mathbf{f}}}{\mathbf{v}_1^\top \hat{\mathbf{f}} + \mathbf{v}_2^\top \hat{\mathbf{f}} + ... + \mathbf{v}_C^\top \hat{\mathbf{f}}},
\end{equation}
where $k$ denotes the index of the selected semantic label in $\mathbf{v}$ and the computation of $\hat{\mathbf{f}}$ is defined in Eq.~\ref{eq:semantic_detokenizer_output_feature}. Here, $(\mathbf{v}_1, \mathbf{v}_2, ..., \mathbf{v}_C)$ are the BERT features that differ in two types of tasks: classification tasks and caption tasks. For the classification task, \emph{i.e.,} skeleton-based action recognition, attribute recognition, the candidate features are the BERT features of all class names in the defined task. For the caption task, \emph{i.e.,} the pedestrian caption, the candidate features are the BERT features of all tokens in the BERT vocabulary.

\subsubsection{Digit Regression Loss}
The digit regression loss aims at predicting 3D coordinate values (3d pose estimation, human mesh recovery), or coordinate sequence values ($x_1y_1x_2y_2$ coordinates in pedestrian detection). Specifically, given the coordinates of the sparse labels, the digit regression is defined as 
\begin{equation}
    \mathcal{L}^d = Dis(\hat{\mathbf{y}}_S^d, \mathbf{y}_S^d),
\end{equation}
where $Dis(\hat{\mathbf{y}}_S^d, \mathbf{y}_S^d)$ measures the distance between predicted $\hat{\mathbf{y}}_S^d$ and ground truth $\mathbf{y}_S^d$, which is different in different tasks, \emph{e.g.,} L1 loss and GIoU loss in the pedestrian detection.

\subsubsection{Overall Objective function}
In general multi-dataset learning of Hulk with $K$ datasets, given the prediction $\hat{\mathbf{y}}_i$ with ground truth $\mathbf{y}_i, i=1,2,...,K$, in $i$-th dataset, the overall objective function is defined as 
\begin{equation}
    \mathcal{L} = \sum^{K}_{i=1}{\lambda_i\mathcal{L}_i(\hat{\mathbf{y}}_i, \mathbf{y}_i)}, \mathcal{L}_i\in\{\mathcal{L}^s,\mathcal{L}^d\},
\end{equation}
where $\mathcal{L}_i$ is the dataset-specific loss with corresponding weight $\lambda_i$. For each task, we illustrate the input, output and the objective function in Fig.~\ref{fig:task_loss}. Detailed losses for every human-centric task are elaborated in the supplemental material.

\section{Experiment}
\subsection{Datasets and Evaluation Metric}
\subsubsection{Training Datasets} \label{sec:traning_datasets}

To establish a human-centric universal knowledge translator, we train the proposed Hulk at scale on a diverse and numerous collection of human-centric datasets. There are two major types of datasets: (1) datasets that leverage professional capture equipment~\cite{3dpw,h36m_pami} or proficient human annotators~\cite{coco,aic,cihp,lip,rapv2,pa,shao2018crowdhuman}. These datasets generally provide supervision in high quality but are less scalable due to high costs. (2) pseudo-annotated datasets~\cite{zuo2023plip,fu2021unsupervised} whose annotations are generated by state-of-the-art methods~\cite{li2020self,radford2019language}. Although the annotations of these data are not of high quality, they can greatly enrich the scenarios of the training data, of which the effectiveness has been proven by existing work~\cite{kirillov2023segment, zuo2023plip}. Generally, 42 publically available datasets are gathered to form the training set for Hulk, containing 30,187,836 training samples and covering eight different human-centric tasks -- human parsing, 2D pose estimation, attribute recognition, pedestrian detection, skeleton-based action recognition, image caption, 3D pose estimation, and mesh recovery. Table~\ref{tab:training datasets} presents the number of datasets and samples in each task. 
Following the de-duplication practice in UniHCP~\cite{ci2023unihcp}, we remove the samples that could appear in the evaluation datasets. More detailed dataset setups can be found in the supplemental material.

\begin{table}[t]
  \centering
  \caption{Statistics of Training Datasets.}
    \begin{tabular}{p{8.69em}lc}
    \toprule
    \multicolumn{1}{l}{Task type} & \multicolumn{1}{l}{Dataset} & \multicolumn{1}{l}{Number of samples} \\
    \midrule
    \multirow{2}[2]{*}{Human parsing} & \multicolumn{1}{l}{Human3.6M~\cite{h36m_pami}} & \multirow{4}[2]{*}{1,419,910} \\
    \multicolumn{1}{l}{} & \multicolumn{1}{l}{LIP~\cite{lip}} &  \\
    \multirow{1}[0]{*}{(7 datasets)} & \multicolumn{1}{l}{CIHP~\cite{cihp}} &  \\
    \multicolumn{1}{l}{} & \multicolumn{1}{l}{…} &  \\
    \midrule
    \multirow{2}[2]{*}{2D pose estimation} & \multicolumn{1}{l}{COCO~\cite{coco}} & \multirow{4}[2]{*}{3,192,532} \\
    \multicolumn{1}{l}{} & \multicolumn{1}{l}{AIC~\cite{aic}} &  \\
    \multirow{1}[0]{*}{(8 datasets)} & \multicolumn{1}{l}{Posetrack~\cite{andriluka2018posetrack}} &  \\
    \multicolumn{1}{l}{} & \multicolumn{1}{l}{…} &  \\
    \midrule
    \multirow{2}[2]{*}{Attribute recognition} & \multicolumn{1}{l}{RAPv2~\cite{rapv2}} & \multirow{4}[2]{*}{10,911,029} \\
    \multicolumn{1}{l}{} & \multicolumn{1}{l}{PA-100k~\cite{pa}} &  \\
    \multirow{1}[0]{*}{(6 datasets)} & \multicolumn{1}{l}{parse27k~\cite{PARSE27k}} &  \\
    \multicolumn{1}{l}{} & \multicolumn{1}{l}{…} &  \\
    \midrule
    \multirow{2}[2]{*}{Pedestrian detection} & \multicolumn{1}{l}{CrowdHuman~\cite{shao2018crowdhuman}} & \multirow{4}[2]{*}{170,687} \\
    \multicolumn{1}{l}{} & \multicolumn{1}{l}{EuroCity~\cite{braun2018eurocity}} &  \\
    \multirow{1}[0]{*}{ptasets)} & \multicolumn{1}{l}{WiderPerson~\cite{zhang2019widerperson}} &  \\
    \multicolumn{1}{l}{} & \multicolumn{1}{l}{…} &  \\
    \midrule
    \multirow{2}[1]{*}{Skeleton-based action} & \multicolumn{1}{l}{NTU60~\cite{shahroudy2016ntu}} & \multirow{4}[2]{*}{391,685} \\
    \multicolumn{1}{l}{} & \multicolumn{1}{l}{GYM~\cite{shao2020finegym}} &  \\
    \multirow{1}[0]{*}{(6 datasets)} & \multicolumn{1}{l}{Diving48~\cite{diving}} &  \\
    \multicolumn{1}{l}{} & \multicolumn{1}{l}{…} &  \\
    \midrule
    Image caption & \multicolumn{1}{l}{CUHK-PEDES~\cite{cuhk}} & \multirow{2}[1]{*}{12,206,283} \\
    \multicolumn{1}{l}{(2 datasets)} & \multicolumn{1}{l}{SYNTH-PEDES~\cite{zuo2023plip}} &  \\
    \midrule
    3D pose estimation & \multicolumn{1}{l}{Human3.6M~\cite{h36m_pami}} & \multirow{4}[4]{*}{1,895,710} \\
    \multicolumn{1}{l}{(7 datasets)} & \multicolumn{1}{l}{Muco~\cite{muco}} &  \\
\cmidrule{1-1}    Mesh recovery & \multicolumn{1}{l}{GTA~\cite{gta}} &  \\
    \multicolumn{1}{l}{(7 datasets)} & \multicolumn{1}{l}{…} &  \\
    \midrule
    Total & 42    & 30,187,836 $\approx$\textbf{30M} \\
    \bottomrule
    \end{tabular}%
  \label{tab:training datasets}%
  \vspace{-1em}
\end{table}%

\subsubsection{Evaluation Datasets}
To evaluate the performance of Hulk across diverse human-centric tasks, following UniHCP~\cite{ci2023unihcp}, we select multiple core datasets to form a comprehensive benchmark. Specifically, CrowdHuman~\cite{shao2018crowdhuman} is selected for evaluation on pedestrian detection, COCO~\cite{coco} and AIC~\cite{aic} are selected for 2D pose estimation, Human3.6M~\cite{h36m_pami}, LIP~\cite{lip} and CIHP~\cite{cihp} are selected for human parsing, PA-100K~\cite{pa} and RAPv2~\cite{rapv2} are selected for attribute recognition.

For newly included tasks, we use the most widely-used benchmarks, \emph{i.e.}, 3DPW~\cite{3dpw} and Human3.6M~\cite{h36m_pami} for 3D pose estimation and mesh recovery, NTU60-XSUB~\cite{shahroudy2016ntu} for skeleton-based action recognition, CUHK-PEDES~\cite{cuhk} for image caption on pedestrians.

\subsection{Implementation Details}
Following UniHCP~\cite{ci2023unihcp} and PATH~\cite{tang2023humanbench}, we use the standard ViT-Base~\cite{dosovitskiy2020image} as the default backbone and initialize it with unsupervised MAE pretraining~\cite{he2022masked}. Unless otherwise specified, the input resolution is consistent within each task, \emph{i.e.}, 256$\times$192 resolution for 2D pose estimation and attribute recognition, 480$\times$480 resolution for human parsing, 224$\times$224 resolution for 3D pose estimation and mesh recovery, 17 joints$\times$175 frames for skeleton-based action recognition, 384$\times$384 resolution for image caption, and a maximum height/width of 1120 for pedestrian detection. Due to the high GPU memory demand of detection tasks, we resize the maximum height/width of the input images from the commonly used 1333 to 1120,  which has a certain impact on the performance of detection tasks but does not affect the superiority of Hulk over other methods.

Considering the computational and communication efficiency during training, we adopt the practices in UniHCP to run one specific task on each GPU with gradient checkpointing~\cite{rojas2020study}. We use the Adafactor optimizer with $\beta_1=0.9$, $\beta_2$ clipped at 0.999 and disable the parameter scaling. We adopt a 60000-iteration schedule with a linear warm-up for 2000 iterations. We set the base learning rate to $10^{-3}$ with a cosine decay scheduler. For ViT-B backbone, we use a drop-path-rate of 0.2, a layer-wise learning rate decay of 0.75, and set weight decay to 0.05. For ViT-L, we use a drop-path-rate of 0.5, a layer-wise learning rate decay of 0.85, and a weight decay of 0.1, following the recommendation setup in ViTDET~\cite{li2022exploring}. To better utilize the knowledge in every dataset, we do not restrict all datasets to take the same epoch. {As different task loss scales vary, to prevent model bias to a certain task, we monitor gradients on task-shared parameters, ensure similar gradients for all tasks, and ablate them for better performance. 
Joint training setups are detailed in Sec. VIII of the supplemental material.} The whole training takes 70, 144 hours in total on 80 NVIDIA A100 GPUS for ViT-B, ViT-L, respectively.

\subsection{Comparison with States-of-the-art Methods}
Extensive evaluations are conducted to demonstrate the capabilities of our Hulk across diverse human-centric tasks. Table~\ref{tab:detection}-\ref{tab:caption} show the experimental results of Hulk on eight different human-centric tasks, i.e., pedestrian detection, 2D pose estimation, human parsing, pedestrian attribute recognition, 3D human pose and mesh recovery, skeleton-based action recognition, and pedestrian image caption. 

To facilitate a more detailed analysis, we categorize the comparative methods into three distince classes: (1) Specialist models, referring to models custom-tailored specifically for the target task; (2) Pretraining models, involving models pre-trained on datasets distinct from the target task, followed by fine-tuning on the target task; and (3) Generalist models, 
which are designed as unified frameworks capable of handling multiple tasks simultaneously.

Following UniHCP, we report two kinds of evaluation results of Hulk: (1) \textbf{direct evaluation}, where the pre-trained models are directly used for evaluation on the target dataset, and (2) \textbf{finetuning}, where the pretrained Hulk is first finetuned with the train split of the target dataset and then evaluated.

\subsubsection{2D Vision Tasks}
As observed, our proposed Hulk demonstrates promising performance on four 2D vision tasks, i.e., pedestrian detection, 2D pose estimation, human parsing, and pedestrian attribute recognition. 

\begin{table}[t]
  \centering
  \caption{Pedestrian detection evaluation on CrowdHuman~\cite{shao2018crowdhuman} with mAP, MR$^{-2}$ and JI. Following UniHCP~\cite{ci2023unihcp}, we report the direct eval and fine-tune (FT) results. \dag indicates using a smaller input resolution with a maximum height/width of 1120.}
  \label{tab:detection}
  \resizebox{\linewidth}{!}{
    \begin{tabular}{cllccc}
    \toprule
    \multicolumn{2}{l}{Method} & Backbone & mAP & MR$^{-2}$$\downarrow$ & JI \\  
    \midrule
    \multirow{8}{*}{Specialist} 
      & DETR~\cite{carion2020end}  & ResNet-50 & 75.9 & 73.2 & 74.4 \\
      & CrowdDet~\cite{chu2020detection}  & ResNet-50 & 90.7 & 41.4 & 82.3 \\
      & V2F-Net~\cite{shang2021v2f}  & ResNet-50 & 91.0 & 42.3 & - \\
      & PEDR~\cite{lin2020detr}  & ResNet-50 & 91.6 & 43.7 & 83.3 \\
      & DDETR~\cite{zhu2020deformable} & ResNet-50 & 91.5 & 43.7 & 83.1 \\
      & Sparse-RCNN~\cite{sun2021sparse} & ResNet-50 & 91.3 & 44.8 & 81.3 \\
      & Iter-DDETR~\cite{zheng2022progressive} & ResNet-50 & 92.1 & 41.5 & 84.0 \\
      & Iter-Sparse-RCNN~\cite{zheng2022progressive} & ResNet-50 & \textbf{92.5} & 42.6 & 83.3 \\
    \cmidrule{2-6}  
      & Iter-DDETR~\cite{zheng2022progressive} & Swin-L & \textbf{94.1} & 37.7 & \textbf{87.1} \\
    \midrule
    \multirow{2}{*}{Pretraining} 
      & PATH~\cite{tang2023humanbench} & ViT-B & 90.9 & - & - \\
    \cmidrule{2-6}  
      & PATH~\cite{tang2023humanbench} & ViT-L & 90.8 & - & - \\
    \midrule
    \multirow{6}{*}{Generalist} 
      & UniHCP~\cite{ci2023unihcp} & ViT-B & 90.0 & 46.6 & 82.2 \\
      & UniHCP-FT~\cite{ci2023unihcp}  & ViT-B & \textbf{92.5} & 41.6 & 85.8 \\
      & Hulk\dag  & ViT-B & 90.7 & 43.8 & 84.0 \\
      & Hulk-FT\dag & ViT-B & 92.4 & \textbf{40.7} & \textbf{86.0} \\
    \cmidrule{2-6}  
      & Hulk\dag  & ViT-L & 92.2 & 40.1 & 85.8 \\
      & Hulk-FT\dag & ViT-L & 93.0 & \textbf{36.5} & 87.0 \\
    \bottomrule
    \end{tabular}%
  }
  \vspace{-0.5em}
\end{table}

\begin{table}[t]
  \centering
  \caption{2D Pose estimation evaluation on COCO~\cite{coco} val set and on AIC~\cite{aic} test set with mAP. We compare our method with other SOTA methods, both specialists and generalists, with an input resolution of 256$\times$192. \dag indicates our implementation. \ddag indicates that multiple datasets are used.}
  \label{tab:pose}
    \begin{tabular}{cllcc}
    \toprule
    \multicolumn{2}{l}{Method } & Backbone & COCO  & AIC \\
    \midrule
    \multirow{10}[6]{*}{Specialist} 
          & HRNet~\cite{sun2019deep} & HRNet-W48 & 75.1  & - \\
          & TokenPose-L/D24~\cite{li2021tokenpose} & HRNet-W48 & 75.8  & - \\
          & HRFormer-B~\cite{yuan2021hrformer} & HRFormer-B & 75.6  & - \\
\cmidrule{2-5} 
& ViTPose-B~\cite{xu2022vitpose} & ViT-B & 75.8  & - \\
          & ViTPose-B‡~\cite{xu2022vitpose} & ViT-B & 77.1  & 32.0  \\
\cmidrule{2-5}          
          & {ED-Pose~\cite{yang2023explicit}}&{Swin-L} &{75.8}&{-}\\
          & ViTPose-L~\cite{xu2022vitpose} & ViT-L & 78.3  & - \\
          & ViTPose-L‡~\cite{xu2022vitpose} & ViT-L & 78.7  & 34.5  \\
    \midrule
    \multirow{5}[2]{*}{Pretraining} 
          & HCMoCo\dag~\cite{hcmoco} & HRNet-W48 & 76.9  & - \\
          & SOLDIER~\cite{solider} & Swin-B & 76.6  & - \\
          & HAP~\cite{yuan2023hap}   & ViT-B & 77.0  & 32.3  \\
          & PATH~\cite{tang2023humanbench} & ViT-B & 76.3  & 35.0  \\
          \cmidrule{2-5}   
          & PATH~\cite{tang2023humanbench} & ViT-L & 77.1  & 36.3  \\
    \midrule
    \multirow{6}[4]{*}{Generalist} & UniHCP~\cite{ci2023unihcp} & ViT-B & 76.1  & 32.5  \\
          & UniHCP-FT~\cite{ci2023unihcp}  & ViT-B & 76.6  & 33.6  \\
          & Hulk  & ViT-B & 77.0  & 34.5  \\
          & Hulk-FT & ViT-B &  \textbf{77.5}     & \textbf{35.6}  \\
\cmidrule{2-5}          & Hulk  & ViT-L & 78.3  & 36.3  \\
          & Hulk-FT & ViT-L &  \textbf{78.7}    & \textbf{37.1} \\
    \bottomrule
    \end{tabular}%
    \vspace{-1em}
\end{table}%

\begin{table}[t]
  \centering
  \caption{Human parsing evaluation (mIoU) on Human3.6m~\cite{h36m_pami}, LIP~\cite{lip} and CIHP~\cite{cihp}. We compare Hulk with other SOTA methods, including specialists, generalists, and pretrainings, with an input resolution of 480$\times$480.~\dag indicated re-implementation.}
  \label{tab:parsing}
  \resizebox{0.95\linewidth}{!}{
    \begin{tabular}{cllccc}
    \toprule
\multicolumn{2}{l}{Method } & Backbone & H3.6M & LIP   & \multicolumn{1}{l}{CIHP} \\
    \midrule
    \multirow{2}[2]{*}{Specialist} 
          & SCHP~\cite{li2020self}   &   ResNet-101    & -     & 59.36  & - \\
          & CDGNet~\cite{liu2022cdgnet} &    ResNet-101   & -     & 60.30  & 65.56  \\
    \midrule
    \multirow{5}[4]{*}{Pretraining} 
          & HCMoCo~\cite{hcmoco} & HRNet-18 & 62.50  & -     & - \\
          & HCMoCo\dag~\cite{hcmoco} & HRNet-48 &   66.01      &   -    & - \\
          & SOLDIER~\cite{solider} & Swin-B & -     & 60.50  & - \\
          & PATH~\cite{tang2023humanbench} & ViT-B & 65.00  & 61.40  & 66.80  \\
\cmidrule{2-6}  
          & PATH~\cite{tang2023humanbench} & ViT-L & 66.20  & 62.60  & 67.50  \\
    \midrule
    \multirow{6}[4]{*}{Generalist} & UniHCP~\cite{ci2023unihcp} & ViT-B & 65.90  & 63.80  & 68.60  \\
          & UniHCP-FT~\cite{ci2023unihcp} & ViT-B & 65.95  & 63.86  & 69.80  \\
          & Hulk  & ViT-B & 68.08  & 63.95  & 70.58  \\
          & Hulk-FT & ViT-B & \textbf{68.56}  & \textbf{63.98}   & \textbf{71.26}\\
\cmidrule{2-6}          & Hulk  & ViT-L & 69.31  & 65.86  & 72.33  \\
          & Hulk-FT & ViT-L & \textbf{69.89}  & \textbf{66.02}  & \textbf{72.68} \\
    \bottomrule
    \end{tabular}%
}
\end{table}%

\noindent \textbf{Pedestrian detection.} In the pedestrian detection task, 
MR$^{-2}$ is employed as a more sensitive and crucial metric in detection in the crowd scenes.
This metric is particularly sensitive to False Positives (FP), with lower values indicating more desirable performance. We use an input image resolution of a maximum height/weight of 1120, while other methods take a large resolution of 1333 by default. As shown in Table ~\ref{tab:detection}, Hulk showcases competitive performance with State-Of-The-ART (SOTA) using a smaller input resolution. Specifically, after finetuning,
our Hulk achieves an improvement of \textbf{-0.7\%} MR$^{-2}$ on ViT-B and  \textbf{-1.2\%} MR$^{-2}$ on ViT-L, respectively, compared to SOTA specialist methods, \emph{e.g.}, Iter-DDETR~\cite{zheng2022progressive}.

\begin{table}[t]
  \centering
  \caption{Pedestrian attribute recognition evaluation on PA-100K~\cite{pa} and RAPv2~\cite{rapv2} with mA reported.}
  \label{tab:attribute}
    \begin{tabular}{cllcc}
    \toprule
    \multicolumn{2}{l}{Method } & Backbone & PA-100K & RAPv2 \\
    \midrule
    \multirow{3}[2]{*}{Specialist} & SSC~\cite{jia2021spatial}   & ResNet-50 & 81.87  & - \\
          & L2L~\cite{li2022label2label}    & ViT-B & 82.24  & - \\
          & DAFL~\cite{jia2022learning}   & ResNet-50 & 83.54  & 81.04  \\
    \midrule
    \multirow{4}[2]{*}{Pretraining} & SOLIDER~\cite{solider} & Swin-B & 86.37  & - \\
          & HAP~\cite{yuan2023hap}   & ViT-B & 86.54  & 82.91  \\
          & PATH~\cite{tang2023humanbench} & ViT-B & 86.90  & 83.10  \\
\cmidrule{2-5}          
          & {PATH~\cite{tang2023humanbench}}  & {ViT-L} & {\textbf{90.80}}  & {\textbf{87.40}}  \\
    \midrule
    \multirow{6}[3]{*}{Generalist} & UniHCP~\cite{ci2023unihcp} & ViT-B & 79.32  & 77.20  \\
          & UniHCP-FT~\cite{ci2023unihcp}  & ViT-B & 86.18  & 82.34  \\
          & Hulk  & ViT-B & 82.85  & 80.90  \\
          & Hulk-FT & ViT-B &   \textbf{87.85}    & \textbf{85.26} \\
\cmidrule{2-5}          & Hulk  & ViT-L & 84.36  & 82.85  \\
          & Hulk-FT & ViT-L &  {88.97}     & {85.86 }\\
    \bottomrule
    \end{tabular}%
\vspace{-1em}
\end{table}%

\noindent \textbf{2D pose estimation.}
To evaluate the performance of Hulk on 2D pose estimation task, we conduct evaluations on both the COCO and AIC datasets. 
The results are listed in Table~\ref{tab:pose}. Hulk
improves previous SOTA by considerable margins on AIC, \emph{i.e.}, \textbf{+0.6\%} and \textbf{+0.8\%} AP on ViT-B and ViT-L, respectively. The performance gain demonstrates the good scalability of Hulk. Hulk achieves on-par performance compared with ViTPose++~\cite{xu2023vitpose++} on COCO. We attribute the relatively small performance gain on COCO to the distribution gap between the training set and val set, where models are trained with on ground-truth boxes but evaluated on estimated boxes with a detector obtaining human AP of 56 on COCO val set~\cite{mmpose2020}.

\begin{table}[t]
  \centering
   \footnotesize
  \caption{Skeleton-based Action Recognition evaluation on NTU60-XSub~\cite{liu2019ntu} and FineGYM~\cite{shao2020finegym} with accuracy reported. }
  \label{tab:skeleton}
   \resizebox{.95\linewidth}{!}{
    \begin{tabular}{lllcc}
   
    \toprule
    Method &       & Backbone  & NTU60 & {GYM}\\
    \midrule
    \multicolumn{1}{c}{\multirow{9}[5]{*}{Specialist}} & ST-GCN~\cite{stgcn} & GCN   & 81.5 &{25.2}  \\
          & Shift-GCN~\cite{shiftgcn} & GCN   & 90.7  &{-} \\
          & CTR-GCN~\cite{ctrgcn} & GCN   & 92.4 &{-}   \\
          & MS-G3D~\cite{g3d} & GCN   & 91.5 &{92.0} \\
        
\cmidrule{2-5}          
          & PoseConv3D~\cite{posec3d} & CNN   & 93.1  &{\textbf{92.4}} \\
          &SELFYNet~\cite{kwon2021learning} &CNN &- &{87.7}\\
                        \cmidrule{2-5}    
& ST-TR~\cite{plizzari2021skeleton} & Transformer & 89.9 &{-}  \\

          & DSTA-Net~\cite{shi2020decoupled} & Transformer & 91.5  &{-} \\
          & STTFormer~\cite{qiu2022spatio} & Transformer & 89.9&{-}   \\
              \midrule
      \multirow{2}[2]{*}{Pretraining}& MotionBERT~\cite{zhu2023motionbert} & DSTformer & 93.0 &{-} \\
     & SkeletonMAE~\cite{yan2023skeletonmae} & GIN  &- &{91.8}  \\
    \midrule
    \multirow{4}[3]{*}{Generalist} & Hulk  & ViT-B & 93.8 &{91.6}\\
          & Hulk-FT & ViT-B &  \textbf{94.0} &{92.2}\\
\cmidrule{2-5}          & Hulk  & ViT-L &  94.1 &{92.3}\\
          & Hulk-FT & ViT-L & \textbf{94.3}  &{\textbf{93.2}}\\
    \bottomrule
    \end{tabular}%
    }
    \vspace{-1.2em}
\end{table}%

\noindent \textbf{Human parsing.} 

We evaluate the performance of Hulk on the human parsing task on Human3.6M, LIP, and CIHP datasets. As shown in Table~\ref{tab:parsing}, when using ViT-B, Hulk surpasses SOTA methods by \textbf{+2.36\%} mIoU, \textbf{+1.46\%} mIoU on Human3.6M and CIHP, respectively. On LIP, although Hulk achieves a relatively small performance gain with ViT-B, Hulk showcases considerable performance with ViT-B, \emph{i.e.}, 66.02\% mIoU, demonstrating excellent scalability of Hulk.

\noindent \textbf{Pedestrian attribute recognition.} 

We evaluate the performance of Hulk on PA-100K and RAPv2 to demonstrate its effectiveness on the pedestrian attribute recognition task. 
As shown in Table~\ref{tab:attribute}, on ViT-B, Hulk outperforms previous methods by obtaining 87.85\% mA and 85.26\% mA on PA-100K and RAPv2, respectively. The performance of the large Hulk with ViT-B further increases to 88.97\% mA and 85.86\% mA, showing competitive results without adopting task-specific heads in human-centric pretraining models~\cite{tang2023humanbench}.

\subsubsection{Skeleton-based Tasks}
In the \textbf{skeleton-based action recognition} task, existing methods utilize three categories of backbones: (1) Graph Convolutional Networks (GCN),  which are the most common architecture; (2) Convolutional Neural Networks (CNN) and Recurrent Neural Networks (RNN), and (3) Transformer architectures. Our Hulk falls within the third category. 
To assess the effectiveness of Hulk, we report results on NTU60-XSub, a well-known dataset on skeleton-based action recognition. 
The results in Table~\ref{tab:skeleton} indicate that even without specific pretraining designs~\cite{zhu2023motionbert}, Hulk obtains a remarkable accuracy of \textbf{93.8\%} on ViT-B, better than the SOTA PoseConv3D~\cite{posec3d} method. After finetuning Hulk with ViT-L backbone, the performance on NTU60-XSub further increases to \textbf{94.3\%}. This improvement not only highlights the effectiveness of Hulk in skeleton-based action recognition, but also demonstrates Hulk's ability of tokenizing skeleton and image inputs into a modality-shared manifold space to extract generalized human-centric knowledge.
{Using the skeleton data in the FineGYM dataset as input, Hulk is also better than the state-of-the-art methods, achieving the accuracies of 92.2 and 93.2 on ViT-B and ViT-L, respectively. }

\begin{table*}[t]
  \centering
  \caption{Monocular 3D human pose and mesh
recovery evaluation on 3DPW~\cite{3dpw} and Human3.6M~\cite{h36m_pami} among image-based methods.}
\label{tab:smpl}
\resizebox{.75\linewidth}{!}{
    \begin{tabular}{cllrrrrr}
    \toprule
    \multicolumn{2}{l}{\multirow{2}[4]{*}{Method}} & \multirow{2}[4]{*}{Backbone } & \multicolumn{3}{c}{3DPW} & \multicolumn{2}{c}{Human3.6M} \\
\cmidrule(r){4-6} \cmidrule(r){7-8}    \multicolumn{2}{l}{} &       & \multicolumn{1}{c}{MPVPE$\downarrow$} & \multicolumn{1}{c}{MPJPE$\downarrow$} & \multicolumn{1}{c}{PA-MPJPE$\downarrow$} & \multicolumn{1}{c}{MPJPE$\downarrow$} & \multicolumn{1}{c}{PA-MPJPE$\downarrow$} \\
    \midrule
    \multirow{7}[2]{*}{Specialist} 
          & METRO~\cite{metro} & HRNet-W64 & \multicolumn{1}{c}{88.2} & \multicolumn{1}{c}{77.1} & \multicolumn{1}{c}{47.9} & \multicolumn{1}{c}{54.0} & \multicolumn{1}{c}{36.7} \\
          & PARE~\cite{kocabas2021pare}  & HRNet-W32 & \multicolumn{1}{c}{88.6} & \multicolumn{1}{c}{74.5} & \multicolumn{1}{c}{46.5} & \multicolumn{1}{c}{-} & \multicolumn{1}{c}{-} \\
          & ProHMR~\cite{prohmr} & ResNet-50      & \multicolumn{1}{c}{-} & \multicolumn{1}{c}{-} & \multicolumn{1}{c}{59.8} & \multicolumn{1}{c}{-} & \multicolumn{1}{c}{41.2} \\
          & FastMETRO~\cite{fastmetro} & ResNet-50 & \multicolumn{1}{c}{90.6} & \multicolumn{1}{c}{77.9} & \multicolumn{1}{c}{48.3} & \multicolumn{1}{c}{53.9} & \multicolumn{1}{c}{37.3} \\
          & FastMETRO~\cite{fastmetro} & HRNet-W64 & \multicolumn{1}{c}{84.1} & \multicolumn{1}{c}{73.5} & \multicolumn{1}{c}{44.6} & \multicolumn{1}{c}{52.2} & \multicolumn{1}{c}{33.7} \\
          & CLIFF~\cite{li2022cliff} & ResNet-50      & \multicolumn{1}{c}{81.2} & \multicolumn{1}{c}{69.0} & \multicolumn{1}{c}{43.0} & \multicolumn{1}{c}{47.1} & \multicolumn{1}{c}{32.7} \\
          & VisDB~\cite{visdb} & ResNet-50      & \multicolumn{1}{c}{85.5} & \multicolumn{1}{c}{73.5} & \multicolumn{1}{c}{44.9} & \multicolumn{1}{c}{51.0} & \multicolumn{1}{c}{34.5} \\
    \midrule
    \multirow{2}[1]{*}{Pretraining} 
          & MotionBERT~\cite{zhu2023motionbert} & DSTformer & \multicolumn{1}{c}{88.1} & \multicolumn{1}{c}{76.9} & \multicolumn{1}{c}{47.2} & \multicolumn{1}{c}{53.8} & \multicolumn{1}{c}{34.9} \\
          & HAP~\cite{yuan2023hap}   & ViT-B & \multicolumn{1}{c}{90.1} & \multicolumn{1}{c}{56.0} & \multicolumn{1}{c}{106.3} & \multicolumn{1}{c}{-} & \multicolumn{1}{c}{-} \\
        \midrule
    \multirow{4}[3]{*}{Generalist} & Hulk  & ViT-B & \multicolumn{1}{c}{\textbf{79.8}} & \multicolumn{1}{c}{\textbf{67.0}} & \multicolumn{1}{c}{\textbf{39.9}} & \multicolumn{1}{c}{\textbf{43.6}} & \multicolumn{1}{c}{\textbf{31.9}} \\
          & Hulk-FT & ViT-B & \multicolumn{1}{c}{80.7} & \multicolumn{1}{c}{68.9} & \multicolumn{1}{c}{41.3} & \multicolumn{1}{c}{44.9} & \multicolumn{1}{c}{32.0} \\
\cmidrule{2-8}          & Hulk  & ViT-L & \multicolumn{1}{c}{\textbf{77.4}} & \multicolumn{1}{c}{\textbf{66.3}} & \multicolumn{1}{c}{\textbf{38.5}} & \multicolumn{1}{c}{\textbf{40.3}} & \multicolumn{1}{c}{\textbf{28.8}} \\
          & Hulk-FT & ViT-L& \multicolumn{1}{c}{79.9} & \multicolumn{1}{c}{68.3} & \multicolumn{1}{c}{40.6} & \multicolumn{1}{c}{41.4} & \multicolumn{1}{c}{30.2} \\
    \bottomrule
    \end{tabular}%
    \vspace{-2.em}
}
\end{table*}%

\begin{table}[t]
  \centering
  \caption{Pedestrian image caption evaluation on CUHK-PEDES~\cite{cuhk}. \dag denotes fixed backbone. PD - Pretrain dataset. TM - whether has pretrained Text Models or not.}
  \label{tab:caption}
   \resizebox{\linewidth}{!}{
    \begin{tabular}{cllccccc}
    \toprule
    \multicolumn{1}{l}{Model} &       & Backbone & {{{Params}}} &{PD} &{TM} & B@4   & CIDEr  \\
    \midrule
    \multirow{4}[1]{*}{Specialist} & BLIP~\cite{li2022blip} &  ViT-B  & {446.0M} &{129M} & {\checkmark} & \textbf{32.9}     & 103.8   \\
          & BLIP-2~\cite{li2023blip} &  {ViT-G\dag} &  {1.1B} &{129M} &{\checkmark}  &  32.8     & \textbf{105.8}   \\
       & {OFA-L~\cite{wang2022ofa}} &  {ResNet-101}  &{472.0M} & {18M} & {$\times$} & {31.5}      & {97.7}   \\
        & {mPLUG-B~\cite{li2022mplug}} &  {ViT-B}   &{350.0M} & {14M} &{\checkmark} &  {32.2}     & {101.1}  \\
    \midrule
    \multirow{4}[4]{*}{Generalist} & Hulk  & ViT-B &{113.7M}&  {12M} & {$\times$} & 31.1     & 91.4 \\
          & Hulk-FT & ViT-B&{113.7M} & {12M} & {$\times$} & 28.3      & 72.8 \\
\cmidrule{2-8}          & Hulk  & ViT-L&{336.0M} &  {12M} & {$\times$} & 31.6     & 94.5 \\
          & Hulk-FT & ViT-L &{336.0M} & {12M} & {$\times$} & 30.5      & 82.8   \\
    \bottomrule
    \end{tabular}}
\vspace{-1.5em}
\end{table}%

\subsubsection{Vision-language Tasks}
We also evaluate Hulk on the CUHK-PEDES dataset for the vision-language task, \emph{i.e.,} \textbf{pedestrian image caption}. 
{For fair comparisons, we re-implemented the state-of-the-art methods, \emph{i.e.}, OFA~\cite{wang2022ofa}, mPLUG~\cite{li2022mplug}, BLIP~\cite{li2022blip} and BLIP-2~\cite{li2023blip}, by finetuning their pre-trained weights on the CUHK-PEDES dataset. As shown in Table~\ref{tab:caption},
Hulk achieves on-par performance with SOTA methods on the B@4 metric but lags behind in terms of CIDEr. This discrepancy in performance can be attributed to two factors: (1) the extensive pretraining of BLIP and BLIP-2 on large-scale vision-language datasets, while our Hulk is trained solely on smaller human-centric vision-language datasets (about 12M). (2) For unifying other human-centric tasks, Hulk does not incorporate a pretrained text encoder/decoder, a crucial module for generating coherent text outputs in modern caption models.}
Furthermore, we observed that finetuning on the CUHK-PEDES dataset led to a decline in the model's prediction accuracy, because better pedestrian
captions can learn from diverse annotations in large-scale 
human-centric datasets.

\subsubsection{3D Vision Tasks}
In the case of the 3D vision task, \emph{i.e.,} \textbf{3D human pose estimation} and \textbf{mesh recovery}, 
our Hulk improves the previous state-of-the-art method CLIFF~\cite{li2022cliff}, by \textbf{-3.8} MPVPE, \textbf{-2.7} MPJPE, \textbf{-4.5} PA-MPJPE on 3DPW dataset, and \textbf{-6.8} MPJPE, \textbf{-3.9} PA-MPJPE on Human3.6M dataset, in direct evaluation scenarios. Interestingly, comparing with direct evaluation results, we observed that finetuning on both 3DPW and Human3.6M datasets will lead to lower performances, showing the effectiveness of pretraining our Hulk on diverse tasks.

\subsection{Ablation Study}

To demonstrate the effectiveness of Hulk, we conduct several ablations on a smaller training set. The ablation set consists of 21 datasets with about 1.7M samples, covering all eight tasks. We provide the detailed dataset combination in the supplemental material. Without otherwise specified, we train Hulk for 20k iterations in all ablation variants. 
While results on different datasets within the same task are highly related, we showcase one dataset per task to represent the performance on the specific task in the main text (all results are provided in the supplemental material). As the metric in 3D pose estimation and mesh recovery computes the error, we use $100-error$ when we need to compute average performance among 8 tasks.

\subsubsection{Weight sharing}
As Hulk demonstrates promising performance in various human-centric tasks while sharing mostly all parameters, we explore the impact of incorporating more task-specific parameters into our learning framework. To assess the effectiveness of our weight-sharing approach, we ablate three weight-sharing variants: 
(a) Our Hulk model, all parameters in both the encoder and decoder are shared across all datasets, and parameters in the tokenizers and de-tokenizers are shared within the same input/output modality. (b) All parameters in both the encoder and decoder are shared across all datasets, but the parameters in the tokenizers and de-tokenizers are dataset-specific. (c) Only encoder parameters are shared. 
In Table~\ref{tab:abl_share}, we observed that more parameter sharing enhances the overall performance of the model. Specifically, introducing decoder parameter sharing (from (b) to (c)) significantly benefits average performances by \textbf{$\textbf{+12.1\%}$}.
This shows that shared human semantic information in different human-related annotations is beneficial to human-centric tasks.
A notable example is the skeleton-based action recognition task. When only the encoder parameters are shared, the model fails to converge because of the small size of the NTU60-XSub dataset, resulting in only $\textbf{1.7}\%$ accuracy. However, by sharing more model parameters, the general human-centric knowledge from other tasks can be substantially beneficial to better accuracy in this task.

\begin{table*}[t]
  \centering
  \caption{Comparison of different parameter-sharing schemes. $\mathcal{E}$, $\mathcal{D}$, $\mathcal{P}$, and $\mathcal{Q}$ denote encoder, decoder, tokenizer, and de-tokenizer, respectively. With more task-specific parameters, the average tasks performance of Hulk declines significantly.}
    \label{tab:abl_share}
    \resizebox{.95\textwidth}{13.2mm}{
    \begin{tabular}{lccccccccccccc}
    \toprule
    \multicolumn{1}{c}{\multirow{3}[6]{*}{Methods}} & \multicolumn{4}{c}{\multirow{2}[4]{*}{Shared module}} & Parsing & 2D Pose & Detection & Attribute & Caption & Skeleton  & Mesh & 3D Pose  & \multirow{3}[6]{*}{Avg} \\
  \cmidrule(r){6-6}   \cmidrule(r){7-7}   \cmidrule(r){8-8}   \cmidrule(r){9-9}   \cmidrule(r){10-10}   \cmidrule(r){11-11}   \cmidrule(r){12-12}   \cmidrule(r){13-13}           & \multicolumn{4}{c}{}          & LIP   & AIC   & CrowdHuman & RAPv2 & CUHK-PEDES & NTU60-XSub & 3DPW  & 3DPW  &  \\
\cmidrule(r){2-5}    \cmidrule(r){6-6}   \cmidrule(r){7-7}   \cmidrule(r){8-8}   \cmidrule(r){9-9}   \cmidrule(r){10-10}   \cmidrule(r){11-11}   \cmidrule(r){12-12}   \cmidrule(r){13-13}        & $\mathcal{E}$ & $\mathcal{D}$ & $\mathcal{P}$ & $\mathcal{Q}$ & mIoU  & AP    & AP    & mA    & B@4   & Acc   & 100-MPVPE & 100-MPJPE &  \\
    \midrule
    (a) Hulk & \checkmark     & \checkmark     & \checkmark     & \checkmark     & 56.8  & 25.5  & 81.8  & 77.9  & 28.0  & 93.2  & -0.3  & 13.2  & 41.8  \\
    (b)   & \checkmark     & \checkmark     &       &       & 56.8  & 25.8  & 80.6  & 78.8  & 26.7  & 93.8  & -3.0  & 10.7  & 41.1  \\
    (c)   & \checkmark     &       &       &       & 56.3  & 25.7  & 75.9  & 81.3  & 28.7  & 1.7   & -11.2  & 2.9   & 29.0  \\
    \bottomrule
    \end{tabular}%
    }
\end{table*}%

\begin{table*}[t]
  \centering
  \caption{Ablation of collaboration and interference between tasks. In \textcolor[rgb]{0.2,0.8,0.2}{green} and \textcolor{red}{red} are the gaps of at least $\pm0.5 $ points.}
  \resizebox{.95\textwidth}{!}{
    \begin{tabular}{lccccccccc}
    \toprule
    \multirow{3}[6]{*}{Method} & Parsing & 2D Pose & Detection & Attribute & Caption & Skeleton & Mesh  & \multicolumn{1}{c}{3D Pose} & \multirow{3}[6]{*}{Avg. }\\
\cmidrule(r){2-2} \cmidrule(r){3-3} \cmidrule(r){4-4} \cmidrule(r){5-5} \cmidrule(r){6-6} \cmidrule(r){7-7} \cmidrule(r){8-8} \cmidrule(r){9-9}     & LIP   & AIC   & CrowdHuman & RAPv2 & CUHK-PEDES & NTU60-XSub & 3DPW  & \multicolumn{1}{c}{3DPW} &\\
\cmidrule(r){2-2} \cmidrule(r){3-3} \cmidrule(r){4-4} \cmidrule(r){5-5} \cmidrule(r){6-6} \cmidrule(r){7-7} \cmidrule(r){8-8} \cmidrule(r){9-9}        & mIoU  & AP    & AP    & mA    & {B@4} & Acc   &100-MPVPE & \multicolumn{1}{c}{100-MPJPE}  \\
    \midrule
    baseline & 56.8  & 25.5  & 81.8  & 77.9  & 28.0    & 93.2  & -0.3 & 13.2    & -\\
    w/o Parsing &   -    & 25.2\smalldown{0.3}  & 78.5\down{3.3}  & 77.3\down{0.6}  & 25.8\down{2.2}  & 93.2\smallup{0.0}  & -2.9\down{2.6}  & 10.8\down{2.4}   & \textcolor{red}{-1.6} \\
    w/o 2D Pose & 52.3\down{4.5}  &   -    & 65.0\down{16.8}  & 75.1\down{2.8}  & 26.6\down{1.5}  & 93.5\smallup{0.3}  & -18.9\down{18.6}  & -3.5\down{16.7}   & \textcolor{red}{-8.6} \\
    w/o Detection & 57.6\up{0.8}  & 26.4\up{0.9}  &   -    & 77.6\smalldown{0.3}  & 27.8\smalldown{0.2}  & 93.1\smalldown{0.1} & -4.8\down{4.5}  & 10.5\down{2.7}   & \textcolor{red}{-0.9}\\
    w/o Attribute & 56.9\smallup{0.1}  & 25.6\smallup{0.1}  & 79.9\down{1.9}  &   -    & 28.2\smallup{0.2}  & 93.1\smalldown{0.1}  & -2.1\down{1.8}  & 13.2\down{1.0}   &\textcolor{red}{-0.6} \\
    w/o Caption  & 57.3\up{0.5}  & 25.6\smallup{0.1}  & 80.2\down{1.6}  & 77.8\smalldown{0.1}  &    -   & 92.9\smalldown{0.3}  & -2.2\down{1.9}  & 11.2\down{2.0}   & \textcolor{red}{-0.8}\\
    w/o Skeleton & 57.4\up{0.6}  & 25.7\smallup{0.2}  & 80.9\down{0.9}  & 77.4\down{0.5}  & 28.2\smallup{0.2}  &    -   & -4.8\down{4.5}  & 9.9\down{3.3}  & \textcolor{red}{-1.2}\\
    w/o Mesh\&3D Pose & 57.4\up{0.6}  & 25.5\smallup{0.0}  & 81.0\down{0.8}  & 78.0\smallup{0.1}  & 27.9\smalldown{0.1}  & 93.1\smalldown{0.1}  &   -    &   -    &  \textcolor[rgb]{0.5,0.6,0.7}{-0.1}\\
        
    \bottomrule
    \end{tabular}%
    }
    \vspace{-1em}
  \label{tab:collaboration and interference}%
\end{table*}%

\begin{table}[t]
  \centering
  \caption{Ablation of how generic tasks help smaller ones.}
  \resizebox{0.95\linewidth}{!}{
    \begin{tabular}{lccc}
    \toprule
    \multirow{3}[2]{*}{Task} & {Detection} & {Attribute} & {Skeleton} \\
    \cmidrule(r){2-2} \cmidrule(r){3-3}  \cmidrule(r){4-4}
          & {CrowdHuman} &{RAPv2} & {NTU60-XSub} \\
          \cmidrule(r){2-2} \cmidrule(r){3-3}  \cmidrule(r){4-4}
          & {AP} & {mA} & {Acc} \\
    \midrule
    baseline & 52.6  & 75.4  & 93.1 \\
    baseline + 2D pose joint train & 71.6  & 78.7  & 93.5 \\
    baseline + 3D pose joint train & 57.8  & 76.5 & 93.2 \\
    \bottomrule
    \end{tabular}%
    }
    \vspace{-2em}
  \label{tab:one_add_one}%
\end{table}%

\subsubsection{Task Collaboration and Interference}
In this section, we explore how various human-centric tasks influence each other. We conduct leave-one-out experiments that we remove one task from our training set every time (as detailed in Table~\ref{tab:collaboration and interference}). Notably, mesh recovery and 3D pose estimation were always conducted together. We compare these scenarios against a baseline to determine if omitting a task would positively or negatively impact the performance of others.

We observe several key findings. 
First, {
while the removal of certain tasks can result in performance improvements in some cases, in most situations, removing one task from training will lead to an average performance decline}. 
{\uline{Among all tasks, 2D pose recognition, as a generic task, is mostly beneficial for all others.}} Particularly, removing the 2D Pose task significantly decreased performance in detection (\textbf{-16.8} AP), mesh recovery (increased error by \textbf{18.6} MPVPE), and 3D pose estimation (increased error by \textbf{16.7} MPJPE). The 2D keypoint information in 2D pose estimation is highly important to the 3D keypoint for 3D pose estimation and can help the model perceive the boundary of pedestrians for pedestrian detection, making the 2D pose estimation task an important complement to training a human-centric foundation model. 
Second, tasks such as the performances of pedestrian detection, mesh recovery, and 3D pose estimation are more sensitive to the removal of other tasks. Their performance markedly declined, indicating a high interdependence among these human-centric tasks during training. 
Third, skeleton-based action recognition was less affected by the absence of other tasks, which may stem from its unique input type (skeleton points) compared to the image-based input of other tasks. 
It is also worth noting that the absence of certain tasks resulted in performance gains for human parsing, indicating potential task conflicts with other human-centric tasks that require further investigation.

\begin{figure}[t]
    \centering
    \includegraphics[width=0.95\linewidth]{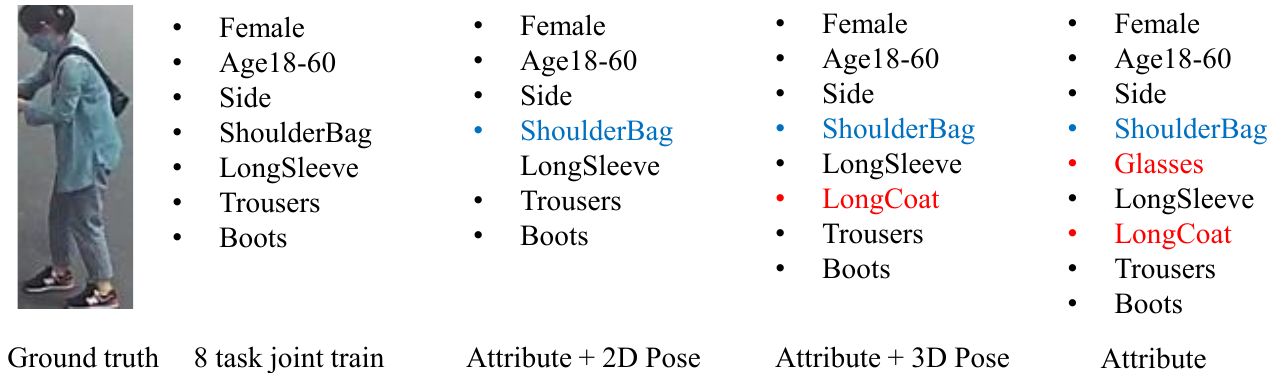}
    \caption{{Visualization of attribute recognition on PA-100K~\cite{pa}. The \textcolor[rgb]{0,0.40,0.701}{missed} and \textcolor{red}{incorrectly identified} attributes are marked in blue and red, respectively.}}
    \label{fig:attr+pose}
    \vspace{-1em}
\end{figure}

\begin{figure}[t]
    \centering
    \includegraphics[width=0.99\linewidth]{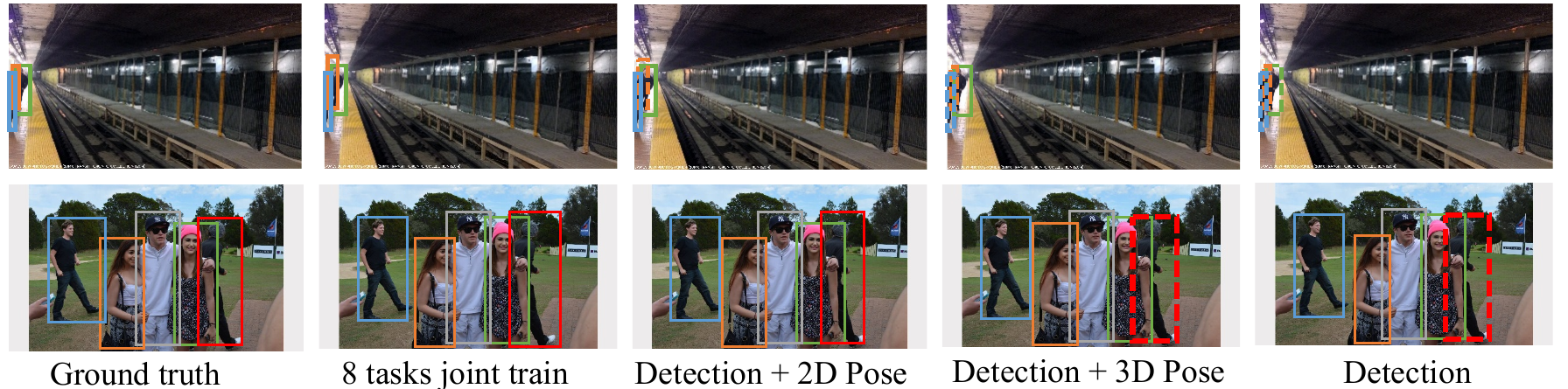}
    \caption{{Visualization of pedestrian detection on CrowdHuman~\cite{shao2018crowdhuman}. The dashed boxes indicate missed detections.}}
    \label{fig:det+pose}
    \vspace{-1.5em}
\end{figure}

{
{After further exploring how generic tasks help the smaller ones during Hulk training, we find that the relations between them determine their mutual help. A strong correlation between the general task and the small task amplifies the benefits of the large task, and conversely. We train Hulk with different pretraining datasets for 20,000 iterations. As shown in Table~\ref{tab:one_add_one} involving generic tasks (2D/3D pose estimation) does not necessarily improve performance on smaller tasks, \emph{e.g.}, pedestrian detection, attribute recognition, and skeleton-based action recognition tasks.} Specifically, adding 2D/3D pose estimation task brings $\textbf{+19.0\%} AP$/$\textbf{+5.2\%} AP$ and $\textbf{+3.3\%} mA$/$\textbf{+1.1\%} mA$ performance gain on CrowdHuman and RAPv2, respectively. We suggest that the 
information in human joint location annotations in 2D and 3D pose estimation helps Hulk to better focus on the region of pedestrians, facilitating its performance on detection and attribute recognition. In contrast, the skeleton-based action recognition task, which focuses on the semantics of human motion sequences, has a weak correlation with the joint location information learned by 2D/3D pose estimation tasks, resulting in slight performance improvements. {Qualitative results in Fig.~\ref{fig:attr+pose} and \ref{fig:det+pose} also show that jointly training with generic tasks can benefit the learning on pedestrian detection and attribute recognition.}
}


\subsubsection{Attention Mask Designs in the Decoder}
\begin{figure}
    \centering
    \includegraphics[width=\linewidth]{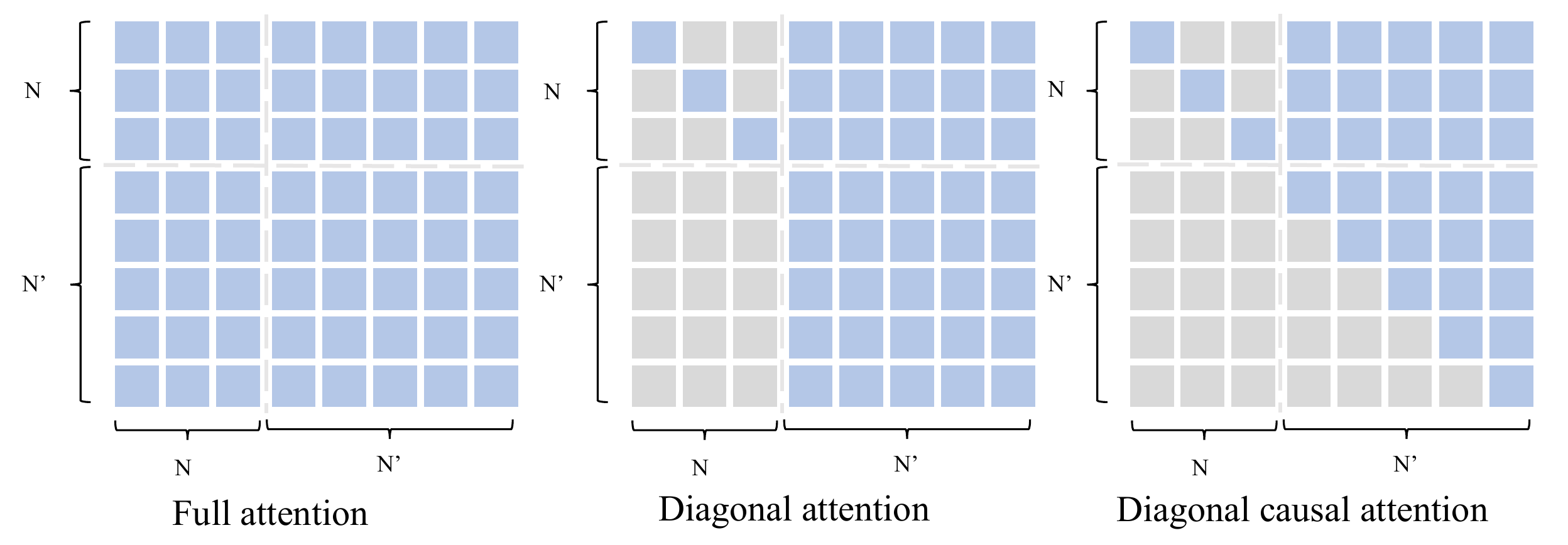}
    \caption{Given $N$ encoded tokens $\mathcal{E}(\mathbf{p})$ with $N'$ modality indicators $\mathcal{I}_{m'}$, we proposed various attention mask in the decoder $\mathcal{D}$ to facilitate diverse tasks. \textbf{Left}: Full attention enables comprehensive interaction between encoder tokens and modality indicators. \textbf{Mid}: Diagonal attention mitigates potential distortions in encoded tokens due to randomly initialized modality indicators. \textbf{Right}: A combination of the causal and diagonal masks for image captioning.}
    \vspace{-1.5em}
    \label{fig:attn mask}
\end{figure}

The receptive field of the attention module in the decoder is crucial for the translation and de-tokenization of features in the unified model. Therefore, we investigate the effects of different attention masks as illustrated in Fig.~\ref{fig:attn mask}: (a) our Hulk model adopts task-specialized attention masks for different tasks, i.e., for pose estimation and mesh recovery tasks, we employ full attention; for parsing, detection, attribute, and skeleton-based tasks, we use diagonal attention; for caption tasks, we apply a combination of causal and diagonal attention. We also experimented with (b) all tasks using full attention and (c) all tasks using diagonal attention. The results, shown in Table~\ref{tab:abl_attn}{(a-c)}, indicate that employing different attention interaction methods for different tasks can improve the performances, \emph{i.e.}, \textbf{+0.4\%} and \textbf{+1.0\%} average performance gain on full attention and diagonal attention, respectively.

\begin{table*}[t]
  \centering
  \caption{Comparison of different attention masks and different initialization parameters.}
  \label{tab:abl_attn}
    \begin{tabular}{lccccccccc}
    \toprule
    \multirow{3}[6]{*}{Methods} & Parsing & 2D Pose & Detection & Attribute & Caption & Skeleton  & Mesh & 3D Pose  & \multirow{3}[6]{*}{Avg} \\
 \cmidrule(r){2-2}   \cmidrule(r){3-3}   \cmidrule(r){4-4}   \cmidrule(r){5-5}  \cmidrule(r){6-6}   \cmidrule(r){7-7}   \cmidrule(r){8-8}   \cmidrule(r){9-9}            & LIP   & AIC   & CrowdHuman & RAPv2 & CUHK-PEDES & NTU60-XSub & 3DPW  & 3DPW  &  \\
 \cmidrule(r){2-2}   \cmidrule(r){3-3}   \cmidrule(r){4-4}   \cmidrule(r){5-5}  \cmidrule(r){6-6}   \cmidrule(r){7-7}   \cmidrule(r){8-8}   \cmidrule(r){9-9}    & mIoU  & AP    & AP    & mA    & B@4   & Acc   & 100-MPVPE & 100-MPJPE &  \\
    \midrule
    (a) Hulk & 56.8  & 25.5  & 81.8  & 77.9  & 28.0  & 93.2  & -0.3  & 13.2  & 47.0  \\
        \midrule
    \multicolumn{10}{l}{Different attention masks}    \\
    \midrule
    (b) full-attn & 57.2  & 25.4  & 79.9  & 77.6  & 28.6  & 91.8  & -1.0  & 13.4  & 46.6  \\
    (c) diag-attn & 56.6  & 25.6  & 80.3  & 78.0  & 28.4  & 93.1  & -4.1  & 10.0  & 46.0  \\
    \midrule
    \multicolumn{10}{l}{Different initialization parameters}    \\
    \midrule
        (d) HAP~\cite{yuan2023hap} & 63.3  & 33.4  & 85.4  & 81.8  & 29.6  & 91.5  & 9.6  & 22.9    & 52.2  \\
    (e) PATH~\cite{tang2023humanbench} & 63.6  & 34.5  & 88.6  & 80.2  & 29.9  & 92.8  & 8.3  & 21.3    & 52.4  \\
    \bottomrule
    \end{tabular}%
    \vspace{-0.5em}
\end{table*}%

\begin{table*}[tbp]
  \centering
  \caption{Performance of Hulk on diverse human-centric tasks when scaling up data size. Experiments are both trained with a \textbf{FULL 60,000-iteration} schedule, showing that learn from additional data and achieve better performance on average. }
    \begin{tabular}{lccccccccc}
    \toprule
    \multirow{3}[6]{*}{Data samples} & Parsing & 2D Pose & Detection & Attribute & Caption & Skeleton & Mesh  & 3D Pose & \multirow{3}[6]{*}{Avg.} \\
  \cmidrule(r){2-2}   \cmidrule(r){3-3}   \cmidrule(r){4-4}   \cmidrule(r){5-5}   \cmidrule(r){6-6}   \cmidrule(r){7-7}   \cmidrule(r){8-8}   \cmidrule(r){9-9}          & LIP   & AIC   & CrowdHuman & RAPv2 & CUHK-PEDES & NTU60-XSub & 3DPW  & 3DPW  &  \\
 \cmidrule(r){2-2}   \cmidrule(r){3-3}   \cmidrule(r){4-4}   \cmidrule(r){5-5}   \cmidrule(r){6-6}   \cmidrule(r){7-7}   \cmidrule(r){8-8}   \cmidrule(r){9-9}        & mIoU  & AP    & AP    & mA    & {B@4} & Acc   & 100-MPVPE & 100-MPJPE &  \\
    \midrule
    1.7M  & 63.5  & 32.2  & 89.1  & 80.8  & 29.0  & 93.7  & -5.8  & 8.5   & 48.9  \\
    30M   & 64.0  & 34.5  & 90.7  & 80.9  & 31.1  & 93.8  & 20.2  & 33.0  & 56.0  \\
    \bottomrule
    \end{tabular}%
  \label{tab:data scale}%
  \vspace{-1.25em}
\end{table*}%

\subsubsection{Scaling up data size}
To assess the impact of the diverse and sufficient human-centric data, we train Hulk using a longer 60000-iteration schedule on the ablation datasets comprising 1.7M training samples, and full datasets comprising 30M training samples, respectively. Results in Table~\ref{tab:data scale} indicate that incorporating more human-centric data brings considerable performance improvement across all tasks, with notable enhancements in 2D pose estimation (\textbf{+2.3\%} AP), pedestrian detection (\textbf{+1.6\%} AP), image caption (\textbf{+2.1} B@4), mesh recovery (\textbf{-26.0} MPVPE) and 3D pose estimation (\textbf{-24.5} MPJPE). Slight improvements are also noted in attribute recognition and skeleton-based action recognition. We attribute these modest gains to the lower quality of data and the significant domain gap in the extended datasets. For instance, in attribute recognition, we include LU-Person~\cite{fu2021unsupervised} with its pseudo labels that offers less informative supervision compared to other training datasets. Similarly, the Kinetics-400 dataset, which has a substantial domain gap from NTU60, influences the results for the skeleton-based action recognition.

\subsubsection{Different initialization weights}
To examine the influence of model initialization on performance, we conduct additional experiments using pre-training parameters from the human-centric models, \emph{i.e.,} HAP\cite{yuan2023hap} and PATH\cite{tang2023humanbench}, and compare them with our baseline Hulk initialized by MAE~\cite{he2022masked}. 
The results, detailed in Table~\ref{tab:abl_attn}{(d-e)}, indicate that initializing Hulk with parameters from human-centric pretraining methods can further enhance its accuracy. This result not only validates the robustness of Hulk but also highlights its potential for further enhancement. Moreover, it is important to note that for a fair comparison with other human-centric models, we refrain from using these specialized pretraining parameters as our default initialization method in main results, ensuring the improvement observed in Hulk are 
attributed to its novel architecture and learning strategy.

\begin{figure}[t]
    \centering
    \includegraphics[width=\linewidth]{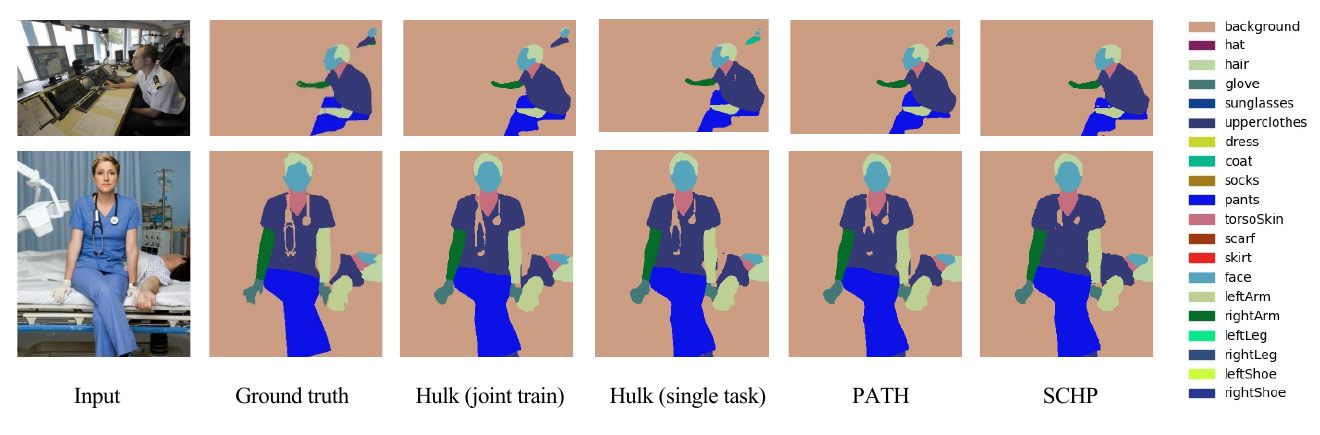}
    \vspace{-2.em}
    \caption{{Qualitative comparison of Hulk and other methods on human parsing.}}
    \label{fig:par}
    \vspace{-1em}
\end{figure}

\begin{figure}[t]
    \centering
    \includegraphics[width=.98\linewidth]{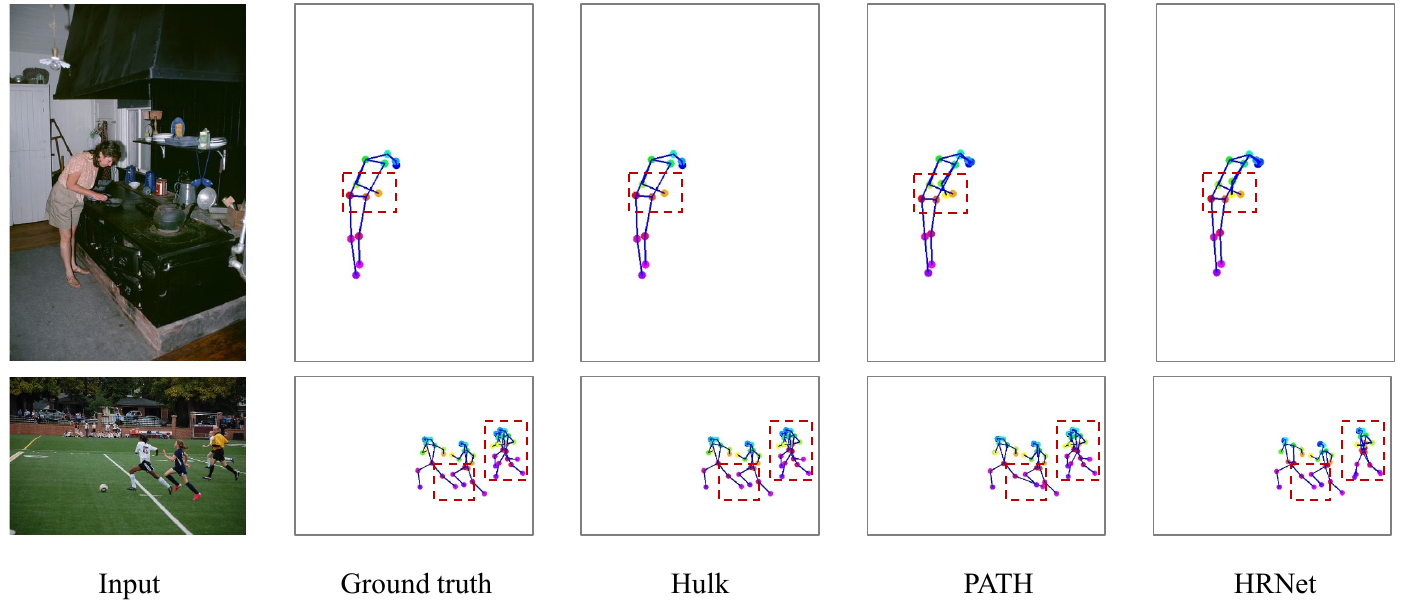}
    \vspace{-0.5em}
    \caption{{Qualitative comparison on 2D pose estimation.}}
    \label{fig:2dpose}
    \vspace{-1.7em}
\end{figure}

\subsection{{Visualization}}
{
Fig.~\ref{fig:par} and Fig.~\ref{fig:2dpose} show qualitative comparisons of Hulk with other SOTA methods on human parsing and 2D pose estimation, respectively. Hulk achieves more precise predictions, especially on detailed regions. More visualization can be found in the Supplementary Material.
}

\section{Conclusion and Limitations}
In this paper, we introduce Hulk, a novel generalist human-centric perceiver that marks the first integration of {common} 2D and 3D vision tasks, skeleton-based models, and vision-language tasks into a unified framework.  Diverging from traditional models that rely on task-specific heads, Hulk adopts an innovative approach by decomposing these into two basic heads, thereby enhancing adaptability and simplifying the process. Trained on extensive datasets, Hulk not only achieves state-of-the-art performance across various tasks but does so without the need for dataset-specific fine-tuning. Beyond the success in human-centric tasks, we also shed light on the essential components for developing a human-centric foundation model. Hulk represents a significant step for human-centric foundation models, offering a versatile and effective framework that paves the way for advanced research.

\noindent\textbf{Limitations.} Currently, the scope and the capabilities of Hulk are limited by resource constraints and training costs, preventing further exploration of other human-centric tasks, \emph{e.g.}, 3D human generation. Overcoming these barriers remains a lot of future work and will enable us to fully realize the theoretical potential of Hulk. Despite the limitations, Hulk stands as a pioneering framework in human-centric foundation models, offering a glimpse into the future where such models are more inclusive and efficient. The potential negative impacts are discussed in the supplemental material.

\section*{Acknowledgement}
This work was supported by the JC STEM Lab of AI for Science and Engineering, funded by The Hong Kong Jockey Club Charities Trust, the Research Grants Council of Hong Kong (Project No. CUHK14213224). This work was also partially supported by the National Key R$\&$D Program of China(NO.2022ZD0160101), National Natural Science Foundation of China (No.62127803), and Key R$\&$D Project of Zhejiang Province (No.2022C01056).

{\small
\bibliographystyle{unsrt}
\bibliography{bib}
}

\begin{IEEEbiography}
[{\includegraphics[width=1in,height=1.25in,clip,keepaspectratio]{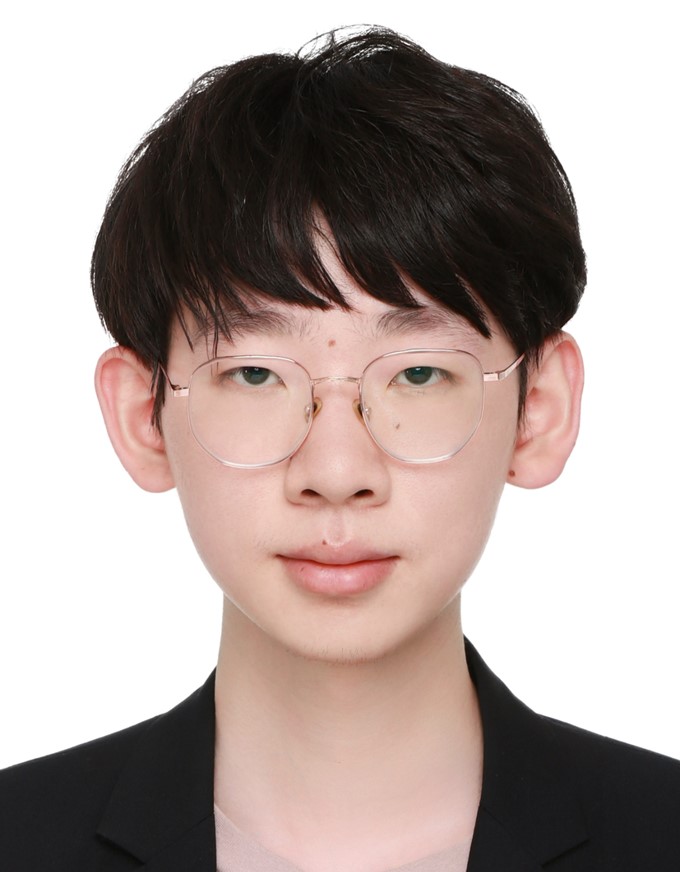}}]{Yizhou Wang}
is a researcher intern at Shanghai Artificial Intelligence Laboratory. He received Master's degree in Control Theory and Control Engineering from Zhejiang University in 2023 and Bachelor's degree in Automation from Zhejiang University in 2020. He's now pursuing his Ph.D. degree in information engineering at the Chinese University of Hong Kong. His research interests are concentrated in the areas of unsupervised learning, transfer learning, and the advancement of human-centric unified models in computer vision.
\end{IEEEbiography}

\begin{IEEEbiography}[{\includegraphics[width=1in,height=1.25in,clip,keepaspectratio]{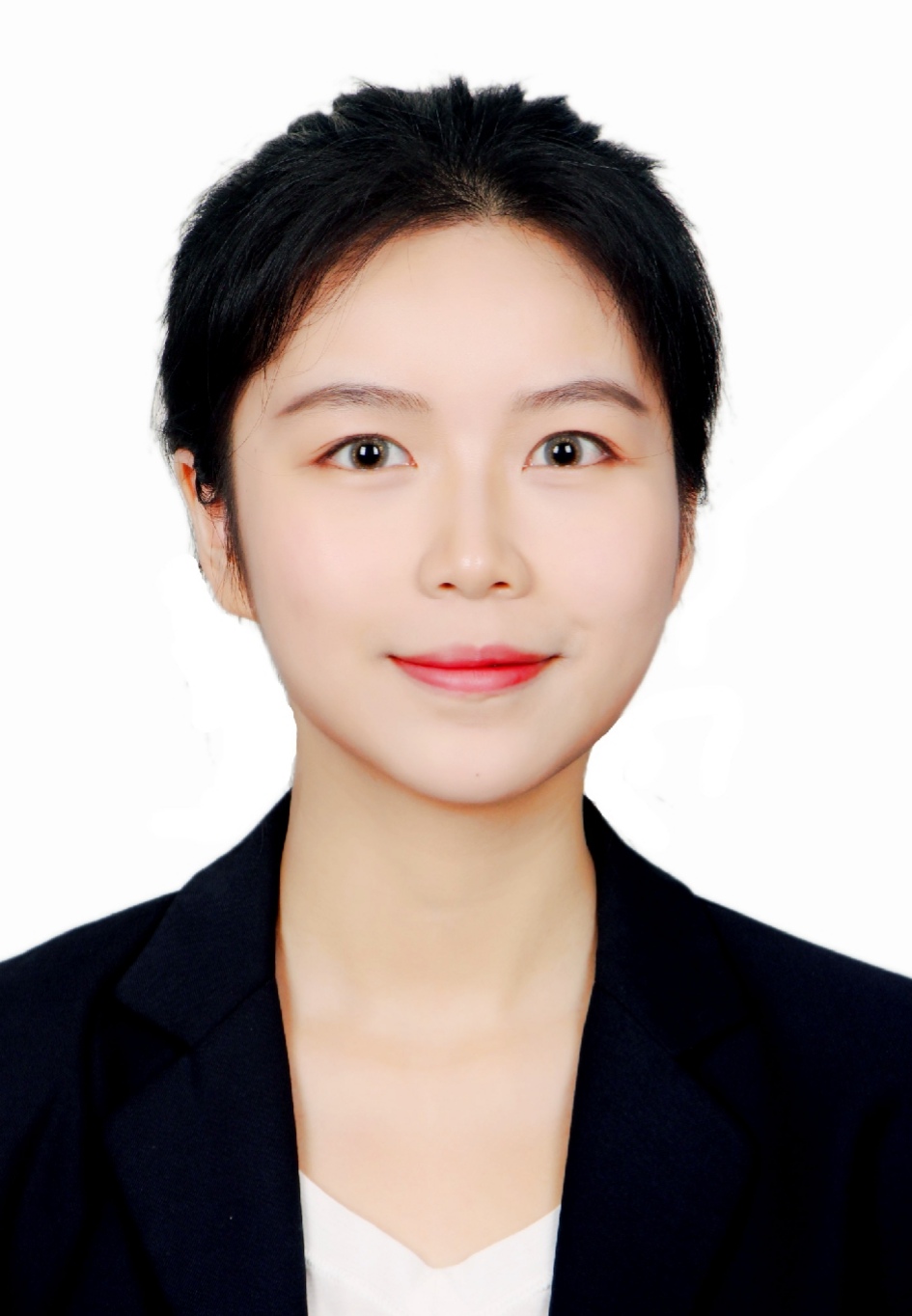}}]{Yixuan Wu} received the B.S. degree from Zhejiang University, Hangzhou, China. She is currently working toward the Ph.D. degree in the major of big data and health science in Zhejiang University, Hangzhou, China. Her research interests include deep learning, multimodal learning and computer vision.
\end{IEEEbiography}

\begin{IEEEbiography}[{\includegraphics[width=1in,height=1.25in,clip,keepaspectratio]{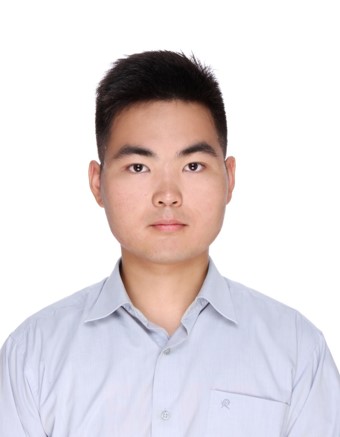}}]{Weizhen He} received the B.S.degree in College of Electrical Engineering from Zhejiang University, Hangzhou, China, in 2021. He is currently working toward the Ph.D. degree in the major of Control Theory and Control Engineering in Zhejiang University. His research interests include human-centric artificial intelligence, person re-identification and object detection.

\end{IEEEbiography}

\begin{IEEEbiography}[{\includegraphics[width=1in,height=1.1in,clip,keepaspectratio]{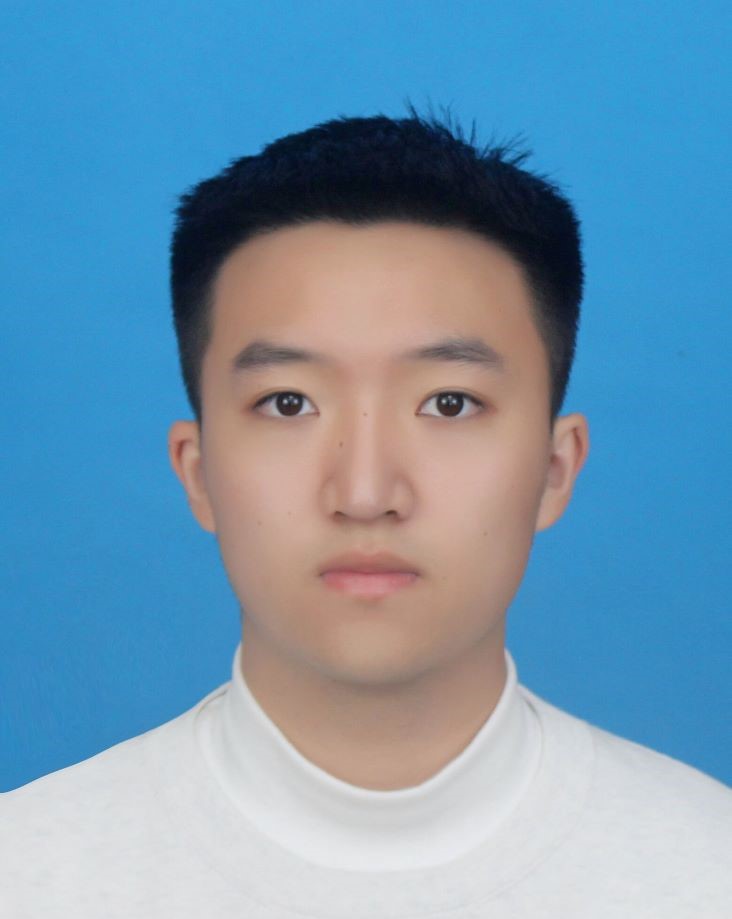}}]{Xun Guo} is now jointly trained by the University of Science and Technology of China and the Institute of Automation, Chinese Academy of Science, where he pursuing the master degree in the major of Control Science and Engineering. His research interests include multimodal learning and generative models.

\end{IEEEbiography}

\begin{IEEEbiography}[{\includegraphics[width=1in,height=1.25in,clip,keepaspectratio]{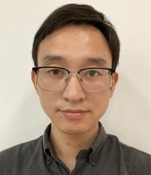}}]{Feng Zhu} is currently a Research Director in Smart
City Group of SenseTime, leading a talented research development team with over 40 full-time researchers, engineers, and interns. His team supports
SenseFoundry, a one-stop software platform tailored
for Smart City management, delivering cutting-edge
Computer Vision and Deep Learning techniques for
large-scale applications in Smart City, such as City
Safety, and Traffic Management. Feng ZHU received
a B.E. degree and Ph.D. degree in Electronic Engineering in 2011 and 2017, respectively, from the
Department of Electronic Engineering and Information Science, University of
Science and Technology of China (USTC).

\end{IEEEbiography}

\begin{IEEEbiography}[{\includegraphics[width=1in,height=1.25in,clip,keepaspectratio]{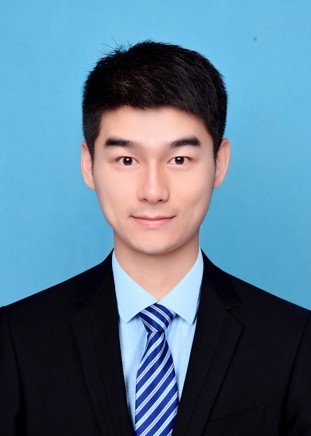}}]{Lei Bai} is a research scientist at Shanghai AI Laboratory. Prior to that, he was a Postdoctoral Research Fellow at the University of Sydney, Australia. He received his PhD degree in Computer Science from UNSW Sydney under the supervision of Prof. Lina Yao and Prof. Salil Kanhere. His research interests lay in Spatial-Temporal Generative Learning and it's applications in cross-discipline scenarios, such as climate and weather forecasting, and intelligent transportation. Dr. Bai has published more than 60 papers in top-tier conferences and journals, including Nature Communications, IEEE TPAMI, IEEE TIP, NeurIPS, CVPR, IJCAI, and KDD.

\end{IEEEbiography}

\begin{IEEEbiography}[{\includegraphics[width=1in,height=1.25in,clip,keepaspectratio]{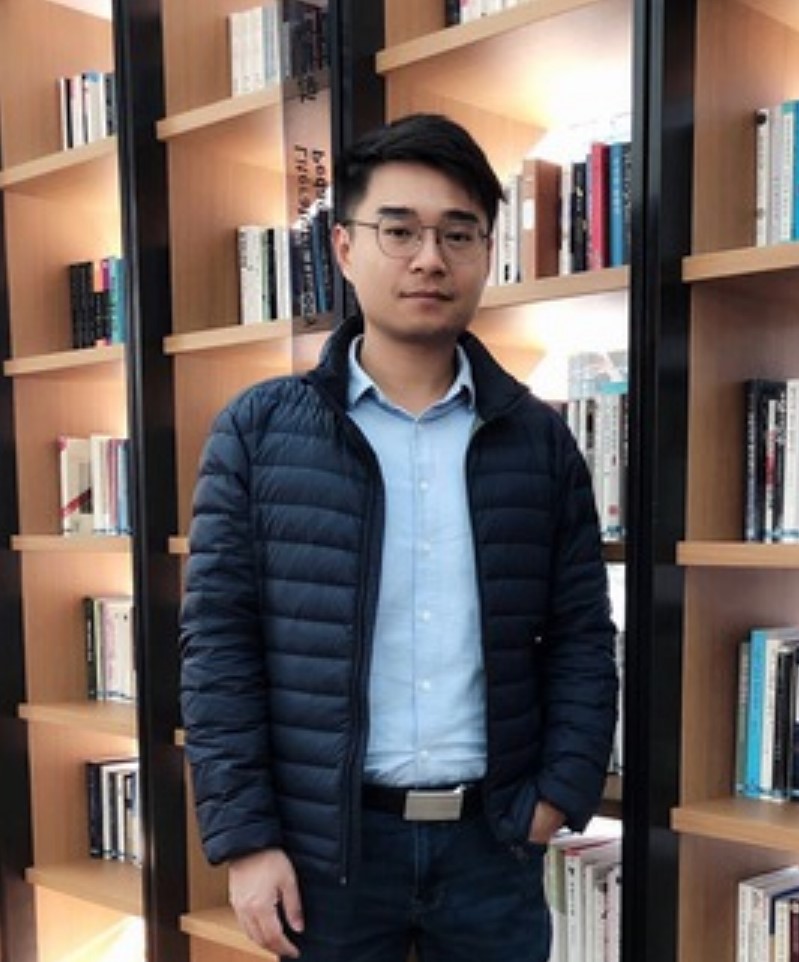}}]{Rui Zhao} is the Research Executive in the Smart
City Group of SenseTime. He is also the lead
researcher at Qing Yuan Research Institute of Shanghai Jiaotong University. He obtained Ph.D. from
the Chinese University of Hong Kong in 2015.
His research interest is computer vision and deep
learning. He has more than 100 publications, with
around 10,000 citations.

\end{IEEEbiography}

\begin{IEEEbiography}[{\includegraphics[width=1in,height=1.25in,clip,keepaspectratio]{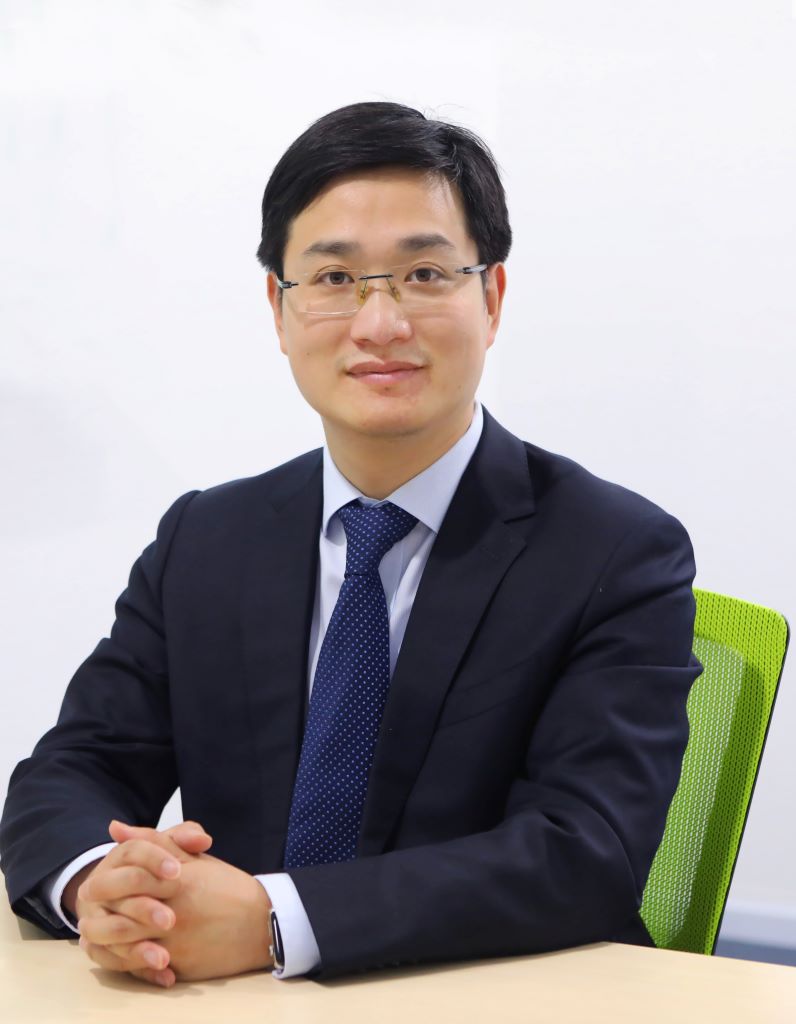}}]{Jian Wu} received the Ph.D. degree in Computer Science and Technology from Zhejiang University in 1998. He is an IEEE member, CFF member, CCF TCSC member, CCF TCAPP member and member of the ``151 Talent Project of Zhejiang Province''. Prof. Jian Wu is recently the director of Research Centre of Zhejiang University and Vice-president of National Research Institute of Big Data of Health and Medical Sciences of Zhejiang University. His research interests include Medical Artificial Intelligence, Service Computing and Data Mining.

\end{IEEEbiography}

\begin{IEEEbiography}[{\includegraphics[width=1in,height=1.25in,clip,keepaspectratio]{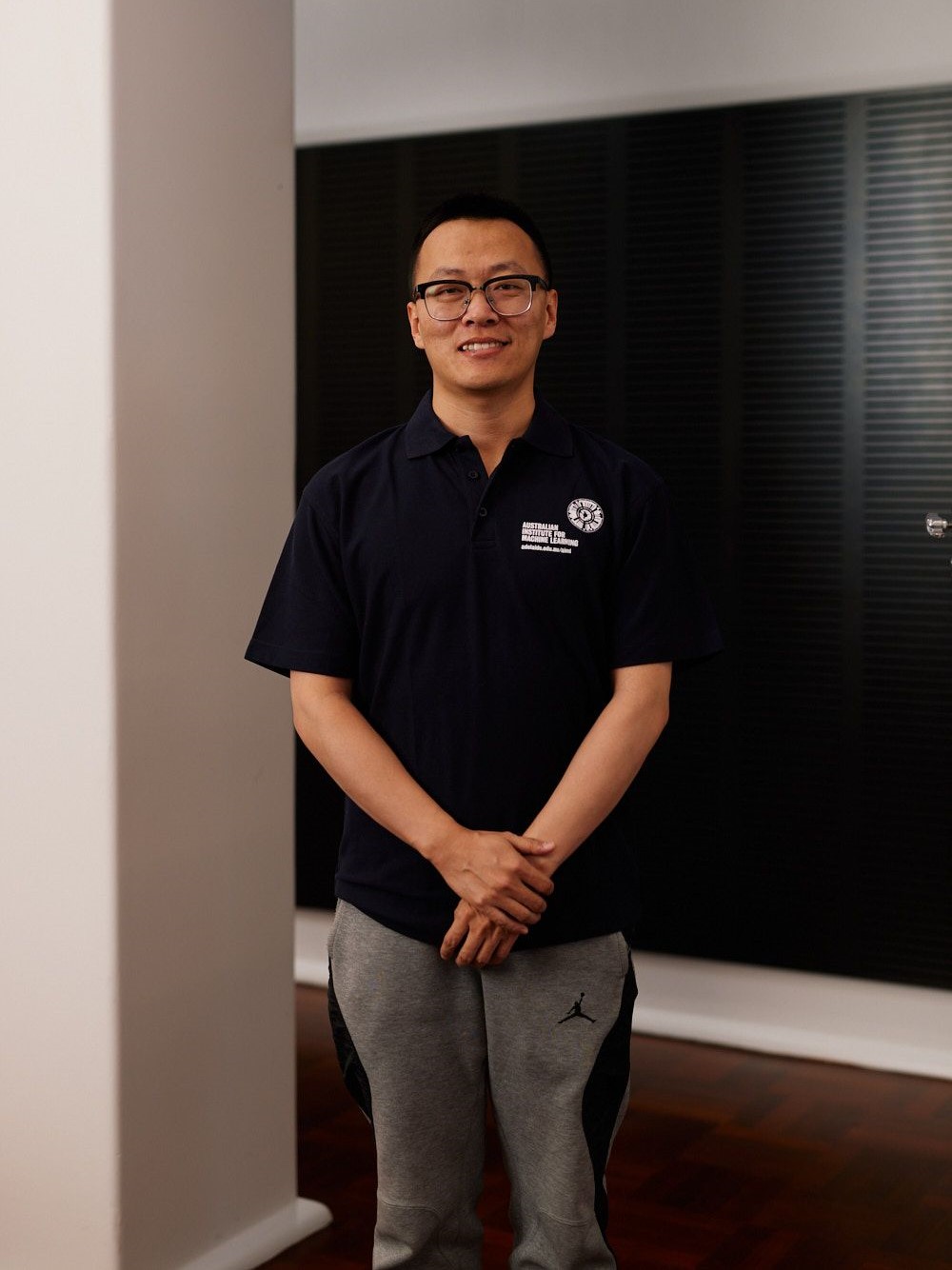}}]{Tong He} received his Ph.D. degree in computer science from the University of Adelaide, Australia, in 2020. He is currently a researcher at Shanghai AI Laboratory. His research interests include computer vision and machine learning.

\end{IEEEbiography}
\begin{IEEEbiography}[{\includegraphics[width=1in,height=1.25in,clip,keepaspectratio]{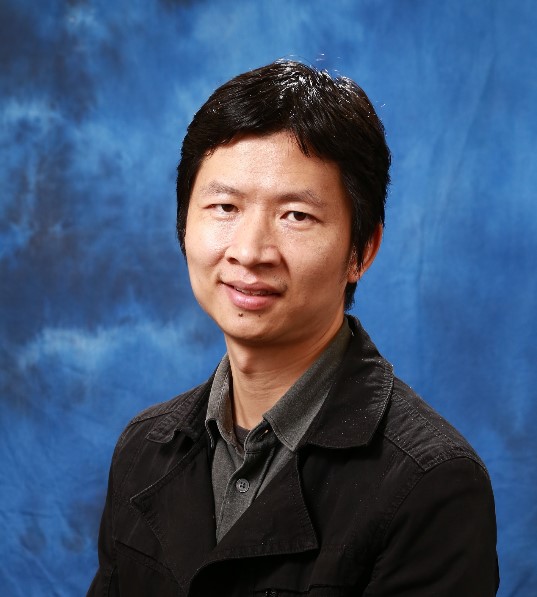}}]{Wanli Ouyang} (Senior Member, IEEE) received
the Ph.D. degree from the Department of Electronic
Engineering, The Chinese University of Hong Kong. He is currently a professor at the Chinese University of Hong Kong and Shanghai Artificial Intelligence Laboratory. His research interests include pattern recognition, machine learning, and AI for Science. 
His research interests include deep learning and its
application to computer vision and pattern recognition, image, and video processing. He was an associate editor of IEEE Transactions on Pattern Analysis and Machine Intelligence, International Journal of Computer Vision, and Public Relations, the Senior Area Chair of CVPR.

\end{IEEEbiography}

\begin{IEEEbiography}[{\includegraphics[width=1in,height=1.25in,clip,keepaspectratio]{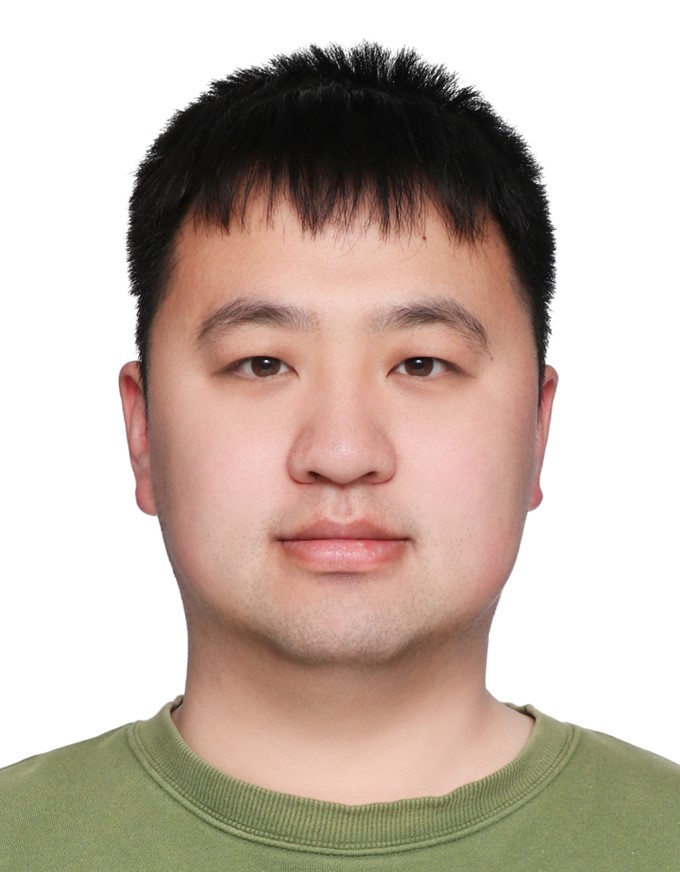}}]{Shixiang Tang} received the Ph.D degree from the University of Sydney. Prior to that, he received the Master of Philosophy from the Chinese University of Hong Kong in 2018 and Bachelor of Science from Fudan University. His interests lie in machine learning and computer vision, especially self-supervised learning and foundation models. He has published about 10 papers in top-tier conferences and journals, e.g., CVPR, ICCV, NeurIPS, Nature Physics and Nature Materials.
\vspace{-2em}
\end{IEEEbiography}

\newpage

 




\newpage

\vfill
\clearpage
\appendices
\section{Full Ablation results}
As shown in Table~\ref{tab:abl}, we present the full experimental results of the ablation studies. Specifically, we explore the effects of (1) different weight-sharing strategies, (2) attention mask designs in the decoder, (3) task collaboration and interference, and (4) different initialization weights. One can see that the same conclusion can be observed that as more model components are shared, Hulk can achieve higher performance. Besides, omitting any task
generally leads to reduced performance. Furthermore, employing different attention interaction methods for
different tasks can optimize the model’s performance, compared with using pure full attention or diagonal attention. Last, using the pre-trained weights from other human-centric models (i.e., HAP~\cite{yuan2023hap} and PATH~\cite{tang2023humanbench}) can further boost our hulk's performance. This result not only validates
the robustness of Hulk but also highlights its potential for
further enhancement.

Moreover, in Table~\ref{tab:abl2}, to assess the impact of the diverse and sufficient human-centric data, we train Hulk using a 60000-iteration schedule on the ablation datasets comprising 1.7M training samples, and full datasets comprising 30M training samples, respectively. Results indicate that incorporating more human-centric data and increasing the training iterations brings considerable performance improvement across all tasks.

\begin{table*}[htbp]
  \centering
  \caption{Complete experimental results of different ablation studies. The average performances in the ablation of task collaboration and interference are not computed and replaced by N/A.}
      \resizebox{\textwidth}{34mm}{
    \begin{tabular}{lcccccccccccccccc}
    \toprule
    \multicolumn{1}{c}{\multirow{3}[6]{*}{Methods}} & \multicolumn{3}{c}{Parsing} & \multicolumn{2}{c}{2D Pose} & Detection & \multicolumn{2}{c}{Attribute} & Caption & Skeleton & \multicolumn{5}{c}{3D Pose \& Mesh}    & \multirow{3}[6]{*}{Avg.} \\
  \cmidrule(r){2-4}   \cmidrule(r){5-6}   \cmidrule(r){7-7}   \cmidrule(r){8-9}   \cmidrule(r){10-10}   \cmidrule(r){11-11}   \cmidrule(r){12-16}             & H3.6  & LIP   & CIHP  & COCO & AIC   & Crowd & PA & rapv2 & CUHK  & ntu60 & \multicolumn{2}{c}{Human3.6M} & \multicolumn{3}{c}{3DPW} &  \\
     \cmidrule(r){2-4}   \cmidrule(r){5-6}   \cmidrule(r){7-7}   \cmidrule(r){8-9}   \cmidrule(r){10-10}   \cmidrule(r){11-11}   \cmidrule(r){12-16}          & mIoU  & mIoU  & mIoU  & AP    & AP    & AP    & mA    & mA    & B@4   & Acc   & 100-MPJPE & 100-PA-MPJPE & 100-MPVPE & 100-MPJPE & 100-PA-MPJPE &  \\
    \midrule
    (a) baseline (Hulk) & 62.4  & 56.8  & 59.7  & 70.8  & 25.5  & 81.8  & 79.4  & 77.9  & 28.0  & 93.2  &  45.9  & 60.5  & -0.3  & 13.2  & 49.0    & 53.6  \\
        \midrule
       \multicolumn{3}{l}{\textit{Weight sharing}} \\
        \midrule
    (b) only enc./dec. shared& 63.4  & 56.8  & 59.6  & 71.1  & 25.8  & 80.6  & 80.8  & 78.8  & 26.7  & 93.8  & 45.2  & 60.3  & -3.0  & 10.7  & 47.0  & 53.2  \\
    (c) only enc. shared & 51.6  & 56.3  & 60.5  & 71.3  & 25.7  & 75.9  & 82.2  & 81.3  & 28.7  & 1.7   & 35.9  & 56.5  & -11.2  & 2.9   & 41.3  & 44.0  \\
        \midrule
       \multicolumn{3}{l}{\textit{Attention Mask Designs in the Decoder}}\\
        \midrule
     full-attn & 62.6  & 57.2  & 59.9  & 70.9  & 25.4  & 79.9  & 79.2  & 77.6  & 28.6  & 91.8  &44.3  & 60.1  & -1.0  & 13.4  & 48.1   & 53.2  \\
     diag-attn & 62.7  & 56.6  & 59.9  & 70.6  & 25.6  & 80.3  & 79.3  & 78.0  & 28.4  & 93.1  & 40.3  & 57.6  & -4.1  & 10.0  & 46.3    & 52.3  \\
        \midrule
      \multicolumn{3}{l}{\textit{Task Collaboration and Interference}} \\
        \midrule
    w/o attribute & 62.4  & 56.9  & 59.9  & 70.7  & 25.6  & 79.9  &   -    &  -     & 28.2  & 93.1  &44.4  & 60.0  & -2.1  & 12.2  & 47.9  & N/A  \\
    w/o caption & 62.7  & 57.3  & 59.9  & 70.7  & 25.6  & 80.2  & 79.3  & 77.8  & -      & 92.9  & 46.5  & 61.3  & -2.2  & 11.2  & 47.5    & N/A  \\
    w/o skeleton & 62.3  & 57.4  & 60.1  & 70.7  & 25.7  & 80.9  & 79.1  & 77.4  & 28.2  & -    &45.7  & 59.6  & -4.8  & 10.0  & 46.7    & N/A  \\
    w/o mesh\&3d pose & 62.6  & 57.4  & 60.3  & 70.8  & 25.5  & 81.0  & 79.8  & 78.0  & 27.9  & 93.1  &  -     &  -     &    -   &  -     &  -     & N/A  \\
    w/o detection & 63.2  & 57.6  & 60.9  & 71.4  & 26.4  &  -     & 79.9  & 77.6  & 27.8  & 93.1  & 46.0  & 59.8  & -4.8  & 10.5  & 47.6  & N/A  \\
    w/o 2d pose & 56.6  & 52.3  & 53.0  &  -     & -      & 65.0  & 75.3  & 75.1  & 26.6  & 93.5  & 24.5  & 48.4  & -18.9  & -3.5  & 37.2   & N/A  \\
    w/o parsing &  -     &  -     &  -     & 70.2  & 25.2  & 78.5  & 79.4  & 77.3  & 25.8  & 93.2  & 42.5  & 57.5  & -2.9  & 10.8  & 46.7    & N/A  \\
        \midrule
       \multicolumn{3}{l}{\textit{Different initialization weights}} \\
        \midrule
    MAE (baseline)   & 62.4  & 56.8  & 59.7  & 70.8  & 25.5  & 81.8  & 79.4  & 77.9  & 28.0  & 93.2  & 45.9  & 60.5  & -0.3  & 13.2  & 49.0   & 53.6  \\
    HAP   & 65.8  & 63.3  & 68.8  & 76.5  & 33.4  & 85.4  & 85.5  & 81.8  & 29.6  & 91.5  & 55.2  & 66.6  & 9.6   & 22.9  & 55.6   & 59.4  \\
    PATH & 66.3  & 63.6  & 69.2  & 76.8  & 34.5  & 88.6  & 83.3  & 80.2  & 29.9  & 92.8  & 54.6  & 66.7  & 8.3   & 21.3  & 54.5    & 59.4  \\
    \bottomrule
    \end{tabular}}
  \label{tab:abl}%
\end{table*}%

\begin{table*}[htbp]
  \centering
  \caption{Complete experimental results of our Hulk performance on diverse human-centric tasks when scaling up data size. Experiments are both trained with a \textbf{FULL 60,000-iteration} schedule, showing that learn from additional data and achieve better performance on average.}
      \resizebox{\textwidth}{10.8mm}{
    \begin{tabular}{lcccccccccccccccc}
    \toprule
    \multicolumn{1}{c}{\multirow{3}[6]{*}{Methods}} & \multicolumn{3}{c}{Parsing} & \multicolumn{2}{c}{2D Pose} & Detection & \multicolumn{2}{c}{Attribute} & Caption & Skeleton & \multicolumn{5}{c}{3D Pose \& Mesh}    & \multirow{3}[6]{*}{Avg} \\
  \cmidrule(r){2-4}   \cmidrule(r){5-6}   \cmidrule(r){7-7}   \cmidrule(r){8-9}   \cmidrule(r){10-10}   \cmidrule(r){11-11}   \cmidrule(r){12-16}            & H3.6  & LIP   & CIHP  & COCO & AIC   & Crowd & PA & rapv2 & CUHK  & ntu60 & \multicolumn{2}{c}{Human3.6M} & \multicolumn{3}{c}{3DPW} &  \\
  \cmidrule(r){2-4}   \cmidrule(r){5-6}   \cmidrule(r){7-7}   \cmidrule(r){8-9}   \cmidrule(r){10-10}   \cmidrule(r){11-11}   \cmidrule(r){12-16}           & mIoU  & mIoU  & mIoU  & AP    & AP    & AP    & mA    & mA    & B@4   & Acc   & 100-MPJPE & 100-PA-MPJPE & 100-MPVPE & 100-MPJPE & 100-PA-MPJPE &  \\
    \midrule
    30M   & 68.1  & 64.0  & 70.6  & 77.0  & 34.5  & 90.7  & 82.9  & 80.9  & 31.1  & 93.8  & 56.4  & 68.1  & 20.2  & 33.0  & 60.1   & 62.1  \\
    1.7M  & 66.9  & 63.5  & 68.4  & 76.0  & 32.2  & 89.1  & 82.1  & 80.8  & 29.0  & 93.7  & 36.0  & 60.3  & -5.8  & 8.5   & 44.9   & 55.0  \\
    \bottomrule
    \end{tabular}}
  \label{tab:abl2}%
\end{table*}%

\section{Additional Architecture Details}
\subsection{Modality indicator}
Modality indicators $\mathcal{I}_{m'}$ are learnable tokens that are appended with encoded tokens $\mathbf{p}$ to generate the output tokens $\mathbf{q}$ through the modality-shared decoder. In the decoder, modality indicators $\mathcal{I}_{m'}$ have a default shape of $\mathcal{R}^{1 \times d_{out}}$, where $d_{out}$ is the dimension of output tokens. For a certain task that requires $N'$ output tokens, we repeat the modality indicator $N'$ times, resulting in a shape of $\mathcal{R}^{N' \times d_{out}}$.

\subsubsection{Weight sharing.} As an indicator, modality indicators $\mathcal{I}_{m'}$ are shared within the same modality but across different tasks, \emph{e.g.}, the dense labeling modality indicator $\mathcal{I}_{D}$ is utilized for both human parsing task and 2D pose estimation task. This design together with the modality-shared encoder and decoder helps Hulk learn human-centric knowledge, remedying task bias from a certain task.

\subsubsection{Shape of modality indicators.}  While the aligning text features $\mathbf{v}\in \mathcal{R}^{C\times d_{out}}$ always have a length of $C$ denoting the number of semantics/classes, $N'$ varies among different human-centric tasks. For \textbf{action and attribute recognition tasks}, $N'=C$ where $C$ is the number of action or attribute classes. After aligning with BERT semantic features $\mathbf{v}$, the similarity in the $i$-th index represents the existence probability of the $i$-th action/attribute. For the \textbf{image caption task},  $N'=40$ means the length of caption tokens. For dense label output tasks, \emph{i.e.}, \textbf{human parsing and 2D pose estimation tasks}, $N'$ equals the number of input image tokens $N$, generating predicted semantic maps with the same resolution. For sparse label output tasks, \emph{i.e.}, \textbf{pedestrian detection, 3D pose estimation, and mesh recovery task}, $N'$ equals the number of predicted boxes/joints/vertices. 

\subsection{Positional Embeddings}
\subsubsection{Positional embeddings in the encoder}
Different from UniHCP~\cite{ci2023unihcp}, we do not use a learnable positional embedding, instead, we use the bilinear interpolation of fixed MAE~\cite{he2022masked} positional embeddings in the encoder.

\subsubsection{Positional embeddings in the decoder}
As the inputs for the decoder contain two parts, \emph{i.e.}, encoded tokens $\mathbf{p}$ and modality indicator $\mathcal{I}_{m'}$, therefore, positional embeddings in the decoder have two parts. 

For the encoded tokens part, positional embeddings are set as the bilinear interpolation of fixed MAE positional embeddings. 

For the modality indicator part, we prepare three types of positional embeddings for better experimental results across diverse human-centric tasks:

 \begin{itemize}
     \item \textbf{Anchor points}. For the pedestrian detection task, we follow anchor DETR~\cite{wang2022anchor} to integrate the location information into modality indicators for achieving better performance. To ensure compatibility with the MAE positional embeddings, which have a shape of $14\times14$, used in the encoded tokens part, we generate $17\times17$ positional embeddings as fixed anchor points. This approach enables Hulk to focus on regressing the distances between these anchor points and the actual ground truth boxes. By doing so, the detection task is simplified, effectively reducing the complexity and difficulty of convergence.
     \item \textbf{3D positional embeddings}. For 3D pose estimation and mesh recovery tasks, Hulk needs to predict 3D digit locations using 3D positional embeddings. Following the practice in~\cite{feichtenhofer2022masked}, we separate the 3D positional embeddings into a fixed $2D$ positional embedding and a fixed $1D$ positional embedding with $PE_{3D}=PE_{2D}+PE_{1D}$, introducing strong 3D location information into 3D regression tasks. The $2D$ positional embedding and $1D$ positional embedding can be generated by interpolating from fixed MAE positional embeddings.
     \item \textbf{Simple interpolation}. For tasks encompassing 2D vision, vision-language, and skeleton-based applications, we employ a straightforward approach to generate the positional embeddings in the modality indicator part. These embeddings are inherently two-dimensional or one-dimensional and are derived through a process of simple interpolation from the fixed MAE positional embeddings. 
 \end{itemize}

\subsection{Objective function details}
In this section, we present the objective functions in Sec.~IV.D in the main text. Given the prediction $\hat{\mathbf{y}}_m'$ where $m'\in\{I, T, S, D\}$ is the output modality, we compute the losses according to different human-centric tasks. As modality-specific de-tokenizers transform output tokens to the semantic part and the digital location part, we adopt two types of losses, \emph{i.e.}, semantic contrastive loss and digit regression loss. 

For \textbf{semantic constrasitve loss}, we first compute the conv-transformed output tokens $\hat{\mathbf{f}}\in \mathbb{R}^{N'\times d_{out}}$ with $\hat{\mathbf{f}} =\text{Conv}(\mathbf{q})$, where $N'$ is the number of output tokens, $d_{out}$ is the output dimension, and $\mathbf{q}\in \mathbb{R}^{N'\times d_{out}}$ are output tokens. Given $\mathbf{v}=(\mathbf{v}_1, \mathbf{v}_2, ..., \mathbf{v}_C)$ containing the BERT features of $C$ ground truth classes, the semantic contrastive loss is defined as
\begin{equation}
    \mathcal{L}^s = \frac{\mathbf{v}_k^\top \hat{\mathbf{f}}}{\mathbf{v}_1^\top \hat{\mathbf{f}} + \mathbf{v}_2^\top \hat{\mathbf{f}} + ... + \mathbf{v}_C^\top \hat{\mathbf{f}}},
\end{equation}
where $k$ is the index of the selected semantic token.

For \textbf{digit regression loss}, we compute the distance between predicted coordinate values $\hat{\mathbf{y}}^d$ and corresponding ground truth coordinate values $\mathbf{y}^d$ with
\begin{equation}
    \mathcal{L}^d = Dis(\hat{\mathbf{y}}^d, \mathbf{y}^d),
\end{equation}
where $Dis(\mathbf{a}, \mathbf{b})$ measures the distance between $\mathbf{a}$ and $\mathbf{b}$, which is different in different tasks, \emph{i.e.}, L1 loss and GIoU loss in the pedestrian detection.

\noindent\textbf{Human Parsing.}
Human parsing aims to segment human parts. As human parts can be represented by their BERT features, we align the output tokens with BERT features of semantic classes using per-pixel contrastive loss. Following the practice in UniHCP~\cite{ci2023unihcp}, we also supervise the global probability of a certain semantic class to enhance the performance. 
Given the conv-transformed output tokens $\hat{\mathbf{f}}$, we simply average pool them to represent the predicted probability of occurrence of certain semantic class $\hat{\mathbf{f}}_{avg}$. Given extracted BERT features $\mathbf{v}$ of $C$ parsing classes,
, the total objective function for human parsing is as follows:
\begin{equation}
\begin{aligned}
     \mathcal{L}_{par} &= \mathcal{L}^{s}(\hat{\mathbf{f}}, \mathbf{v}_{pix}) + \mathcal{L}^{s}(\hat{\mathbf{f}}_{avg}, \mathbf{v}) + \mathcal{L}_{dice}(\hat{\mathbf{y}}, y), 
\end{aligned}
\end{equation}
where $\mathbf{v}_{pix}\in \mathbb{R}^{N\times C\times d_{out}}$ denotes the per-pixel BERT features, $L_{dice}$ is adopted as an auxiliary loss to enhance the human parsing performance. In the dice loss, $y \in \mathbb{R}^{N\times C}$ denotes the ground truth heatmaps with $C$ parsing classes and $\hat{\mathbf{y}}_D\in \mathbb{R}^{N\times C}$ is the predicted heatmaps computed by $\hat{\textbf{y}} = \{\hat{y}_j  = \text{argmax}_{k\in [1, C]} \mathbf{v}^\top\text{Upsample}( \hat{\mathbf{f_{j}}}) \}, j\in [1, N']$. 

\noindent\textbf{2D Pose Estimation.} We follow the common top-down setting for 2D pose estimation. As we leverage the heatmap-based methods for 2D pose estimation for better experimental results, the objective function is similar to that in human parsing. Mathematically, given extracted BERT features $\mathbf{v}$ of $C$ joint names, we have
\begin{equation}
    \mathcal{L}_{pose} = \mathcal{L}^{s}(\hat{\mathbf{f}}, \mathbf{v}_{pix}) + \mathcal{L}^{s}(\hat{\mathbf{f}}_{avg}, \mathbf{v}),
\end{equation}
where $\mathbf{v}_{pix}\in \mathbb{R}^{N\times C\times d_{out}}$ denotes the per-pixel BERT features.

\noindent\textbf{Pedestrian Attribute Recognition.} Pedestrian Attribute Recognition predicts whether an attribute exists in the input image. Therefore, we only supervise the output tokens $\hat{\mathbf{f}}\in \mathbb{R}^{C\times d_{out}}$ by aligning them to $C$ different attribute semantics. High similarity means the existence of a certain attribute. Given extracted attribute BERT features $\mathbf{v}\in \mathbb{R}^{C\times d_{out}}$, the objective function can be computed by:
\begin{equation}
    \mathcal{L}_{attr} = \mathcal{L}^{s}(\hat{\mathbf{f}}, \mathbf{v}).
\end{equation}

\noindent\textbf{Pedestrian Image Caption.} We predict the image caption in an auto-regressive manner. For each predicted token, we adopt the semantic constrastive loss $\mathcal{L}^s$ to supervise. The total objective function is similar to that in attribute recognition,
\begin{equation}
    \mathcal{L}_{cap} = \mathcal{L}^{s}(\hat{\mathbf{f}}, \mathbf{v}),
\end{equation}
where $\hat{\mathbf{f}}\in \mathbb{R}^{N'\times d_{out}}$ are the features representing predicted caption tokens, $N'$ is 40 by default. $\mathbf{v}\in \mathbb{R}^{C \times d_{out}}$ denotes the features of BERT vocabulary, where $C = 30522$ denotes the length of BERT vocabulary.

\noindent\textbf{Pedestrian Detection.} Different from widely adopted designs in transformer-based pedestrian detectors that rely on object queries, we directly decode $N'$ modality indicators as the sparse prediction results. Given predicted digital location $\hat{\mathbf{y}}^d$, predicted semantic features $\hat{\mathbf{f}}$, ground truth location $\mathbf{y}^d$ and ground truth semantic features $\mathbf{v}$ representing the word ``pedestrian", we utilize optimal bipartite matching to determine the matched pairs. The objective function can be computed by the combination of semantic contrastive loss and digit regression loss:
\begin{equation}
    \mathcal{L}_{det} = \mathcal{L}_{L1}(\hat{\mathbf{y}}^d, \mathbf{y}^d) + \mathcal{L}_{iou}(\hat{\mathbf{y}}^d, \mathbf{y}^d) + \mathcal{L}^s(\hat{\mathbf{f}}, \mathbf{v}),
\end{equation}
where $\mathcal{L}_{L1}$, $\mathcal{L}_{iou}$ are L1 loss and GIoU loss, respectively. 

\noindent\textbf{3D Pose Estimation \& Mesh Recovery.} Following common practice~\cite{fastmetro,metro,li2022cliff}, we train Hulk to learn 3D pose estimation task and mesh recovery task jointly. Similar to the pedestrian detection task, the 3D pose estimation task also has a combined objective function of semantic contrastive loss and digital regression loss while the mesh recovery task only has the digit regression loss due to undefined semantics of numerous vertices. Given predicted digital location $\hat{\mathbf{y}}^d$, predicted semantic features $\hat{\mathbf{f}}$ representing $C$ different joint semantics, we compute the objective function for 3D pose estimation and mesh recovery as 
\begin{equation}
\begin{aligned}
    \mathcal{L}_{3dpos}+\mathcal{L}_{mesh}=&\mathcal{L}_{2d}(\hat{\mathbf{y}}^d, \mathbf{y}) 
    +\mathcal{L}_{3d} (\hat{\mathbf{y}}^d, \mathbf{y})\\
    &+\mathcal{L}_{vertice}(\hat{\mathbf{y}}^d, \mathbf{y}) 
    +\mathcal{L}_{normal}(\hat{\mathbf{y}}^d, \mathbf{y})\\
    &+\mathcal{L}_{edge}(\hat{\mathbf{y}}^d, \mathbf{y})+\mathcal{L}^s(\hat{\mathbf{f}}, \mathbf{v}),
\end{aligned}
\end{equation}
where $\mathbf{y}$ is the ground truth location, $\mathbf{v}$ denotes BERT features of joint names. $\mathcal{L}_{2d}$, $\mathcal{L}_{3d}$, $\mathcal{L}_{vertice}$, $\mathcal{L}_{normal}$, and $\mathcal{L}_{edge}$ are all L1 loss.

\noindent\textbf{Skeleton-based action Recognition.} Skeleton-based action recognition predicts the action based on input skeleton sequence. Therefore, given action semantic features $\mathbf{v} \in \mathbb{R}^{C\times d_{out}}$ where $C$ is the number of actions, and the predicted semantic tokens $\hat{\mathbf{f}}\in \mathbb{R}^{C\times d_{out}}$, the objective function can be computed by:
\begin{equation}
    \mathcal{L}_{ske}=\mathcal{L}^s(\hat{\mathbf{f}}, \mathbf{v}).
\end{equation}

\section{Potential Negative Impacts}
As a foundation model for human-centric tasks, Hulk's extensive data requirements and prolonged training time raise concerns about environmental impact due to high energy consumption. To mitigate this, future research should focus on improving computational efficiency. Additionally, Hulk is trained and evaluated on several human-centric datasets, thus may contain biases in these datasets, which need continuous effort to address to ensure ethical application.

\section{\textcolor{black}{Transfer to other human-centric datasets}}
\subsection{\textcolor{black}{COCO-Wholebody}}
\textcolor{black}{
To evaluate the ability of Hulk on fine-grained whole-body pose estimation task, we finetune Hulk on COCO-WholeBody V1.0 dataset~\cite{jin2020whole} following most of the default training and evaluation settings of mmpose~\cite{mmpose2020}. Table~\ref{tab:wholebody} shows the experimental results comparing Hulk with the state-of-the-art top-down methods. The results show that Hulk achieves the best whole-body pose estimation accuracy (61.2\% AP and 72.1\% AR) across different methods with diverse backbones. When compared using the same plain backbone, Hulk outperforms ViTPose~\cite{xu2022vitpose} by a large margin ($\textbf{+5.3}$ AP), demonstrating the general capability of Hulk on human-centric tasks.
}

\begin{table*}[htbp]\color{black}
  \centering
  \caption{OKS-based Average Precision (AP) and Average Recall (AR) on the COCO-WholeBody V1.0 dataset. All methods are evaluated with 256$\times$192 input images. Style: \textbf{best}, \underline{second}.}
\resizebox{\linewidth}{!}{
    \begin{tabular}{lllcccccccccc}
    \toprule
    Method &       & Backbone & Body AP & Body AR & Foot AP & Foot AR & Face AP & Face AR & Hand AP & Hand AR & Whole AP & Whole AR \\
    \midrule
    \multirow{8}[0]{*}{Specalist} & SimpleBase~\cite{xiao2018simple}& ResNet-50 & 65.2  & 73.8  & 61.5  & 74.9  & 60.6  & 71.5  & 46.0  & 58.4  & 52.1  & 63.3  \\
    & SimpleBase~\cite{xiao2018simple}& ResNet-152 & 68.2  & 76.4  & 66.1  & 78.7  & 62.3  & 72.8  & 48.1  & 60.7  & 54.8 & 66.1  \\
          & HRNet~\cite{sun2019deep} & HRNet-W32 & 67.8  & 75.5  & 54.3  & 66.1  & 63.0  & 70.8  & 46.7  & 56.6  & 53.6  & 63.6  \\
          & HRNet~\cite{sun2019deep} & HRNet-W48 & \underline{70.1}  & \textbf{77.6}  & 67.5  & 78.7  & 65.6  & 74.3  & \textbf{53.5}  & \underline{63.9}  & 57.9  & 68.1  \\
          & TCFormer~\cite{zeng2022not} & TCFormer & 69.1  & 77.0  & \underline{69.8}  & \textbf{81.3}  & 64.9  & 74.6  & \textbf{53.5} & \textbf{65.0}  & 57.2  & 67.8  \\
          & RTMPose-m~\cite{jiang2023rtmpose} & CSPNeXt-m & 67.3  & 75.0  & 61.5  & 75.2  & 81.3  & 87.1  & 47.5  & 58.9  & 58.2  & 67.4  \\
          & DWPose-m~\cite{yang2023effective} & CSPNeXt-m & 68.5  & 76.1  & 63.6  & 77.2  & \underline{82.8}  & \underline{88.1}  & 52.7  & 63.4  & \underline{60.6}  & \underline{69.5}  \\
          & ViTPose-B~\cite{xu2022vitpose} & ViT-B & 69.6  & -  & \textbf{70.1}  & -  & 62.1  & -  & 50.8  & -  & 56.3  & - \\
          \midrule
    Generalist & Hulk-FT  & ViT-B & \textbf{70.2}  & \underline{77.5}  & 65.9  & \underline{79.1}  & \textbf{84.2}  & \textbf{90.1}  & \underline{53.2}  & 63.1  & \textbf{61.6}  & \textbf{72.1}  \\
    \bottomrule
    \end{tabular}%
    }
  \label{tab:wholebody}%
\end{table*}%

\subsection{\textcolor{black}{ICFG-PEDES}}
\textcolor{black}{
To further evaluate the effectiveness of Hulk on the pedestrian captioning task, we conduct transfer learning with ICFG-PEDES~\cite{ding2021semantically}, another pedestrian dataset with detailed language captions. All pretrained image caption methods are fine-tuned on the training set with five epochs, and evaluated on the test set. As shown in Table~\ref{tab:icfg}, Hulk outperforms powerful BLIP~\cite{li2022blip} and BLIP2~\cite{li2023blip} with $\textbf{3.2}$ B@4 and $\textbf{1.5}$ B@4, \textbf{20.1} CIDEr and \textbf{12.8} CIDEr, respectively. Using fewer vision-language pretraining data (12M vs. 129M), the superior transfer performance of Hulk demonstrates its general capabilities on human-centric perception tasks.
}

\begin{table}[tbp]\color{black}
  \centering
  \caption{Pedestrian image caption finetuning results on ICFG-PEDES~\cite{ding2021semantically}. * denotes fixed backbone.}
  \resizebox{\linewidth}{!}{
    \begin{tabular}{lllcccc}
    \toprule
    \multicolumn{1}{l}{Method} &       & Backbone & {B@4} & CIDEr & Trainable Params & Total Params \\
    \midrule
    \multirow{2}[1]{*}{Specialist} & BLIP~\cite{li2022blip}  & ViT-B & 22.5  & 36.9 & 446.0M & 446.0M \\
          & BLIP-2~\cite{li2023blip} & ViT-G* & 24.2  & 44.2 & 188.0M &1.1B \\
    \midrule
    Generalist & Hulk  & ViT-B & \textbf{25.7}  & \textbf{57.0} & 113.7M & 113.7M\\
    \bottomrule
    \end{tabular}%
  \label{tab:icfg}%
  }
\end{table}%

\section{\textcolor{black}{Model Parameters and FLOPs}}
\textcolor{black}{
We compute the model parameters and FLOPs of Hulk and some representative methods on all eight human-centric tasks in Table~\ref{tab:response_param}. Hulk-B has 113.7M parameters which is comparable with SOTA methods on most human-centric tasks. The only exceptions are skeleton-based action recognition and pedestrian image caption tasks. Compared with specialist models for the skeleton-based action recognition task, Hulk needs more parameters to enable its general abilities on other more complex human-centric tasks. Compared with specialist models for the pedestrian image caption task, notably,  Hulk achieves comparable performance but with a more efficient architecture that eliminates the need for an additional text encoder, resulting in fewer parameters and lower computational requirements(GFLOPs).
}
\begin{table*}[htbp]\color{black}
  \centering
  \caption{Model parameters and GFLOPs of Hulk and other methods. GFLOPs are computed with the input size reported in corresponding papers. \dag denotes fixed backbone. \ddag We compute the GFLOPs of autoregressively forwarding one step in the pedestrian image caption task.}
  \resizebox{.8\linewidth}{!}{
    \begin{tabular}{llllccc}
    \toprule
    Task  & Dataset & Method & Backbone & Performance & Params & GFLOPs \\
    \midrule
    \multirow{4}[4]{*}{Detection} & \multirow{4}[4]{*}{CrowdHuman} & DETR  & ResNet-50 & 75.9\% mAP & 41.0M & 187.0  \\
          &       & DDETR & ResNet-50 & 91.5\% mAP & 40.0M   & 173.0  \\
          &       & PATH-B & ViT-B & 90.9\% mAP & 106.0M & 341.3  \\
\cmidrule{3-7} 
&       & Hulk-B & ViT-B & 92.4\% mAP & 113.7M & 348.2  \\
    \midrule
    \multirow{5}[3]{*}{2D Pose} & \multirow{5}[3]{*}{COCO} & HRNet & HRNET-W48      & 75.1\% AP & 63.6M & 32.9  \\
          &       & TokenPose-L/D24 &   HRNET-W48-S3    & 75.8\% AP & 27.5M & 11.0  \\
          &       & ViTPose-B  & ViT-B & 75.8\% AP & 86.0M  & 17.1  \\
          &       & PATH-B & ViT-B & 76.3\% AP & 106.0M & 18.2  \\
\cmidrule{3-7}
&       & Hulk-B & ViT-B & 77.5\% AP & 113.7M & 19.1  \\
\midrule
    \multirow{4}[3]{*}{Parsing} & \multirow{4}[3]{*}{LIP} & SCHP  & ResNet-101 & 59.4\% mIoU & 66.7M & 77.2  \\
          &       & PCNet & ResNet-101 & 57.0\% mIoU & 53.8M & 232.5  \\
          &       & PATH-B & ViT-B & 61.4\% mIoU & 106.0M & 87.3  \\
\cmidrule{3-7}
&       & Hulk-B & ViT-B & 64.0\% mIoU & 113.7M & 91.4  \\
    \midrule
    \multirow{4}[4]{*}{Attribute} & \multirow{4}[4]{*}{PA-100K} & C-Tran & ResNet-101 & 81.5\% mA & 67.8M & 11.1  \\
          &       & DAFL  & ResNet-50 & 83.5\% mA & {154.2M} & 24.8  \\
          &       & PATH-B & ViT-B & 86.9\% mA & 106.0M & 18.2  \\
\cmidrule{3-7}          &       & Hulk-B & ViT-B & 87.9\% mA & 113.7M & 19.1  \\
    \midrule
    \multirow{4}[4]{*}{Skeleton} & \multirow{4}[4]{*}{NTU60-XSub} & ST-GCN  & GCN   & 81.5\% acc & 1.2M  & 1.7  \\
          &       & AGCN  & GCN   & 88.5\% acc & 1.6M  & 2.1  \\
          &       & PoseConv3D & CNN   & 93.1\% acc & 2.0M  & 15.9  \\
\cmidrule{3-7} 
&       & Hulk-B & ViT-B & 94.0\% acc & 113.7M & 20.1  \\
    \midrule
    \multirow{4}[4]{*}{3D Pose \& Mesh} & \multirow{4}[4]{*}{Human3.6M} & METRO & HRNet-W64 & 56.5 MPJPE & 230.4M & 27.4  \\
          &       & MeshGraphormer & HRNet-W64 & 51.2 MPJPE & 226.5M & 28.9  \\
          &       & FastMETRO & HRNet-W64 & 52.2 MPJPE & 153.0M & 21.2  \\
\cmidrule{3-7}   
&       & Hulk-B & ViT-B & 43.6 MPJPE & 113.7M & 19.7  \\
    \midrule
    \multirow{3}[3]{*}{Caption} & \multirow{3}[3]{*}{CUHK-PEDES} & BLIP  & ViT-B & 32.9 B@4 & 446.0M & 506.1\ddag \\
          &       & BLIP-2 & ViT-G\dag & 32.8 B@4 & 1.1B  & 2814.7\ddag \\
\cmidrule{3-7}  
&       & Hulk-B & ViT-B & 31.1 B@4 & 113.7M & 58.7\ddag \\
\bottomrule

    \end{tabular}%
    }
  \label{tab:response_param}%
\end{table*}%

\section{\textcolor{black}{Additional Visualization}}
\subsection{\textcolor{black}{Visualization of Eight Tasks}}
\begin{figure*}[t]
    \centering
    \includegraphics[width=\linewidth]{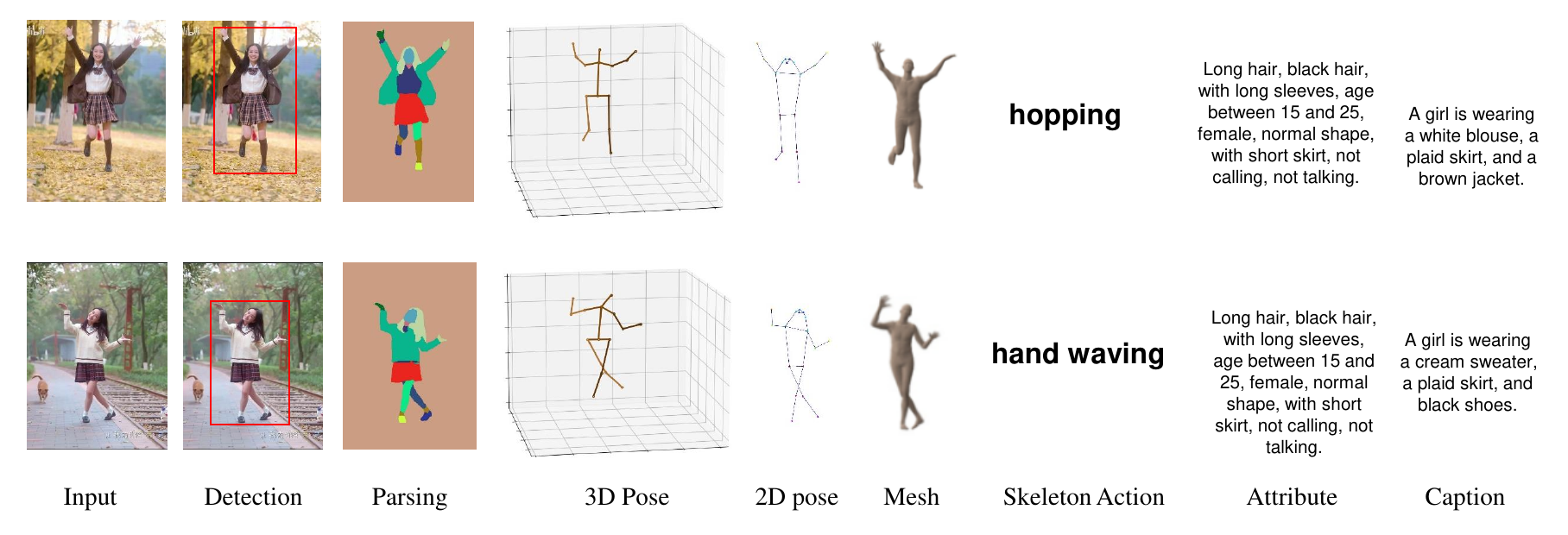}
    \caption{Visualization of Hulk's performance on eight human-centric tasks for the same input.}
    \label{fig:8task}
\end{figure*}

\textcolor{black}{
In Figure~\ref{fig:8task}, we utilize Hulk to simultaneously predict eight human-centric tasks on the same input image, demonstrating excellent performance across all tasks simultaneously.
}

\subsection{\textcolor{black}{Visualization For Task Collaboration}}
\textcolor{black}{
To qualitatively demonstrate how one task helps the other in Hulk, we provide visualizations in two ways, \emph{i.e.}, visualization of prediction results and intermediate feature maps in diverse tasks.}

\subsubsection{\textcolor{black}{Visualization of prediction results in different tasks}}
\begin{figure*}[t]
    \centering
    \includegraphics[width=0.9\linewidth]{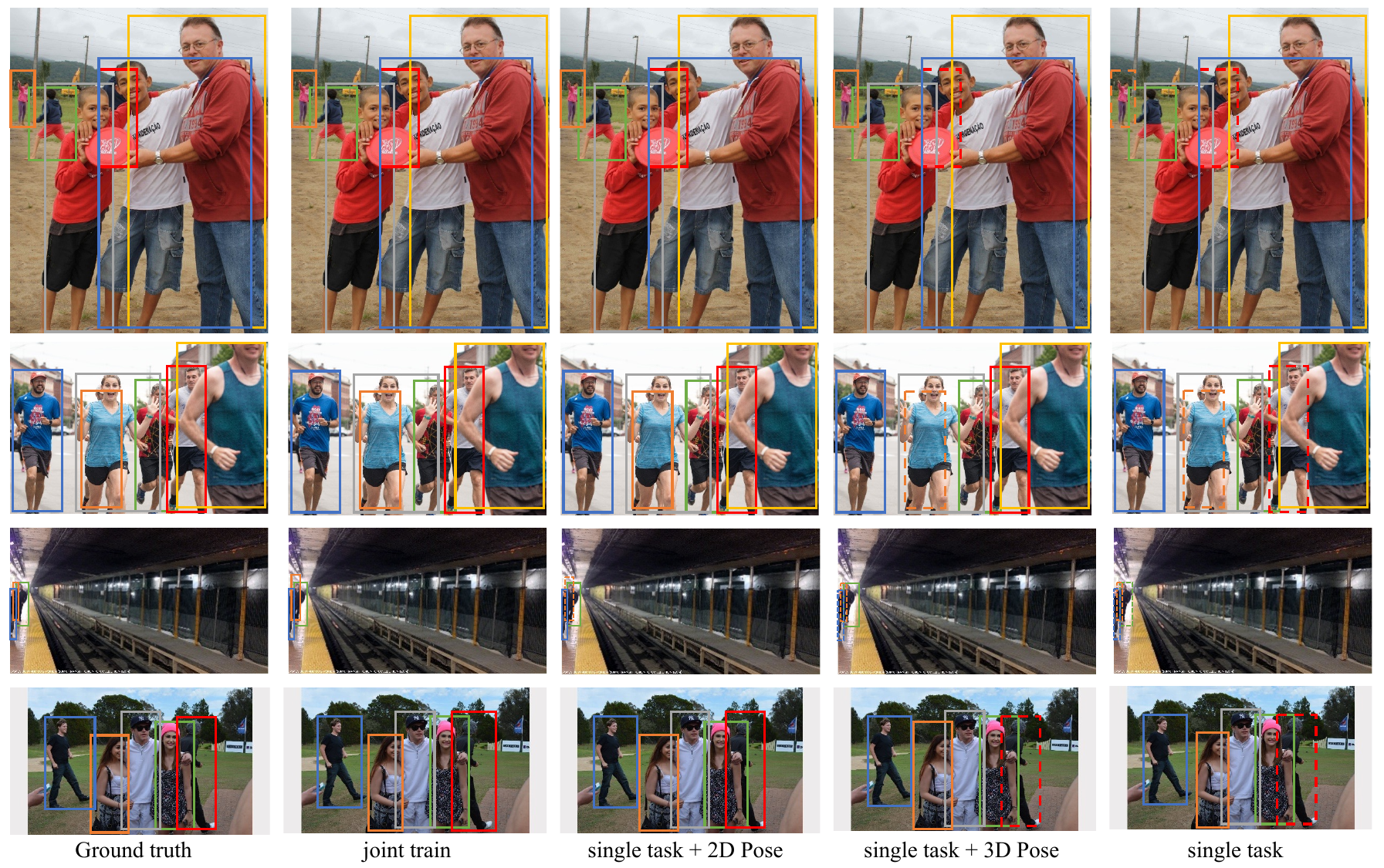}
    \caption{Visualization of pedestrian detection tasks on CrowdHuman~\cite{shao2018crowdhuman} dataset. The dashed boxes indicate missed detections.}
    \label{fig:det}
\end{figure*}

\begin{figure*}[t]
    \centering
    \includegraphics[width=0.9\linewidth]{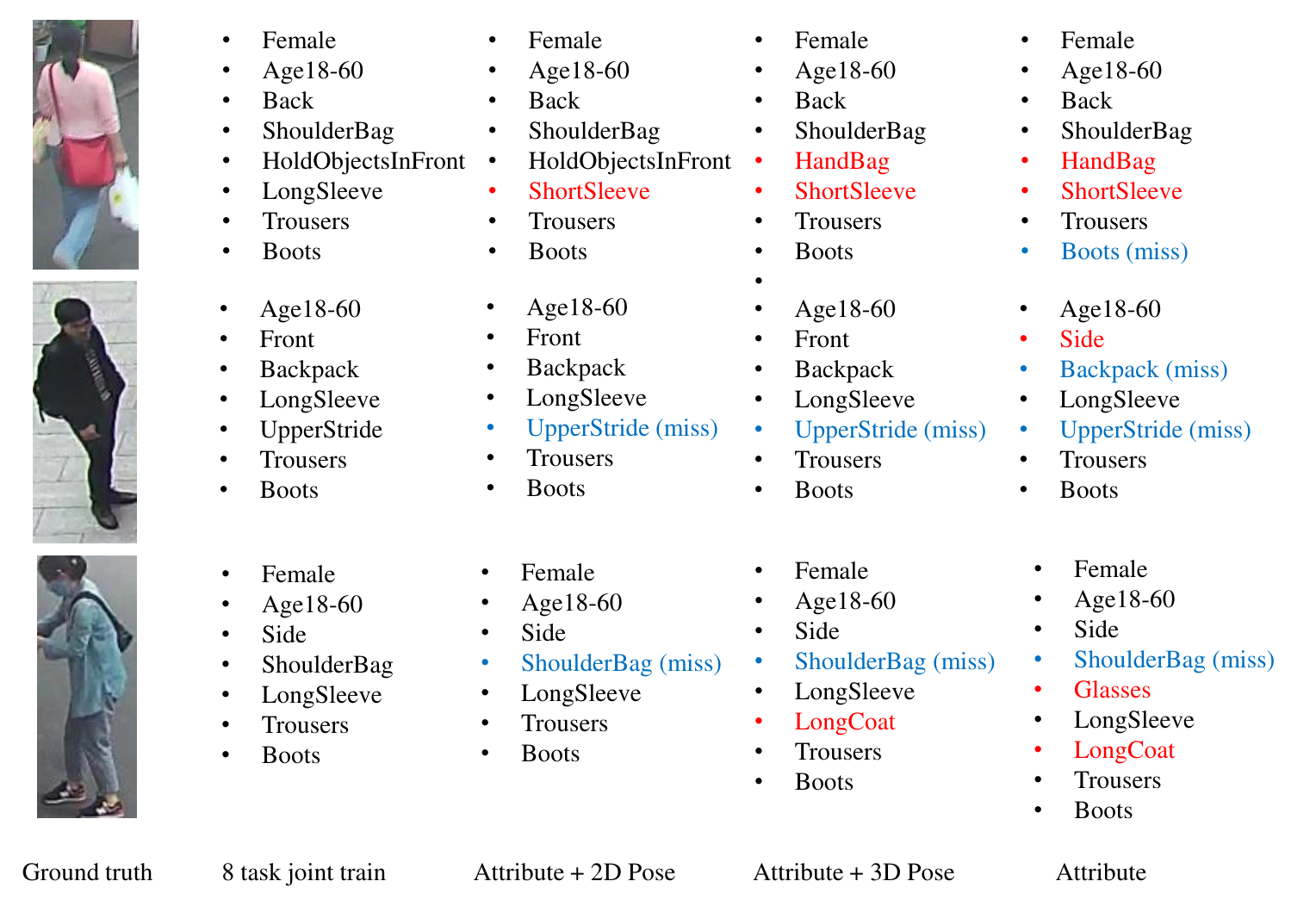}
    \caption{Visualization of pedestrian attribute recognition task on the PA-100K~\cite{pa} dataset. The \textcolor[rgb]{0,0.40,0.701}{missed} and \textcolor{red}{incorrectly identified} attributes are marked in blue and red, respectively.
    }
    \label{fig:attri}
\end{figure*}

\textcolor{black}{We compare the predicted results on small tasks, \emph{e.g.,} pedestrian detection and attribute recognition tasks, with different training sets to visualize potential task collaborations: a) one dataset from a small task; b) one dataset from a small task + one dataset from a generic task; and c) datasets from 8 tasks. 
In the small ablation training set of Hulk, we choose \underline{detection and attribute recognition as small tasks} and choose \underline{2D pose estimation and 3D pose estimation as generic tasks}: 
\begin{itemize}
    \item We provide qualitative visualizations for pedestrian detection on the CrowdHuman validation set to visualize potential collaborations of tasks. As shown in Figure~\ref{fig:det}, the dashed boxes indicate missed detection. It can be observed that when trained on the single detection task, the model fails to detect some occluded individuals or small individuals (column 5). However, Hulk can detect previously missed individuals by jointly training additional generic tasks (such as 2D pose and 3D pose) and pedestrian detection tasks (columns 3 and 4). Furthermore, when Hulk is jointly trained with all eight tasks, all previously missed individuals are successfully detected (column 2), indicating potential collaborations among human-centric tasks.
    \item We also provide the qualitative visualizations for another smaller task, \emph{i.e.,} pedestrian attribution recognition. As shown in Figure~\ref{fig:attri} (column 5), when only training on the attribute recognition task, Hulk misses some easy attributes (marked blue) and predicts wrong attributes (marked red) in the images, \emph{e.g.}, boots, backpacks, and shoulder bags. However, when 2D or 3D pose estimation tasks are incorporated into the joint training, most of these issues are resolved (columns 3 and 4). Furthermore, when we train Hulk on all 8 tasks (see column 2), even highly concealed attributes can be recognized, such as the UpperStride (stride T-shirt, a specific attribute in PA-100K) that the man wears under the coat in the second-row. These visualizations indicate that jointly training with generic tasks can benefit the learning on pedestrian attribute recognition, which is also a small task. 
\end{itemize}
}

\subsubsection{\textcolor{black}{Visualization of intermediate feature maps}}
\begin{figure*}[t]
    \centering
    \includegraphics[width=0.9\linewidth]{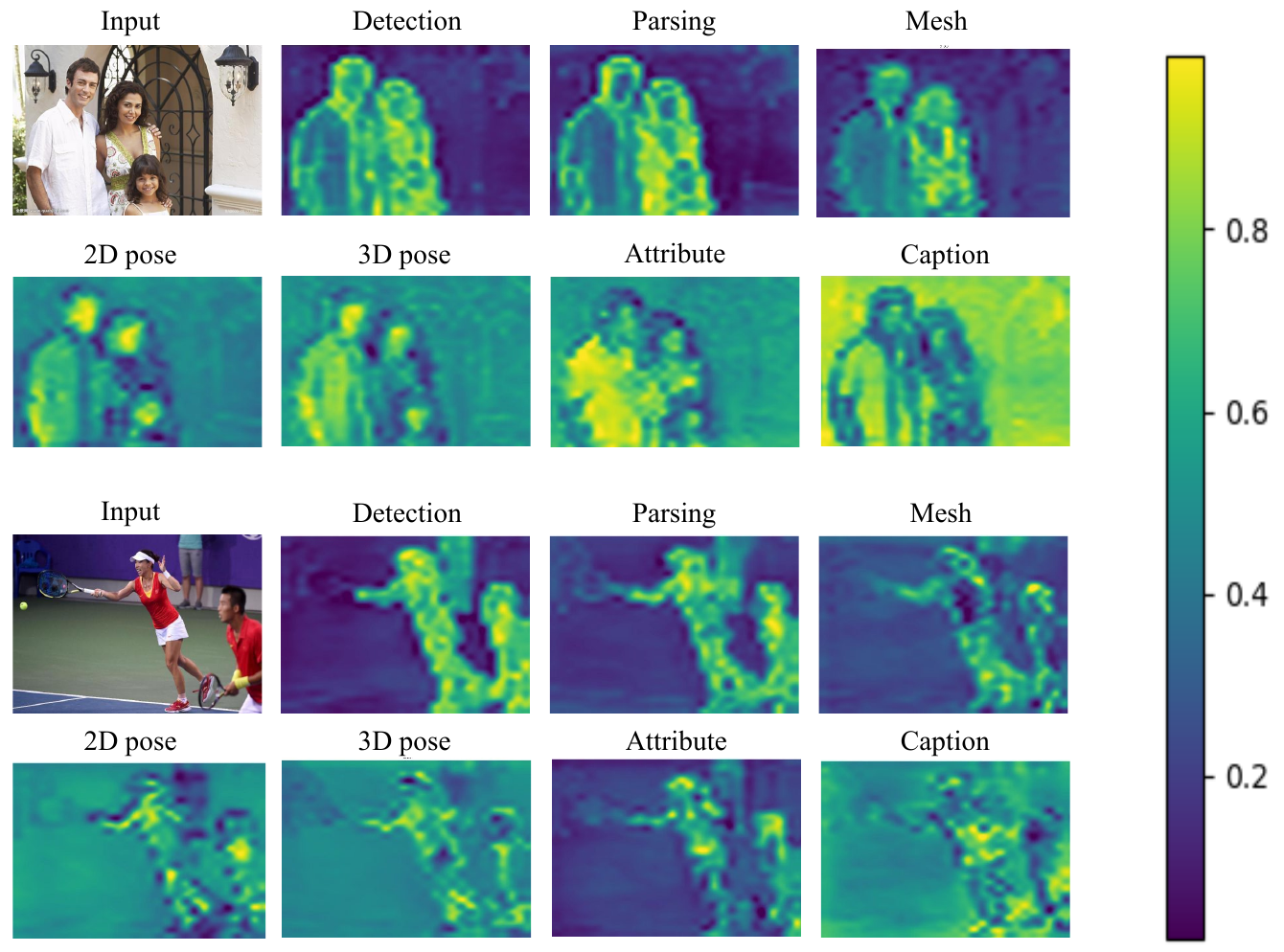}
    \caption{Visualization of the feature map from the last layer of the decoder on seven human-centric tasks.}
    \label{fig:hm}
\end{figure*}

\textcolor{black}{
In Figure~\ref{fig:hm}, we visualize the feature map of the shared decoder's last layer to confirm the consistency of regions of interest across different human-centric tasks. The similarities of intermediate feature maps help to visualize the improved performance of small tasks with the help of some generic tasks.}

\textcolor{black}{Specifically, 
2D pose estimation and 3D pose estimation tasks highlight similar regions around human body joints, illustrating the high correlation between these two tasks. The phenomenon is consistent with the quantitative results in Table XII, where removing the 2D pose estimation task during joint training leads to a notable increased error (\textbf{+16.7} MPJPE) in 3D pose estimation. Additionally, the feature map of 2D pose estimation task shows clear boundaries of the pedestrians (the low light part between the background and pedestrians), which helps us to understand why removing 2D pose task significantly decreases performance (\textbf{-16.8\%} AP) in detection in Table XII. On the contrary, the feature maps of attribute and mesh recovery tasks show a low similarity. We suggest that features of mesh recovery can not help the model highlight regions of attributes, \emph{e.g.}, the upper clothing of the man in see the top example in Figure~\ref{fig:hm}, resulting in slight performance changes (\textbf{+0.1} mA) in attribute recognition after removing the mesh recovery task.
}

\section{Details of training datasets}
\subsection{Full setting}
As shown in Table~\ref{tab:fullset}, we utilize 42 publically available datasets to form the training set for training our Hulk model, containing 30,187,836 training samples and covering eight different human-centric tasks, including human parsing, 2D pose estimation, attribute recognition, pedestrian detection, skeleton-based action recognition, image caption, 3D pose estimation, and mesh recovery. 

\subsection{Ablation setting}
In ablation studies, to demonstrate the effectiveness of Hulk and further explore the potential of a human-centric foundation model, we conduct several ablations on a smaller training set. In Table~\ref{tab:subset}, the ablation set consists of 21 datasets with about 1.7M samples, covering all eight tasks. 

\begin{table}[t]
  \centering
  \caption{Statistics of Training Datasets in the full setting.}
    \begin{tabular}{p{8.69em}lc}
    \toprule
    \multicolumn{1}{l}{Task type} & \multicolumn{1}{l}{Dataset} & \multicolumn{1}{l}{Number of samples} \\
    \midrule
     \multicolumn{1}{l}{} & \multicolumn{1}{l}{Human3.6M~\cite{h36m_pami}} & \multicolumn{1}{c}{62,668} \\
      \multicolumn{1}{l}{} & \multicolumn{1}{l}{LIP~\cite{lip}} & \multicolumn{1}{c}{30,462} \\
    \multirow{2}[0]{*}{Human parsing} & \multicolumn{1}{l}{CIHP~\cite{cihp}} & \multicolumn{1}{c}{28,280} \\
    \multicolumn{1}{l}{} & \multicolumn{1}{l}{ModaNet~\cite{zheng2018modanet}} & \multicolumn{1}{c}{52,245} \\
    \multirow{1}[0]{*}{(7 datasets)} & \multicolumn{1}{l}{VIP~\cite{vip}} & \multicolumn{1}{c}{18,469} \\
    \multicolumn{1}{l}{} & \multicolumn{1}{l}{Deep fashion~\cite{ge2019deepfashion2}} & \multicolumn{1}{c}{191,961} \\
    \multicolumn{1}{l}{} & \multicolumn{1}{l}{Paper Doll~\cite{paper}} & \multicolumn{1}{c}{1,035,825} \\
    \midrule
         \multicolumn{1}{l}{} & \multicolumn{1}{l}{COCO~\cite{coco}} & \multicolumn{1}{c}{149,813} \\
      \multicolumn{1}{l}{} & \multicolumn{1}{l}{AIC~\cite{aic}} & \multicolumn{1}{c}{378,352} \\
    \multirow{2}[0]{*}{2D pose estimation} & \multicolumn{1}{l}{Human3.6M (pose)~\cite{h36m_pami}} & \multicolumn{1}{c}{312,187} \\
    \multicolumn{1}{l}{} & \multicolumn{1}{l}{Posetrack~\cite{andriluka2018posetrack}} & \multicolumn{1}{c}{97,174} \\
    \multirow{1}[0]{*}{(8 datasets)} & \multicolumn{1}{l}{3DPW~\cite{3dpw}} & \multicolumn{1}{c}{68,663} \\
    \multicolumn{1}{l}{} & \multicolumn{1}{l}{JRDB-Pose~\cite{jrdb}} & \multicolumn{1}{c}{139,385} \\
    \multicolumn{1}{l}{} & \multicolumn{1}{l}{MPI-INF-3DHP~\cite{mehta2017monocular}} & \multicolumn{1}{c}{1,031,701} \\
    \multicolumn{1}{l}{} & \multicolumn{1}{l}{AIST++~\cite{aist}} & \multicolumn{1}{c}{1,015,257} \\
    \midrule
      \multicolumn{1}{l}{} & \multicolumn{1}{l}{RAPv2~\cite{rapv2}} & \multicolumn{1}{c}{67,943} \\
    \multirow{2}[0]{*}{Attribute recognition} & \multicolumn{1}{l}{PA-100k~\cite{pa}} & \multicolumn{1}{c}{90,000} \\
    \multicolumn{1}{l}{} & \multicolumn{1}{l}{Parse27k~\cite{PARSE27k}} & \multicolumn{1}{c}{27,482} \\
    \multirow{1}[0]{*}{(6 datasets)} & \multicolumn{1}{l}{Market~\cite{market}} & \multicolumn{1}{c}{12,926} \\
    \multicolumn{1}{l}{} & \multicolumn{1}{l}{HARDHC~\cite{HARDHC}} & \multicolumn{1}{c}{28,336} \\
    \multicolumn{1}{l}{} & \multicolumn{1}{l}{LU-Person~\cite{fu2021unsupervised}} & \multicolumn{1}{c}{10,684,342} \\
    \midrule
          \multicolumn{1}{l}{} & \multicolumn{1}{l}{CrowdHuman~\cite{shao2018crowdhuman}} & \multicolumn{1}{c}{15,000} \\
    \multirow{2}[0]{*}{Pedestrian detection} & \multicolumn{1}{l}{EuroCity~\cite{braun2018eurocity}} & \multicolumn{1}{c}{21,785} \\
    \multicolumn{1}{l}{} & \multicolumn{1}{l}{CityPersons~\cite{zhang2017citypersons}} & \multicolumn{1}{c}{2,778} \\
    \multirow{1}[0]{*}{(6 datasets)} & \multicolumn{1}{l}{WiderPerson~\cite{zhang2019widerperson}} & \multicolumn{1}{c}{9,000} \\
    \multicolumn{1}{l}{} & \multicolumn{1}{l}{WiderPedestrian~\cite{loy2019wider}} & \multicolumn{1}{c}{58,009} \\
    \multicolumn{1}{l}{} & \multicolumn{1}{l}{COCO-person~\cite{coco}} & \multicolumn{1}{c}{64,115} \\
    \midrule
              \multicolumn{1}{l}{} & \multicolumn{1}{l}{NTU60~\cite{shahroudy2016ntu}} & \multicolumn{1}{c}{40,091} \\
    \multirow{2}[0]{*}{Skeleton-based action} & \multicolumn{1}{l}{NTU120~\cite{liu2019ntu}} & \multicolumn{1}{c}{63,026} \\
    \multicolumn{1}{l}{} & \multicolumn{1}{l}{GYM~\cite{shao2020finegym}} & \multicolumn{1}{c}{2,778} \\
    \multirow{1}[0]{*}{(6 datasets)} & \multicolumn{1}{l}{Diving48~\cite{diving}} & \multicolumn{1}{c}{15,027} \\
    \multicolumn{1}{l}{} & \multicolumn{1}{l}{UCF101~\cite{soomro2012ucf101}} & \multicolumn{1}{c}{13,320} \\
    \multicolumn{1}{l}{} & \multicolumn{1}{l}{Kinetics-400~\cite{kay2017kinetics}} & \multicolumn{1}{c}{239,737} \\
    \midrule
    Image caption & \multicolumn{1}{l}{CUHK-PEDES~\cite{cuhk}} & \multicolumn{1}{c}{68,126} \\
    \multicolumn{1}{l}{(2 datasets)} & \multicolumn{1}{l}{SYNTH-PEDES~\cite{zuo2023plip}} & \multicolumn{1}{c}{12,138,157} \\
    \midrule
 \multicolumn{1}{l}{} & \multicolumn{1}{l}{Human3.6M~\cite{h36m_pami}} & \multicolumn{1}{c}{312,188} \\
      \multicolumn{1}{l}{} & \multicolumn{1}{l}{3DPW~\cite{3dpw}} & \multicolumn{1}{c}{22,735} \\
    \multirow{2}[0]{*}{3D pose \& Mesh} & \multicolumn{1}{l}{COCO~\cite{coco}} & \multicolumn{1}{c}{40,055} \\
    \multicolumn{1}{l}{} & \multicolumn{1}{l}{MUCO~\cite{muco}} & \multicolumn{1}{c}{101,883} \\
    \multirow{1}[0]{*}{(7 datasets)} & \multicolumn{1}{l}{UP-3D~\cite{up3d}} & \multicolumn{1}{c}{7,126} \\
    \multicolumn{1}{l}{} & \multicolumn{1}{l}{MPII~\cite{mpii}} & \multicolumn{1}{c}{14,810} \\
    \multicolumn{1}{l}{} & \multicolumn{1}{l}{GTA~\cite{gta}} & \multicolumn{1}{c}{1,396,913} \\
    \midrule
    Total & 42    & 30,187,836$\approx$\textbf{30M} \\
    \bottomrule
    \end{tabular}
  \label{tab:fullset}%
\end{table}%

\begin{table}[t]
  \centering
  \caption{Statistics of Training Datasets in the ablation setting.}
    \begin{tabular}{p{8.69em}lc}
    \toprule
    \multicolumn{1}{l}{Task type} & \multicolumn{1}{l}{Dataset} & \multicolumn{1}{l}{Number of samples} \\
    \midrule
     \multirow{2}[0]{*}{Human parsing} & \multicolumn{1}{l}{Human3.6M~\cite{h36m_pami}} & \multicolumn{1}{c}{62,668} \\
      \multicolumn{1}{l}{} & \multicolumn{1}{l}{LIP~\cite{lip}} & \multicolumn{1}{c}{30,462} \\
    \multirow{1}[0]{*}{(3 datasets)} & \multicolumn{1}{l}{CIHP~\cite{cihp}} & \multicolumn{1}{c}{28,280} \\

    \midrule
        2D pose estimation & \multicolumn{1}{l}{COCO~\cite{coco}} & \multicolumn{1}{c}{149,813} \\
     (2 datasets) & \multicolumn{1}{l}{AIC~\cite{aic}} & \multicolumn{1}{c}{378,352} \\

    \midrule
      \multicolumn{1}{l}{} & \multicolumn{1}{l}{RAPv2~\cite{rapv2}} & \multicolumn{1}{c}{67,943} \\
    \multirow{2}[0]{*}{Attribute recognition} & \multicolumn{1}{l}{PA-100k~\cite{pa}} & \multicolumn{1}{c}{90,000} \\
    \multicolumn{1}{l}{} & \multicolumn{1}{l}{Parse27k~\cite{PARSE27k}} & \multicolumn{1}{c}{27,482} \\
    \multirow{1}[0]{*}{(5 datasets)} & \multicolumn{1}{l}{Market~\cite{market}} & \multicolumn{1}{c}{12,926} \\
    \multicolumn{1}{l}{} & \multicolumn{1}{l}{HARDHC~\cite{HARDHC}} & \multicolumn{1}{c}{28,336} \\
    \midrule
        Pedestrian detection & \multicolumn{1}{l}{CrowdHuman~\cite{shao2018crowdhuman}} & \multicolumn{1}{c}{15,000} \\

    \midrule
          Skeleton action & \multicolumn{1}{l}{NTU60~\cite{shahroudy2016ntu}} & \multicolumn{1}{c}{40,091} \\
     (2 datasets) & \multicolumn{1}{l}{NTU120~\cite{liu2019ntu}} & \multicolumn{1}{c}{63,026} \\

    \midrule
    Image caption & \multicolumn{1}{l}{CUHK-PEDES~\cite{cuhk}} & \multicolumn{1}{c}{68,126} \\
    \multicolumn{1}{l}{(2 datasets)} & \multicolumn{1}{l}{SYNTH-PEDES~\cite{zuo2023plip}} & \multicolumn{1}{c}{132,000} \\
    \midrule
 \multicolumn{1}{l}{} & \multicolumn{1}{l}{Human3.6M~\cite{h36m_pami}} & \multicolumn{1}{c}{312,188} \\
      \multicolumn{1}{l}{} & \multicolumn{1}{l}{3DPW~\cite{3dpw}} & \multicolumn{1}{c}{22,735} \\
    \multirow{2}[0]{*}{3D pose \& Mesh} & \multicolumn{1}{l}{COCO~\cite{coco}} & \multicolumn{1}{c}{40,055} \\
    \multicolumn{1}{l}{} & \multicolumn{1}{l}{MUCO~\cite{muco}} & \multicolumn{1}{c}{101,883} \\
    \multirow{1}[0]{*}{(6 datasets)} & \multicolumn{1}{l}{UP-3D~\cite{up3d}} & \multicolumn{1}{c}{7,126} \\
    \multicolumn{1}{l}{} & \multicolumn{1}{l}{MPII~\cite{mpii}} & \multicolumn{1}{c}{14,810} \\

    \midrule
    Total & 21    & 1,693,302$\approx$\textbf{1.7M} \\
    \bottomrule
    \end{tabular}%
  \label{tab:subset}%
\end{table}%

\section{Details of dataset-wise configurations}
\subsection{Training}
We provide detailed dataset-wise training configurations in Table~\ref{tab:train}, including total batch size, batch size per GPU, the number of GPUs, and task weights. In order to better unify different tasks and different datasets, the different datasets of certain tasks (i.e., attribute recognition, pedestrian detection, skeleton-based action, image caption, 3D pose, and mesh recovery) are stacked together as an integrated dataset for training, and the datasets of some other tasks (i.e., human parsing and 2D pose estimation) are separately trained on different GPUs with diverse task weights.

\subsection{\textcolor{black}{Steps to adjust task weight}}
\textcolor{black}{
 (1) In Hulk, there are two steps to adjust task weights: 
\begin{itemize}
    \item \textbf{Step1: Make sure different human-centric tasks provide similar gradients to task-shared parameters.} Since de-tokenizers in Hulk are not shared across all human-centric tasks, we use the last parameter (Layer Norm) in the decoder of Hulk as a probe during Hulk's training on the small ablation set. The L2-normed gradients of this probe from different tasks should be similar. 
    \item {\textbf{Step2: Ablate the generated task weights by $0.1\times$, $0.3\times$, $3\times$, $10\times$, etc., for better performance of all eight tasks.}} We provide the L2-normed gradients of all eight tasks in Table~\ref{tab:gradnorm}. Even after adjusting task weights, the L2-normed gradients of all eight tasks are almost in the same order of magnitude. 
\end{itemize}
}

\begin{table}[htbp]
  \centering
  \caption{For eight tasks, we compute the L2-normed gradient using the last parameter in the decoder of Hulk.}
  \resizebox{\linewidth}{!}{
    \begin{tabular}{ccccccc}
    \toprule
    Attribute & Caption & Skeleton & 3D Pose\&Mesh & Detection   & 2D pose & Parsing \\
    \midrule
    0.93  & 1.08  & 0.48  & 3.83  & 5.01  & 3.35  & 0.83 \\
    \bottomrule
    \end{tabular}%
    }
  \label{tab:gradnorm}%
\end{table}%

\subsection{Fine-tuning}
For the fine-tuning process, we carefully tune the learning rate, total batch size, the number of iterations, and drop path rate, and report the best performances. In detail, the ViT-base backbone (i.e., in Table~\ref{tab:ft}) and ViT-large backbone (i.e., in Table~\ref{tab:ft_large}) are tuned, respectively.

\begin{table*}[htbp]
  \centering
  \caption{Hulk joint training setup.}
    \begin{tabular}{clcccc}
    \toprule
    Task Type  & \multicolumn{1}{c}{Dataset} & Batch Size & Batch Size per GPU & GPUs  & Task Weight \\
    \midrule
    \multirow{7}[2]{*}{Human parsing } & LIP   & 108   & 27    & 4     & 1.8 \\
          & CIHP  & 104   & 26    & 4     & 3.6 \\
          & Human3.6M & 217   & 31    & 7     & 2.25 \\
          & ModaNet  & 27    & 27    & 1     & 0.021 \\
          & VIP   & 27    & 27    & 1     & 0.021 \\
          & Deepfashion & 54    & 27    & 2     & 0.042 \\
          & PaperDoll & 27    & 27    & 1     & 0.021 \\
    \midrule
    \multirow{8}[2]{*}{2D pose estimation} & COCO  & 528   & 176   & 3     & 28000 \\
          & AIC   & 1323  & 189   & 7     & 56000 \\
          & Human3.6M & 264   & 132   & 2     & 3192 \\
          & Posetrack & 340   & 170   & 2     & 12335 \\
          & JRDB  & 340   & 170   & 2     & 8223 \\
          & MPI-INF-3DHP & 340   & 170   & 2     & 8223 \\
          & 3DPW  & 170   & 170   & 1     & 2055 \\
          & AIST++ & 170   & 170   & 1     & 2055 \\
    \midrule
    \multirow{6}[2]{*}{Attribute recognition } & RAPv2 & \multirow{5}[1]{*}{147} & \multirow{5}[1]{*}{147} & \multirow{5}[1]{*}{1} & \multirow{5}[1]{*}{5} \\
          & PA-100k    &       &       &       &  \\
          & Parse27k &       &       &       &  \\
          & Market &       &       &       &  \\
          & HARDHC &       &       &       &  \\
          \cmidrule{2-6}  
          & LUPerson & 300   & 300   & 1     & 5 \\
    \midrule
    \multirow{6}[4]{*}{Pedestrian detection } & CrowdHuman & 32    & 4     & 8     & 15 \\
\cmidrule{2-6}          & EuroCity  & \multirow{5}[2]{*}{80} & \multirow{5}[2]{*}{4} & \multirow{5}[2]{*}{20} & \multirow{5}[2]{*}{42.4} \\
          & CityPersons &       &       &       &  \\
          & WiderPerson &       &       &       &  \\
          & WiderPedestrian &       &       &       &  \\
          & COCO  &       &       &       &  \\
    \midrule
    \multirow{6}[4]{*}{Skeleton action} & Ntu60 & \multirow{3}[2]{*}{240} & \multirow{3}[2]{*}{120} & \multirow{3}[2]{*}{2} & \multirow{3}[2]{*}{4.4} \\
          & Ntu120 &       &       &       &  \\
          & gym   &       &       &       &  \\
\cmidrule{2-6}          & Diving48 & \multirow{3}[2]{*}{90} & \multirow{3}[2]{*}{90} & \multirow{3}[2]{*}{1} & \multirow{3}[2]{*}{1} \\
          & UCF101 &       &       &       &  \\
          & K400  &       &       &       &  \\
    \midrule
    \multirow{2}[2]{*}{Image caption} & SYNTH-PEDES & \multirow{2}[2]{*}{300} & \multirow{2}[2]{*}{100} & \multirow{2}[2]{*}{3} & \multirow{2}[2]{*}{90} \\
          & CUHK-PEDES &       &       &       &  \\
    \midrule
    \multirow{7}[2]{*}{3D pose \& Mesh} & 3DPW  & \multirow{7}[2]{*}{495} & \multirow{7}[2]{*}{165} & \multirow{7}[2]{*}{3} & \multirow{7}[2]{*}{0.5} \\
          & Human3.6M &       &       &       &  \\
          & COCO  &       &       &       &  \\
          & MUCO &       &       &       &  \\
          & UP-3D &       &       &       &  \\
          & MPII  &       &       &       &  \\
          & GTA   &       &       &       &  \\
    \bottomrule
    \end{tabular}%
  \label{tab:train}%
\end{table*}%

\begin{table*}[htbp]
  \centering
  \caption{Detailed finetuning configs \textbf{with ViT-base backbone} for human-centric tasks.}
    \begin{tabular}{clcccc}
    \toprule
    Task Type  & \multicolumn{1}{c}{Dataset} & \multicolumn{1}{c}{Learning Rate} & \multicolumn{1}{c}{Batch Size} & \multicolumn{1}{c}{Iterations} & \multicolumn{1}{c}{Drop Path Rate} \\
    \midrule
    \multirow{2}[2]{*}{2D pose estimation} & COCO  & \multicolumn{1}{c}{3.00E-05} & \multicolumn{1}{c}{1024} & \multicolumn{1}{c}{20k} & \multicolumn{1}{c}{0.3} \\
          & AIC   & \multicolumn{1}{c}{3.00E-04} & \multicolumn{1}{c}{1024} & \multicolumn{1}{c}{15k} & \multicolumn{1}{c}{0.2} \\
    \midrule
    \multirow{2}[2]{*}{3D pose \& Mesh} & 3DPW  & \multicolumn{1}{c}{3.00E-05} & \multicolumn{1}{c}{512} & \multicolumn{1}{c}{10k} & \multicolumn{1}{c}{0.2} \\
          & Human3.6M & \multicolumn{1}{c}{3.00E-05} & \multicolumn{1}{c}{512} & \multicolumn{1}{c}{10k} & \multicolumn{1}{c}{0.2} \\
    \midrule
    Pedestrian detection & CrowdHuman & 1.00E-04      &   32    &   80k    & 0.2 \\
    \midrule
    \multirow{3}[2]{*}{Human parsing} & LIP   & 1.00E-04      &  1024     &  20k     & 0.3 \\
          & CIHP  & 1.00E-04      & 1024      &  20k     & 0.3 \\
          & Human3.6M & 5.00E-04      &  1024     &  20k     & 0.3 \\
    \midrule
    \multirow{2}[2]{*}{Attribute recognition } & RAPv2 &  6.00E-04     &   128    &  8k     & 0.2 \\
          & PA100k    &  3.00E-04     &    128   &   5k    & 0.2 \\
    \midrule
    Skeleton action & NTU60 & 1.00E-04      &  48     &   10k    & 0.2 \\
    \midrule
    Image caption & CUHK-PEDES &  1.00E-06     &   256    &  3k     &  0.2\\
    \bottomrule
    \end{tabular}%
  \label{tab:ft}%
\end{table*}%

\begin{table*}[htbp]
  \centering
  \caption{Detailed finetuning configs \textbf{with ViT-large backbone} for human-centric tasks.}
    \begin{tabular}{clcccc}
    \toprule
    Task Type  & \multicolumn{1}{c}{Dataset} & \multicolumn{1}{c}{Learning Rate} & \multicolumn{1}{c}{Batch Size} & \multicolumn{1}{c}{Iterations} & \multicolumn{1}{c}{Drop Path Rate} \\
    \midrule
    \multirow{2}[2]{*}{2D pose estimation} & COCO  & \multicolumn{1}{c}{3.00E-05} & \multicolumn{1}{c}{1024} & \multicolumn{1}{c}{10k} & \multicolumn{1}{c}{0.6} \\
          & AIC   & \multicolumn{1}{c}{3.00E-04} & \multicolumn{1}{c}{1024} & \multicolumn{1}{c}{15k} & \multicolumn{1}{c}{0.3} \\
    \midrule
    \multirow{2}[2]{*}{3D pose \& Mesh} & 3DPW  & \multicolumn{1}{c}{3.00E-05} & \multicolumn{1}{c}{512} & \multicolumn{1}{c}{10k} & \multicolumn{1}{c}{0.2} \\
          & Human3.6M & \multicolumn{1}{c}{3.00E-05} & \multicolumn{1}{c}{512} & \multicolumn{1}{c}{10k} & \multicolumn{1}{c}{0.2} \\
    \midrule
    Pedestrian detection & CrowdHuman &   1.00E-04    & 32      &    10k   & 0.5 \\
    \midrule
    \multirow{3}[2]{*}{Human parsing} & LIP   &  5.00E-05     &  512     &   10k    & 0.5 \\
          & CIHP  &  5.00E-05     &  512     &   10k    & 0.5 \\
          & Human3.6M & 1.00E-05      &  512     &  10k     & 0.5 \\
    \midrule
    \multirow{2}[2]{*}{Attribute recognition } & RAPv2 &  1.00E-03     &   128    &   6k    & 0.5 \\
          & PA-100k    &   7.00E-04    &   128    &   3k    & 0.5 \\
    \midrule
    Skeleton action & NTU60 & 1.00E-05      &  48     &  10k     & 0.5 \\
    \midrule
    Image caption & CUHK-PEDES &    1.00E-06     &   256    &  3k     &  0.5 \\
    \bottomrule
    \end{tabular}%
  \label{tab:ft_large}%
\end{table*}%

\section{Experimental results with complete comparison methods}
Due to the limited length of our manuscript, we provide experimental results with complete comparison methods here.

\begin{table}[t]
  \centering
  \caption{Pedestrian detection evaluation on CrowdHuman~\cite{shao2018crowdhuman} with mAP, MR$^{-2}$ and JI. Following UniHCP~\cite{ci2023unihcp}, we report the direct eval and fine-tune (FT) results. \dag indicates using a smaller input resolution with a maximum height/width of 1120.}
  \label{tab:detection}
  \resizebox{\linewidth}{!}{
    \begin{tabular}{cllccc}
    \toprule
    \multicolumn{2}{l}{Method} & Backbone & mAP & MR$^{-2}$$\downarrow$ & JI \\  
    \midrule
    \multirow{8}{*}{Specialist} 
      & DETR~\cite{carion2020end}  & \textcolor{black}{ResNet-50} & 75.9 & 73.2 & 74.4 \\
      & \textcolor{black}{CrowdDet~\cite{chu2020detection}}  & \textcolor{black}{ResNet-50} & 90.7 & 41.4 & 82.3 \\
      & \textcolor{black}{V2F-Net~\cite{shang2021v2f}}  & \textcolor{black}{ResNet-50} & 91.0 & 42.3 & - \\
      & PEDR~\cite{lin2020detr}  & \textcolor{black}{ResNet-50} & 91.6 & 43.7 & 83.3 \\
      & DDETR~\cite{zhu2020deformable} & \textcolor{black}{ResNet-50} & 91.5 & 43.7 & 83.1 \\
      & Sparse-RCNN~\cite{sun2021sparse} & \textcolor{black}{ResNet-50} & 91.3 & 44.8 & 81.3 \\
      & Iter-DDETR~\cite{zheng2022progressive} & \textcolor{black}{ResNet-50} & 92.1 & 41.5 & 84.0 \\
      & Iter-Sparse-RCNN~\cite{zheng2022progressive} & \textcolor{black}{ResNet-50} & \textbf{92.5} & 42.6 & 83.3 \\
    \cmidrule{2-6}  
      & Iter-DDETR~\cite{zheng2022progressive} & Swin-L & \textbf{94.1} & 37.7 & \textbf{87.1} \\
    \midrule
    \multirow{2}{*}{Pretraining} 
      & PATH~\cite{tang2023humanbench} & ViT-B & 90.9 & - & - \\
    \cmidrule{2-6}  
      & PATH~\cite{tang2023humanbench} & ViT-L & 90.8 & - & - \\
    \midrule
    \multirow{6}{*}{Generalist} 
      & UniHCP~\cite{ci2023unihcp} & ViT-B & 90.0 & 46.6 & 82.2 \\
      & UniHCP-FT~\cite{ci2023unihcp}  & ViT-B & \textbf{92.5} & 41.6 & 85.8 \\
      & Hulk\dag  & ViT-B & 90.7 & 43.8 & 84.0 \\
      & Hulk-FT\dag & ViT-B & 92.4 & \textbf{40.7} & \textbf{86.0} \\
    \cmidrule{2-6}  
      & Hulk\dag  & ViT-L & 92.2 & 40.1 & 85.8 \\
      & Hulk-FT\dag & ViT-L & 93.0 & \textbf{36.5} & 87.0 \\
    \bottomrule
    \end{tabular}%
  }
\end{table}

\begin{table}[t]
  \centering
  \caption{2D Pose estimation evaluation on COCO~\cite{coco} val set and on AIC~\cite{aic} test set with mAP. We compare our method with other SOTA methods, both specialists and generalists, with an input resolution of 256$\times$192. \dag indicates our implementation. \ddag indicates that multiple datasets are used.}
  \label{tab:pose}
    \begin{tabular}{cllcc}
    \toprule
    \multicolumn{2}{l}{Method } & Backbone & COCO  & AIC \\
    \midrule
    \multirow{10}[6]{*}{Specialist} 
    & HRNet~\cite{sun2019deep} & HRNet-W32 & 74.4  & - \\
          & HRNet~\cite{sun2019deep} & HRNet-W48 & 75.1  & - \\
          & TokenPose-L/D24~\cite{li2021tokenpose} & HRNet-W48 & 75.8  & - \\
          & HRFormer-B~\cite{yuan2021hrformer} & HRFormer-B & 75.6  & - \\
\cmidrule{2-5}          & ViTPose-B~\cite{xu2022vitpose} & ViT-B & 75.8  & - \\
          & ViTPose-B‡~\cite{xu2022vitpose} & ViT-B & 77.1  & 32.0  \\
          & ViTPose++-B‡~\cite{xu2023vitpose++} & ViT-B & 77.0  & 31.9  \\
\cmidrule{2-5}          
          & \textcolor{black}{ED-Pose~\cite{yang2023explicit}}&\textcolor{black}{Swin-L} &\textcolor{black}{75.8}&\textcolor{black}{-}\\
          & ViTPose-L~\cite{xu2022vitpose} & ViT-L & 78.3  & - \\
          & ViTPose-L‡~\cite{xu2022vitpose} & ViT-L & 78.7  & 34.5  \\
          & ViTPose++-L‡~\cite{xu2023vitpose++} & ViT-L & 78.6  & 34.3  \\
    \midrule
    \multirow{5}[2]{*}{Pretraining} 
          & HCMoCo\dag~\cite{hcmoco} & HRNet-W48 & 76.9  & - \\
          & SOLDIER~\cite{solider} & Swin-B & 76.6  & - \\
          & HAP~\cite{yuan2023hap}   & ViT-B & 77.0  & 32.3  \\
          & PATH~\cite{tang2023humanbench} & ViT-B & 76.3  & 35.0  \\
          \cmidrule{2-5}   
          & PATH~\cite{tang2023humanbench} & ViT-L & 77.1  & 36.3  \\
    \midrule
    \multirow{6}[4]{*}{Generalist} & UniHCP~\cite{ci2023unihcp} & ViT-B & 76.1  & 32.5  \\
          & UniHCP-FT~\cite{ci2023unihcp}  & ViT-B & 76.6  & 33.6  \\
          & Hulk  & ViT-B & 77.0  & 34.5  \\
          & Hulk-FT & ViT-B &  \textbf{77.5}     & \textbf{35.6}  \\
\cmidrule{2-5}          & Hulk  & ViT-L & 78.3  & 36.3  \\
          & Hulk-FT & ViT-L &  \textbf{78.7}    & \textbf{37.1} \\
    \bottomrule
    \end{tabular}%
\end{table}%

\begin{table}[t]
  \centering
  \caption{Human parsing evaluation on Human3.6m~\cite{h36m_pami}, LIP~\cite{lip} and CIHP~\cite{cihp} with mIoU. We compare our method with other SOTA methods, including specialists, generalists, and pretrainings, with an input resolution of 480$\times$480.~\dag indicated the re-implementation.}
  \label{tab:parsing}
    \begin{tabular}{cllccc}
    \toprule
\multicolumn{2}{l}{Method } & Backbone & H3.6M & LIP   & \multicolumn{1}{l}{CIHP} \\
    \midrule
    \multirow{4}[2]{*}{Specialist} 
          & SNT~\cite{ji2020learning}  & ResNet-101      & -     & 54.73  & 60.87  \\
          & PCNet~\cite{zhang2020part} &  ResNet-101     & -     & 57.03  & 61.05  \\
          & SCHP~\cite{li2020self}   &   ResNet-101    & -     & 59.36  & - \\
          & CDGNet~\cite{liu2022cdgnet} &    ResNet-101   & -     & 60.30  & 65.56  \\
    \midrule
    \multirow{5}[4]{*}{Pretraining} 
          & HCMoCo~\cite{hcmoco} & HRNet-18 & 62.50  & -     & - \\
          & HCMoCo\dag~\cite{hcmoco} & HRNet-48 &   66.01      &   -    & - \\
          & SOLDIER~\cite{solider} & Swin-B & -     & 60.50  & - \\
          & PATH~\cite{tang2023humanbench} & ViT-B & 65.00  & 61.40  & 66.80  \\
\cmidrule{2-6}  
          & PATH~\cite{tang2023humanbench} & ViT-L & 66.20  & 62.60  & 67.50  \\
    \midrule
    \multirow{6}[4]{*}{Generalist} & UniHCP~\cite{ci2023unihcp} & ViT-B & 65.90  & 63.80  & 68.60  \\
          & UniHCP-FT~\cite{ci2023unihcp} & ViT-B & 65.95  & 63.86  & 69.80  \\
          & Hulk  & ViT-B & 68.08  & 63.95  & 70.58  \\
          & Hulk-FT & ViT-B & \textbf{68.56}  & \textbf{63.98}   & \textbf{71.26}\\
\cmidrule{2-6}          & Hulk  & ViT-L & 69.31  & 65.86  & 72.33  \\
          & Hulk-FT & ViT-L & \textbf{69.89}  & \textbf{66.02}  & \textbf{72.68} \\
    \bottomrule
    \end{tabular}%

\end{table}%

\begin{table}[t]
  \centering
  \caption{Pedestrian attribute recognition evaluation on PA-100K~\cite{pa} and RAPv2~\cite{rapv2} with mA reported.}
  \label{tab:attribute}
    \begin{tabular}{cllcc}
    \toprule
    \multicolumn{2}{l}{Method } & Backbone & PA-100K & RAPv2 \\
    \midrule
    \multirow{5}[2]{*}{Specialist} & SSC~\cite{jia2021spatial}   & ResNet-50 & 81.87  & - \\
          & C-Tran~\cite{lanchantin2020general} & ResNet-101 & 81.53  & - \\
          & Q2L~\cite{liu2021query2label}    & TResNetL & 80.72  & - \\
          & L2L~\cite{li2022label2label}    & ViT-B & 82.24  & - \\
          & DAFL~\cite{jia2022learning}   & ResNet-50 & 83.54  & 81.04  \\
    \midrule
    \multirow{4}[2]{*}{Pretraining} & SOLIDER~\cite{solider} & Swin-B & 86.37  & - \\
          & HAP~\cite{yuan2023hap}   & ViT-B & 86.54  & 82.91  \\
          & PATH~\cite{tang2023humanbench} & ViT-B & 86.90  & 83.10  \\
\cmidrule{2-5}          
          & \textcolor{black}{PATH~\cite{tang2023humanbench}}  & \textcolor{black}{ViT-L} & \textcolor{black}{\textbf{90.80}}  & \textcolor{black}{\textbf{87.40}}  \\
    \midrule
    \multirow{6}[3]{*}{Generalist} & UniHCP~\cite{ci2023unihcp} & ViT-B & 79.32  & 77.20  \\
          & UniHCP-FT~\cite{ci2023unihcp}  & ViT-B & 86.18  & 82.34  \\
          & Hulk  & ViT-B & 82.85  & 80.90  \\
          & Hulk-FT & ViT-B &   \textbf{87.85}    & \textbf{85.26} \\
\cmidrule{2-5}          & Hulk  & ViT-L & 84.36  & 82.85  \\
          & Hulk-FT & ViT-L &  {88.97}     & {85.86 }\\
    \bottomrule
    \end{tabular}%
\vspace{-0.5em}
\end{table}%

\begin{table}[t]
  \centering
   \footnotesize
  \caption{Skeleton-based Action Recognition evaluation on NTU60-XSub~\cite{liu2019ntu} and FineGYM~\cite{shao2020finegym} with accuracy reported. }
  \label{tab:skeleton}
   \resizebox{.95\linewidth}{!}{
    \begin{tabular}{lllcc}
   
    \toprule
    Method &       & Backbone  & NTU60 & \textcolor{black}{GYM}\\
    \midrule
    \multicolumn{1}{c}{\multirow{9}[5]{*}{Specialist}} & ST-GCN~\cite{stgcn} & GCN   & 81.5 &\textcolor{black}{25.2}  \\
          & AS-GCN~\cite{li2019actional} & GCN   & 86.8  &\textcolor{black}{-}  \\

          & AGCN~\cite{agcn}  & GCN   & 88.5  &\textcolor{black}{-} \\

          & Shift-GCN~\cite{shiftgcn} & GCN   & 90.7  &\textcolor{black}{-} \\
          & CrosSCLR~\cite{sclr} & GCN   & 86.2 &\textcolor{black}{-}  \\
          & MCC~\cite{mcc}   & GCN   & 89.7 &\textcolor{black}{-}  \\
          & SCC~\cite{scc}   & GCN   & 88.0 &\textcolor{black}{-}  \\
          & UNIK~\cite{unik}  & GCN   & 86.8 &\textcolor{black}{-}  \\
          & CTR-GCN~\cite{ctrgcn} & GCN   & 92.4 &\textcolor{black}{-}   \\
          & MS-G3D~\cite{g3d} & GCN   & 91.5 &\textcolor{black}{92.0} \\
\cmidrule{2-5}          
& IndRNN~\cite{li2018independently} & RNN   & 81.8 &\textcolor{black}{-}  \\
          & MTCNN~\cite{ke2018learning} & CNN   & 81.1 &\textcolor{black}{-}  \\
          & HCN~\cite{li2018co}   & CNN   & 86.5 &\textcolor{black}{-}  \\
          & PoseConv3D~\cite{posec3d} & CNN   & 93.1  &\textcolor{black}{\textbf{92.4}} \\
          &SELFYNet~\cite{kwon2021learning} &CNN &- &\textcolor{black}{87.7}\\
                        \cmidrule{2-5}    
& ST-TR~\cite{plizzari2021skeleton} & Transformer & 89.9 &\textcolor{black}{-}  \\

          & DSTA-Net~\cite{shi2020decoupled} & Transformer & 91.5  &\textcolor{black}{-} \\
          & STTFormer~\cite{qiu2022spatio} & Transformer & 89.9&\textcolor{black}{-}   \\
              \midrule
      \multirow{2}[2]{*}{Pretraining}& MotionBERT~\cite{zhu2023motionbert} & DSTformer & 93.0 &\textcolor{black}{-} \\
     & SkeletonMAE~\cite{yan2023skeletonmae} & GIN  &- &\textcolor{black}{91.8}  \\
    \midrule
    \multirow{4}[3]{*}{Generalist} & Hulk  & ViT-B & 93.8 &\textcolor{black}{91.6}\\
          & Hulk-FT & ViT-B &  \textbf{94.0} &\textcolor{black}{92.2}\\
\cmidrule{2-5}          & Hulk  & ViT-L &  94.1 &\textcolor{black}{92.3}\\
          & Hulk-FT & ViT-L & \textbf{94.3}  &\textcolor{black}{\textbf{93.2}}\\
    \bottomrule
    \end{tabular}%
    }
    \vspace{-1em}
\end{table}%

\begin{table*}[t]
  \centering
  \caption{Monocular 3D human pose and mesh
recovery evaluation on 3DPW~\cite{3dpw} and Human3.6M~\cite{h36m_pami} among image-based methods.}
\label{tab:smpl}
\resizebox{.75\linewidth}{!}{
    \begin{tabular}{cllrrrrr}
    \toprule
    \multicolumn{2}{l}{\multirow{2}[4]{*}{Method}} & \multirow{2}[4]{*}{Backbone } & \multicolumn{3}{c}{3DPW} & \multicolumn{2}{c}{Human3.6M} \\
\cmidrule(r){4-6} \cmidrule(r){7-8}    \multicolumn{2}{l}{} &       & \multicolumn{1}{c}{MPVPE$\downarrow$} & \multicolumn{1}{c}{MPJPE$\downarrow$} & \multicolumn{1}{c}{PA-MPJPE$\downarrow$} & \multicolumn{1}{c}{MPJPE$\downarrow$} & \multicolumn{1}{c}{PA-MPJPE$\downarrow$} \\
    \midrule
    \multirow{20}[2]{*}{Specialist} 
     & HMR~\cite{hmr}   & ResNet-50 & \multicolumn{1}{c}{-} & \multicolumn{1}{c}{130.0 } & \multicolumn{1}{c}{76.7 } & \multicolumn{1}{c}{88.0 } & \multicolumn{1}{c}{56.8} \\
           & GraphCMR~\cite{graphcmr} & ResNet-50 & \multicolumn{1}{c}{-} & \multicolumn{1}{c}{-} & \multicolumn{1}{c}{70.2} & \multicolumn{1}{c}{-} & \multicolumn{1}{c}{50.1} \\
           & SPIN~\cite{spin}  & ResNet-50 & \multicolumn{1}{c}{116.4} & \multicolumn{1}{c}{96.9} & \multicolumn{1}{c}{59.2} & \multicolumn{1}{c}{62.5} & \multicolumn{1}{c}{41.1} \\
           & I2LMeshNet~\cite{moon2020i2l} & ResNet-50 & \multicolumn{1}{c}{-} & \multicolumn{1}{c}{93.2} & \multicolumn{1}{c}{57.7} & \multicolumn{1}{c}{55.7} & \multicolumn{1}{c}{41.1} \\
           & PyMAF~\cite{zhang2021pymaf} & ResNet-50 & \multicolumn{1}{c}{110.1} & \multicolumn{1}{c}{92.8} & \multicolumn{1}{c}{58.9} & \multicolumn{1}{c}{57.7} & \multicolumn{1}{c}{40.5} \\
          & ROMP~\cite{romp}  & ResNet-50 & \multicolumn{1}{c}{105.6} & \multicolumn{1}{c}{89.3} & \multicolumn{1}{c}{53.5} & \multicolumn{1}{c}{-} & \multicolumn{1}{c}{-} \\
          & ROMP~\cite{romp}  & HRNet-W32 & \multicolumn{1}{c}{103.1} & \multicolumn{1}{c}{85.5} & \multicolumn{1}{c}{53.3} & \multicolumn{1}{c}{-} & \multicolumn{1}{c}{-} \\
          & PARE~\cite{kocabas2021pare}  & ResNet-50 & \multicolumn{1}{c}{99.7} & \multicolumn{1}{c}{82.9} & \multicolumn{1}{c}{52.3} & \multicolumn{1}{c}{-} & \multicolumn{1}{c}{-} \\
          & METRO~\cite{metro} & ResNet-50 & \multicolumn{1}{c}{-} & \multicolumn{1}{c}{-} & \multicolumn{1}{c}{-} & \multicolumn{1}{c}{56.5} & \multicolumn{1}{c}{40.6} \\
          & METRO~\cite{metro} & HRNet-W64 & \multicolumn{1}{c}{88.2} & \multicolumn{1}{c}{77.1} & \multicolumn{1}{c}{47.9} & \multicolumn{1}{c}{54.0} & \multicolumn{1}{c}{36.7} \\
          & PARE~\cite{kocabas2021pare}  & HRNet-W32 & \multicolumn{1}{c}{88.6} & \multicolumn{1}{c}{74.5} & \multicolumn{1}{c}{46.5} & \multicolumn{1}{c}{-} & \multicolumn{1}{c}{-} \\
          & MeshGraphormer~\cite{graphormer} & HRNet-W64 & \multicolumn{1}{c}{87.7} & \multicolumn{1}{c}{74.7} & \multicolumn{1}{c}{45.6} & \multicolumn{1}{c}{51.2} & \multicolumn{1}{c}{34.5} \\
          & ProHMR~\cite{prohmr} & ResNet-50      & \multicolumn{1}{c}{-} & \multicolumn{1}{c}{-} & \multicolumn{1}{c}{59.8} & \multicolumn{1}{c}{-} & \multicolumn{1}{c}{41.2} \\
          & OCHMR~\cite{ochmr} & ResNet-50      & \multicolumn{1}{c}{107.1} & \multicolumn{1}{c}{89.7} & \multicolumn{1}{c}{58.3} & \multicolumn{1}{c}{-} & \multicolumn{1}{c}{-} \\
          & 3DCrowdNet~\cite{crowdnet} & ResNet-50      & \multicolumn{1}{c}{98.3} & \multicolumn{1}{c}{81.7} & \multicolumn{1}{c}{51.5} & \multicolumn{1}{c}{-} & \multicolumn{1}{c}{-} \\
          & FastMETRO~\cite{fastmetro} & ResNet-50 & \multicolumn{1}{c}{90.6} & \multicolumn{1}{c}{77.9} & \multicolumn{1}{c}{48.3} & \multicolumn{1}{c}{53.9} & \multicolumn{1}{c}{37.3} \\
          & FastMETRO~\cite{fastmetro} & HRNet-W64 & \multicolumn{1}{c}{84.1} & \multicolumn{1}{c}{73.5} & \multicolumn{1}{c}{44.6} & \multicolumn{1}{c}{52.2} & \multicolumn{1}{c}{33.7} \\
          & CLIFF~\cite{li2022cliff} & ResNet-50      & \multicolumn{1}{c}{81.2} & \multicolumn{1}{c}{69.0} & \multicolumn{1}{c}{43.0} & \multicolumn{1}{c}{47.1} & \multicolumn{1}{c}{32.7} \\
          & VisDB~\cite{visdb} & ResNet-50      & \multicolumn{1}{c}{85.5} & \multicolumn{1}{c}{73.5} & \multicolumn{1}{c}{44.9} & \multicolumn{1}{c}{51.0} & \multicolumn{1}{c}{34.5} \\
    \midrule
    \multirow{2}[1]{*}{Pretraining} 
          & MotionBERT~\cite{zhu2023motionbert} & DSTformer & \multicolumn{1}{c}{88.1} & \multicolumn{1}{c}{76.9} & \multicolumn{1}{c}{47.2} & \multicolumn{1}{c}{53.8} & \multicolumn{1}{c}{34.9} \\
          & HAP~\cite{yuan2023hap}   & ViT-B & \multicolumn{1}{c}{90.1} & \multicolumn{1}{c}{56.0} & \multicolumn{1}{c}{106.3} & \multicolumn{1}{c}{-} & \multicolumn{1}{c}{-} \\
        \midrule
    \multirow{4}[3]{*}{Generalist} & Hulk  & ViT-B & \multicolumn{1}{c}{\textbf{79.8}} & \multicolumn{1}{c}{\textbf{67.0}} & \multicolumn{1}{c}{\textbf{39.9}} & \multicolumn{1}{c}{\textbf{43.6}} & \multicolumn{1}{c}{\textbf{31.9}} \\
          & Hulk-FT & ViT-B & \multicolumn{1}{c}{80.7} & \multicolumn{1}{c}{68.9} & \multicolumn{1}{c}{41.3} & \multicolumn{1}{c}{44.9} & \multicolumn{1}{c}{32.0} \\
\cmidrule{2-8}          & Hulk  & ViT-L & \multicolumn{1}{c}{\textbf{77.4}} & \multicolumn{1}{c}{\textbf{66.3}} & \multicolumn{1}{c}{\textbf{38.5}} & \multicolumn{1}{c}{\textbf{40.3}} & \multicolumn{1}{c}{\textbf{28.8}} \\
          & Hulk-FT & ViT-L& \multicolumn{1}{c}{79.9} & \multicolumn{1}{c}{68.3} & \multicolumn{1}{c}{40.6} & \multicolumn{1}{c}{41.4} & \multicolumn{1}{c}{30.2} \\
    \bottomrule
    \end{tabular}%
    \vspace{-0.5em}
}
\end{table*}%

\clearpage

\end{document}